\documentclass[11pt]{article}
\usepackage[preprint]{acl}
\usepackage{latexsym}
\usepackage[utf8]{inputenc}
\usepackage[T1]{fontenc}
\usepackage{inconsolata}
\usepackage{xspace}

\usepackage{graphicx}
\usepackage{xcolor}
\definecolor{deepslate}{HTML}{1E3A66}
\definecolor{coolblue}{HTML}{4F8FD6}
\definecolor{softperi}{HTML}{8FA6E3}
\definecolor{warmcream}{HTML}{F2D6A6}
\definecolor{lavblue}{HTML}{C5B5E8}
\definecolor{coralred}{HTML}{E76B6B}
\definecolor{nearwhite}{HTML}{F7F8FB}
\usepackage{colortbl}
\usepackage[most]{tcolorbox}

\usepackage{booktabs}
\usepackage{multirow}
\usepackage{adjustbox}
\usepackage{makecell}
\usepackage{tabularx}
\usepackage{longtable}
\usepackage{array}
\usepackage{subcaption}
\usepackage{wrapfig}

\usepackage[patch=none]{microtype}
\usepackage{times}
\usepackage{caption}
\usepackage{stfloats}
\usepackage{lineno}
\usepackage{textcomp}
\usepackage{balance}
\usepackage{enumitem}

\usepackage[ruled,linesnumbered]{algorithm2e}
\usepackage{algpseudocode}
\usepackage{listings}
\usepackage{verbatim}

\usepackage{amsmath}
\usepackage{amsfonts}
\usepackage{bm}
\usepackage{amssymb}
\usepackage{amsthm}

\usepackage{pifont}

\SetKwInput{KwInput}{Input}

\SetKwInput{KwOutput}{Output}

\SetKwComment{Block}{$\triangleright$\ }{}

\SetKwComment{Comment}{//  }{}

\DontPrintSemicolon

\SetKwIF{If}{ElseIf}{Else}{if}{}{else if}{else}{end if}%

\def\HiLi{\leavevmode\rlap{\hbox to \hsize{\color{gray!20}\leaders\hrule height .8\baselineskip depth .5ex\hfill}}}

\definecolor{darkblue}{rgb}{0, 0, 0.5}
\definecolor{bestcell}{HTML}{D4EDDA}
\definecolor{worstcell}{HTML}{F8D7DA}
\definecolor{baselinerow}{HTML}{EAEAEA}

\hypersetup{
  colorlinks=true,   
  citecolor=darkblue, 
  linkcolor=darkblue, 
  urlcolor=darkblue   
}

\theoremstyle{plain}

\theoremstyle{definition}

\theoremstyle{remark}

\def\eqref#1{equation~\ref{#1}}

\def\1{\bm{1}}

\DeclareMathAlphabet{\mathsfit}{\encodingdefault}{\sfdefault}{m}{sl}
\SetMathAlphabet{\mathsfit}{bold}{\encodingdefault}{\sfdefault}{bx}{n}

\newcommand{\WithinReduction}{8.5}
\newcommand{\WithinCovarianceShare}{76.4}
\newcommand{\WithinLowerCount}{271}

\newcommand{\portbench}{\textsc{PortBench}\xspace}
\newcommand{\ceps}{CEPS\xspace}
\newcommand{\mbdataset}{market base dataset\xspace}

\title{\portbench: A Correlation-Aware, Full-Pipeline Benchmark\\
for LLM-Driven Portfolio Management
}

\author{%
  Yuxuan Zhao\textsuperscript{1,2} \quad
  Sijia Chen\textsuperscript{2} \quad
  Ningxin Su\textsuperscript{2} \\
  \textsuperscript{1}Yantai Research Institute of Harbin Engineering University \\
  \textsuperscript{2}The Hong Kong University of Science and Technology (Guangzhou) \\
  \texttt{sijiachen@hkust-gz.edu.cn}\\
  Project page: \href{https://portbench.github.io/}{\textcolor[HTML]{DA3D8F}{https://portbench.github.io/}}
}

\begin{document}
\maketitle

\begin{abstract}
Large language models (LLMs) have shown strong performance across diverse financial tasks, yet portfolio management (PM) remains poorly evaluated. Existing benchmarks exhibit two gaps: they are often equity-only and ignore cross-asset correlations; they provide limited insight into where errors arise and how they affect subsequent decisions. We introduce \portbench, a benchmark with market data spanning six asset classes from 2015 to 2025. \portbench comprises a static QA dataset of 6,269 questions across seven task templates and a dynamic five-stage portfolio evaluation. Dynamic evaluation uses a class-level correlation score for allocation and \ceps to summarize reference-based stage scores. We further evaluate under three stress windows and three risk profiles, and support real-time evaluation on newly arriving market data. Evaluating ten LLMs reveals a disconnect: high financial QA scores do not ensure consistent portfolio outperformance. LLMs concentrate exposure in correlated assets, and their portfolios achieve higher Sharpe than equal weighting in 54.2\% of 120 evaluations across four market periods. Our source code is available at \href{https://github.com/AgenticFinLab/portbench}{this https URL}.
\end{abstract}

\addtocontents{toc}{\protect\setcounter{tocdepth}{-1}}

\section{Introduction}
\label{sec:intro}

Large language models (LLMs) have shown increasing capabilities across diverse financial tasks, spurring benchmarks for financial knowledge, numerical reasoning, and investment decision-making~\citep{convfinqa-emnlp22, pixiu-neurips23, finben-neurips24, fineval-naacl25, financereasoning-acl25}. Portfolio management (PM), however, remains inadequately evaluated. PM requires diversified allocations to reduce volatility and risk, which in turn requires accounting for intra- and inter-class correlations~\citep{portfolio-selection-jof,risk-parity-panagora}. Investment decisions put capital at risk. Assessing the reliability of PM systems therefore requires evaluating each stage and how its errors affect later decisions.

However, existing financial benchmarks fail to comprehensively evaluate PM due to two gaps. First, they often restrict coverage to a single asset class~\citep{finrl-neurips22,finben-neurips24,cryptotrade-emnlp24,stockbench-arxiv25,democratizing-alpha-icaif25}; even in multi-asset settings, assets are evaluated in isolation~\citep{investorbench-acl25}, which ignores cross-asset correlation structures. This design fails to distinguish between highly concentrated portfolios and genuinely diversified ones, even when their returns are identical. Additionally, LLM-based multi-agent systems for portfolio construction are consistently evaluated on equities alone with proprietary backtests that differ in data period, stock pools, and metrics, making cross-method comparison infeasible~\citep{Fincon-neurips24,MASS-arxiv25}.

Second, existing benchmarks provide limited diagnosis of the investment decision process. Most financial benchmarks rely on static QA that probes knowledge without linking answers to realistic allocation~\citep{pixiu-neurips23,finben-neurips24,financereasoning-acl25}. The few that test portfolio decisions use single-step predictions or cover only part of the workflow~\citep{stockbench-arxiv25,finripple-acl25,agents-for-investment-management-icaif25}; none spans the full cycle of market interpretation, signal generation, weight optimization, execution, and risk monitoring. These evaluations leave unclear where errors arise and how they affect subsequent decisions. Most further assume one implicit risk profile in normal markets, leaving the resilience of PM strategies under stress and alignment with investor-specific risk tolerances largely untested~\citep{bench-prioritize-risk-arxiv25, finsaber-kdd26}.

\begin{figure*}[t]
\centering
\includegraphics[width=\textwidth]{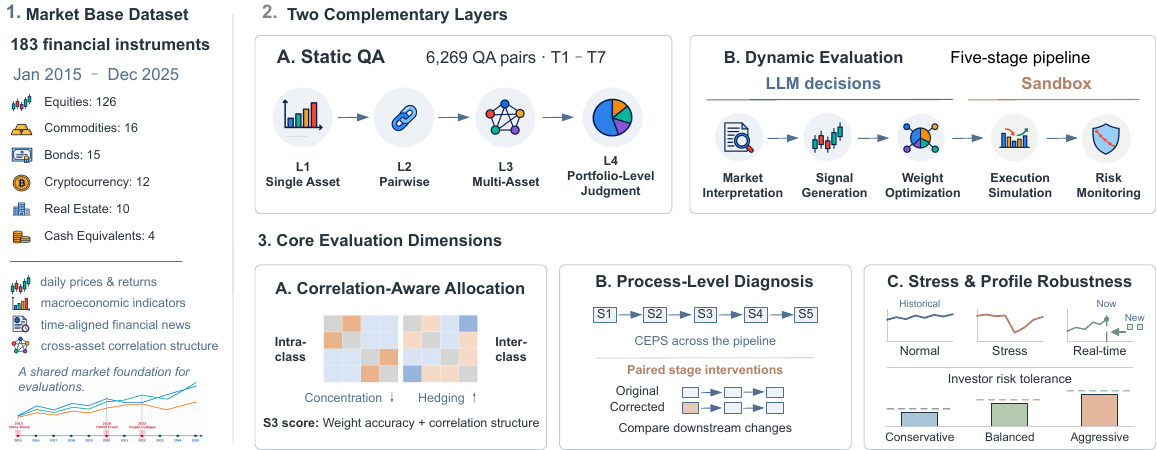}
\caption{Overview of \portbench. A shared multi-asset market base supports complementary static QA and dynamic five-stage evaluation. The benchmark assesses correlation-aware allocation, stagewise diagnostics through \ceps and paired interventions, and robustness across evaluation periods and investor risk profiles.}
\label{fig:portbench_overview}
\end{figure*}

To address these gaps, we introduce \portbench, a benchmark for LLM-driven PM with market data spanning six asset classes from 2015 to 2025. \portbench evaluates LLMs through two complementary layers: a static QA dataset probing financial reasoning, and a dynamic five-stage sandbox that evaluates decisions at each stage of the PM cycle under sequential market conditions. Beyond historical replay, \portbench also supports real-time evaluation on up-to-date market data. Figure~\ref{fig:portbench_overview} provides an overview. Our contributions are:
\begin{itemize}[leftmargin=*,topsep=0pt,itemsep=-0.2em]
    \item \textbf{A Multi-Asset Market Dataset.} We construct a 2015--2025 corpus of 183 market series across six asset classes, with prices, macro series, and time-aligned news, enabling correlation-aware PM evaluation beyond single-asset settings.
    \item \textbf{A Dual-Layer Evaluation Framework.} We pair static QA with a five-stage pipeline to evaluate financial reasoning and its use in portfolio decisions. Within the pipeline, \ceps summarizes reference-based stage scores, while paired interventions trace how local changes affect later decisions. We then evaluate the resulting portfolios across historical stress windows and investor profiles, with real-time evaluation extending coverage beyond historical replay.
    \item \textbf{A QA--Portfolio Performance Gap.} Across the ten evaluated LLMs, high financial QA scores do not ensure consistent portfolio outperformance. Models concentrate exposure in correlated assets, and their portfolios beat equal weighting on Sharpe in 54.2\% of 120 evaluations across four historical market periods.
\end{itemize}

\section{PortBench}
\label{sec:method}

\subsection{Benchmark Construction}
\label{sec:data}

We construct the \textbf{\mbdataset}, covering six asset classes: equities (126 tickers), commodities (16 tickers), bonds (15 series), cryptocurrency (12 tickers), real estate (10 series), and cash equivalents (4 series). In total, the dataset comprises 183 market series spanning January~2015 through December~2025. For each instrument, we collect daily price histories, return series, and associated news text; macroeconomic indicators include interest rates, inflation measures, credit spreads, and volatility indices. Stress windows are drawn from within this range and do not overlap with the normal period. Correlation analysis of the \mbdataset reveals that inter-class average correlations are generally low, while intra-class correlations are strongly positive; see Appendix~\ref{app:market_base_dataset} for details.

\subsection{Evaluation Framework}
\label{sec:eval_framework}

\begin{figure*}[t]
\centering
\includegraphics[width=\textwidth]{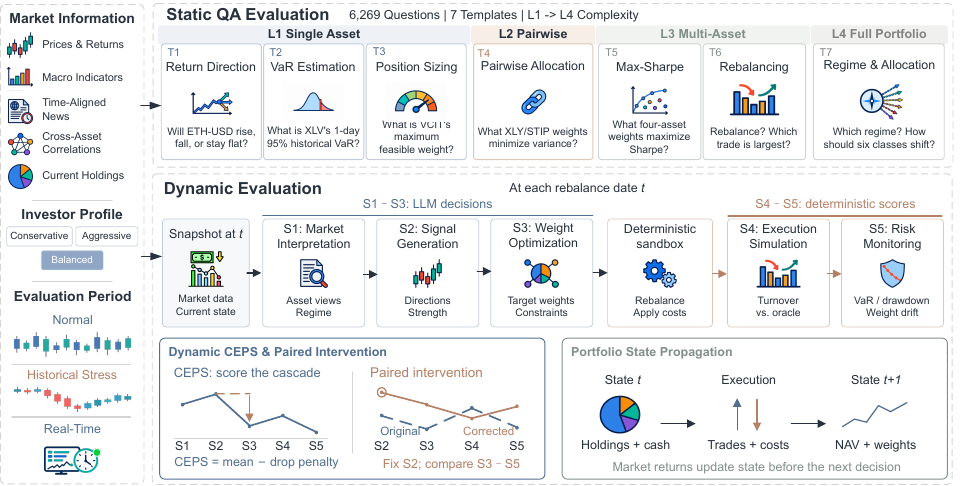}
\caption{Overview of the evaluation framework.
\textbf{Top:} static QA spans seven templates (T1--T7), from single-asset tasks at L1 to full-portfolio judgment at L4.
\textbf{Bottom:} at each rebalance, the LLM produces S1--S3, while the sandbox executes the allocation and evaluates execution and risk at S4--S5. \ceps summarizes reference-based stage scores, paired interventions test downstream effects, and the portfolio state carries into the next decision.}
\label{fig:evaluation_framework}
\end{figure*}

Building on the \mbdataset, \portbench evaluates LLMs through two complementary layers, as illustrated in Figure~\ref{fig:evaluation_framework}.

\textbf{QA evaluation.} In the static layer, 6,269 QA pairs are generated across seven task templates spanning four complexity levels (L1--L4), probing financial reasoning abilities from single-asset prediction and sizing (L1: T1--T3) through pairwise allocation (L2: T4) and constrained multi-asset optimization (L3: T5--T6) to regime-driven judgment (L4: T7). Automated generation from historical data enables scalable template additions and dataset regeneration without manual annotation. Representative QA samples are in Appendix~\ref{app:showcase_qa}.

\textbf{Dynamic evaluation.} The \mbdataset is replayed point-in-time across the evaluation window. At each rebalance, the model receives a market snapshot and sequentially produces S1 (Market Interpretation), which assigns sentiment scores and identifies the prevailing regime; S2 (Signal Generation), which maps these views to directional signals; and S3 (Weight Optimization), which proposes portfolio weights. The sandbox executes the S3 allocation under fixed transaction costs. S4 (Execution Simulation) scores turnover agreement with the reference, while S5 (Risk Monitoring) compares lookback VaR, drawdown, and a rule-based rebalance flag computed from the proposed and reference allocations. Both deterministically evaluate S3 under identical execution and risk rules across models. A stateful sandbox records stage scores, proposed weights, and the resulting NAV trajectory, linking allocations to realized costs, returns, and risk under a common execution policy. Beyond historical replay, \portbench supports real-time evaluation on recent market data. Appendix~\ref{app:evaluation} details evaluation and replay; Appendix~\ref{app:showcase} shows snapshots and full pipeline traces.

\textbf{Pipeline metrics.} \ceps summarizes reference-based stage scores with a penalty for downstream score drops. Let $\sigma_t \in [0,1]$ denote the normalized score at pipeline stage $t \in \{1,\ldots,5\}$, and let $\bar{\sigma} = \frac{1}{5}\sum_{t=1}^{5} \sigma_t$ be the mean stage score. With cascade penalty weight $\lambda = 0.1$, $\text{\ceps} = \operatorname{clip}\!\left(\bar{\sigma} - \lambda \sum_{t=1}^{4} \max(\sigma_t - \sigma_{t+1},\ 0),\ 0,\ 1\right)$. At the same mean score, \ceps assigns a lower value to a pipeline with larger downstream drops, penalizing higher upstream scores followed by lower downstream scores. For \textbf{S3}, we combine reference-weight similarity with class-level correlation structure: $\sigma_3 = \alpha\, s_\text{acc}(\mathbf{w}, \mathbf{w}^*) + (1-\alpha)\, s_\text{corr}(\mathbf{w})$, with $\alpha = 0.5$. Here $s_\text{acc}(\mathbf{w}, \mathbf{w}^*)=1-\frac12\sum_i|w_i-w_i^*|$ measures total-variation proximity to a signal- and profile-constrained reference allocation $\mathbf{w}^*$ estimated from the lookback window, while $s_\text{corr}$ combines exposure to internally correlated classes with cross-class hedging. The lookback reference supplies a reproducible decision-time target; returns and risk are evaluated from the resulting portfolios. Asset-level covariance analysis complements the class-level score by testing how within-class weighting affects portfolio risk at fixed asset selection and class exposure, see Section~\ref{sec:analysis_baselines} for details. Appendix~\ref{app:metrics} defines the scores, and Appendix~\ref{app:parameter_sensitivity} examines ranking sensitivity to $\lambda$ and $\alpha$.

\textbf{Paired stage interventions.} To measure downstream effects, \portbench replaces one stage output with its reference output, reruns the subsequent stages, and compares their scores with the original branch. This auxiliary diagnostic scores generated S4/S5 plans; the main ranking scores turnover and lookback risk-state agreement (Appendix~\ref{app:paired_details}).

\textbf{Stress windows and investor profiles.} We report normal-period CEPS with a stress drawdown gate under three historical windows (2015 China Shock, 2020 COVID Crash, 2022 Market Decline), and across three investor profiles (conservative, balanced, aggressive) whose exposure and drawdown constraints condition the S1/S2 prompts and their resulting signals. We compare matched allocations and realized returns and risk across profiles. Appendix~\ref{app:stress_profiles} gives profile limits and gate criteria.

\section{Experiments}
\label{sec:experiments}

\subsection{Experimental Setup}
\label{sec:setup}

\textbf{LLMs.} We evaluate ten frontier LLMs: Kimi-K2.6~\citep{kimi-k2.6-blog}, Qwen3.7-Max~\citep{qwen3.7-blog}, Qwen3.6-Plus~\citep{qwen3.6plus-blog}, Qwen3.6-35B-A3B~\citep{qwen36_35b_a3b}, GLM-5.1~\citep{glm-5-arxiv26}, Doubao-Seed-2.0-Lite and Doubao-Seed-2.0-Pro (DB-2.0-Lite/Pro)~\citep{doubaoseed2.0-blog}, Hunyuan3-Preview (HY3-Preview)~\citep{hy3-preview-blog}, and DeepSeek-V4-Flash and DeepSeek-V4-Pro (DS-V4-Flash/Pro)~\citep{deepseekv4-hf}.

\textbf{Evaluation Protocol.} We set the temperature to 0 and maximum output length to 8192 tokens, with thinking mode enabled by default. For static QA, each model answers the same 50 test questions per template using zero-shot prompting. For dynamic evaluation, models produce S1--S3 outputs and are evaluated through the five-stage pipeline under three investor profiles, using monthly decisions in the normal period and weekly decisions during stress windows to capture portfolio responses to rapidly changing market conditions. We use a 2024 bull-market window as the normal period and three historical stress windows for the stress gate and cross-period Sharpe comparisons. All strategies use the same eligible assets, next-close execution, and transaction costs; see Appendix~\ref{app:evaluation} for details.

\textbf{Baselines.} We compare LLM-driven portfolios against six classical strategies: (1) Equal-Weight (EqW, $1/n$) allocates capital uniformly across assets; (2) 60/40 allocates 60\% to equities and 40\% to bonds; (3) Risk Parity (RiskPar) weights assets inversely to volatility; (4) Covariance Risk Parity (CovRiskPar) extends RiskPar using the full covariance matrix for equal risk contribution; (5) Minimum Variance (MinVar) selects the long-only portfolio on the Markowitz efficient frontier that minimizes variance~\citep{portfolio-selection-jof}; (6) Black--Litterman (BL) blends prior returns with simple views. These strategies retain their standard rules across profiles and are evaluated on Sharpe ratio, maximum drawdown, volatility, and total return.

\subsection{Static QA Evaluation}
\label{sec:qa_results}

\begin{table}[t]
\centering
\begin{adjustbox}{width=\columnwidth,center}
\begin{tabular}{lccccccccc}
\toprule
\textbf{Model} & \textbf{T1} & \textbf{T2} & \textbf{T3} & \textbf{T4} & \textbf{T5} & \textbf{T6} & \textbf{T7} & \textbf{Mean} \\
\midrule
DS-V4-Flash      & \textbf{.520} & .843 & .945 & \textbf{1.00} & .932 & .652 & \textbf{.843} & \textbf{.819} \\
Qwen3.7-Max      & .500 & \textbf{.859} & .951 & \textbf{1.00} & .954 & .724 & .742 & \textbf{.819} \\
DS-V4-Pro        & \textbf{.520} & .837 & .963 & \textbf{1.00} & \textbf{.992} & .652 & .760 & .818 \\
DB-2.0-Lite      & .460 & .798 & .957 & .956 & .897 & .810 & .747 & .804 \\
DB-2.0-Pro       & .440 & .847 & .963 & .991 & .912 & .824 & .530 & .787 \\
Qwen3.6-Plus     & .440 & .858 & \textbf{.968} & \textbf{1.00} & .804 & .640 & .768 & .783 \\
GLM-5.1          & .440 & .855 & .964 & \textbf{1.00} & .421 & \textbf{.882} & .738 & .757 \\
Qwen3.6-35B-A3B  & .460 & .808 & .961 & \textbf{1.00} & .230 & .564 & .763 & .684 \\
HY3-Preview      & .460 & .386 & .336 & .975 & .958 & .468 & .783 & .624 \\
Kimi-K2.6        & .420 & .422 & .493 & .956 & .280 & .684 & .320 & .511 \\
\bottomrule
\end{tabular}
\end{adjustbox}
\caption{Mean QA score by task template.}
\label{tab:qa_results}
\end{table}

\textbf{Portfolio QA performance is task-dependent: models excel at formula-based allocation but remain weak on prediction and rebalancing.} Table~\ref{tab:qa_results} shows that eight of ten models exceed 0.94 on T3; six score 1.00 on T4, and all reach at least 0.956. In contrast, none exceeds 0.520 on return prediction (T1), while rebalancing scores (T6) range from 0.468 to 0.882. T4--T5 (allocation with supplied covariance statistics) average 0.863, versus 0.652 for T1--T2 and T6--T7 (prediction, risk estimation, rebalancing, and regime detection); this 0.211 gap favors T4--T5 for eight of ten models (Appendix~\ref{app:formula_judgment}). Supplementary structured T3/T4 tasks show that several LLMs retain near-ceiling performance on position-sizing and allocation-constraint reasoning, with details in Appendix~\ref{app:qa_saturation}.

\subsection{Pipeline Evaluation}
\label{sec:pipeline_results}

\begin{table*}[t]
\centering
\small
\begin{adjustbox}{max width=\textwidth,center}
\begin{tabular}{ll rrrrrr rrrr c}
\toprule
\textbf{Profile} & \textbf{Model} & \textbf{S1} & \textbf{S2} & \textbf{S3} & \textbf{S4} & \textbf{S5} & \textbf{CEPS}$\uparrow$ & \textbf{Sharpe}$\uparrow$ & \textbf{Ret\%}$\uparrow$ & \textbf{MaxDD\%}$\uparrow$ & \textbf{Vol\%}$\downarrow$ & \textbf{Gate} \\
\midrule
\multirow{10}{*}{\rotatebox[origin=c]{90}{\textbf{Conservative}}}
& GLM-5.1           & .763 & .438 & .301 & .304 & .091 & \cellcolor{bestcell}\textbf{.299} & 0.707 & 11.82 & $-$7.87 & 11.00 & \cellcolor{worstcell}$\times$ \\
& Qwen3.6-Plus      & .774 & \cellcolor{bestcell}\textbf{.491} & \cellcolor{bestcell}\textbf{.310} & .153 & \cellcolor{bestcell}\textbf{.136} & .292 & 0.875 & \cellcolor{bestcell}\textbf{14.97} & $-$9.56 & 12.22 & \cellcolor{worstcell}$\times$ \\
& Qwen3.6-35B-A3B   & .754 & .405 & .304 & .217 & .045 & .263 & \cellcolor{bestcell}\textbf{0.992} & 13.55 & $-$6.19 & 9.21 & \cellcolor{worstcell}$\times$ \\
& Kimi-K2.6         & .768 & .394 & .303 & .269 & .000 & .261 & 0.445 & 8.25 & $-$6.59 & 9.94 & \cellcolor{worstcell}$\times$ \\
& DS-V4-Flash       & .742 & .315 & .298 & \cellcolor{bestcell}\textbf{.320} & .022 & .253 & $-$0.029 & 3.32 & $-$8.87 & 9.51 & \cellcolor{worstcell}$\times$ \\
& HY3-Preview       & \cellcolor{bestcell}\textbf{.778} & .489 & .290 & .121 & .000 & .251 & 0.547 & 9.00 & $-$6.38 & 9.21 & \cellcolor{worstcell}$\times$ \\
& DS-V4-Pro         & .749 & .339 & .302 & .217 & .000 & .239 & 0.616 & 10.07 & $-$7.50 & 9.87 & \cellcolor{worstcell}$\times$ \\
& DB-2.0-Pro        & .750 & .403 & .295 & .143 & .000 & .236 & 0.571 & 8.77 & \cellcolor{bestcell}\textbf{$-$5.73} & \cellcolor{bestcell}\textbf{8.34} & \cellcolor{worstcell}$\times$ \\
& Qwen3.7-Max       & .746 & .385 & .291 & .151 & .000 & .232 & 0.578 & 9.16 & $-$5.97 & 8.94 & \cellcolor{worstcell}$\times$ \\
& DB-2.0-Lite       & .751 & .339 & .299 & .143 & .000 & .223 & 0.740 & 10.79 & $-$6.46 & 8.98 & \cellcolor{worstcell}$\times$ \\
\midrule
\multirow{10}{*}{\rotatebox[origin=c]{90}{\textbf{Balanced}}}
& GLM-5.1           & .773 & .472 & \cellcolor{bestcell}\textbf{.303} & \cellcolor{bestcell}\textbf{.296} & \cellcolor{bestcell}\textbf{.409} & \cellcolor{bestcell}\textbf{.384} & 0.603 & 13.10 & $-$14.20 & 15.89 & \cellcolor{bestcell}\checkmark \\
& Qwen3.6-Plus      & \cellcolor{bestcell}\textbf{.793} & \cellcolor{bestcell}\textbf{.531} & .298 & .209 & .091 & .300 & 0.798 & 13.99 & $-$10.11 & 12.35 & \cellcolor{bestcell}\checkmark \\
& Qwen3.7-Max       & .774 & .451 & .297 & .158 & .182 & .292 & 0.783 & \cellcolor{bestcell}\textbf{14.97} & $-$11.02 & 13.95 & \cellcolor{bestcell}\checkmark \\
& Kimi-K2.6         & .786 & .464 & .297 & .247 & .000 & .274 & 0.769 & 14.29 & $-$9.34 & 13.32 & \cellcolor{bestcell}\checkmark \\
& Qwen3.6-35B-A3B   & .772 & .476 & .295 & .237 & .000 & .272 & 0.740 & 12.87 & $-$7.77 & 11.91 & \cellcolor{bestcell}\checkmark \\
& DS-V4-Pro         & .764 & .387 & .301 & .289 & .005 & .266 & 0.542 & 9.46 & $-$6.93 & 10.29 & \cellcolor{bestcell}\checkmark \\
& DB-2.0-Pro        & .778 & .483 & .294 & .179 & .000 & .263 & 0.662 & 12.09 & $-$9.83 & 12.36 & \cellcolor{bestcell}\checkmark \\
& HY3-Preview       & .791 & \cellcolor{bestcell}\textbf{.531} & .264 & .133 & .000 & .258 & 0.650 & 11.52 & $-$8.05 & 11.66 & \cellcolor{bestcell}\checkmark \\
& DB-2.0-Lite       & .773 & .387 & .298 & .137 & .045 & .244 & \cellcolor{bestcell}\textbf{0.838} & 14.01 & $-$8.89 & 11.70 & \cellcolor{bestcell}\checkmark \\
& DS-V4-Flash       & .774 & .321 & .290 & .172 & .000 & .226 & 0.609 & 9.89 & \cellcolor{bestcell}\textbf{$-$6.56} & \cellcolor{bestcell}\textbf{9.68} & \cellcolor{worstcell}$\times$ \\
\midrule
\multirow{10}{*}{\rotatebox[origin=c]{90}{\textbf{Aggressive}}}
& GLM-5.1           & .779 & .510 & .295 & \cellcolor{bestcell}\textbf{.310} & \cellcolor{bestcell}\textbf{.409} & \cellcolor{bestcell}\textbf{.393} & 0.930 & 21.00 & $-$15.13 & 17.92 & \cellcolor{bestcell}\checkmark \\
& Qwen3.6-Plus      & .802 & .566 & .297 & .162 & \cellcolor{bestcell}\textbf{.409} & .366 & \cellcolor{bestcell}\textbf{1.034} & \cellcolor{bestcell}\textbf{21.84} & $-$13.05 & 16.57 & \cellcolor{bestcell}\checkmark \\
& Qwen3.6-35B-A3B   & .786 & .553 & .299 & .252 & .318 & .365 & 0.845 & 19.49 & $-$16.83 & 18.33 & \cellcolor{bestcell}\checkmark \\
& DB-2.0-Lite       & .787 & .478 & .300 & .138 & .364 & .331 & 0.916 & 18.48 & $-$12.76 & 15.45 & \cellcolor{bestcell}\checkmark \\
& DB-2.0-Pro        & .787 & .494 & .300 & .168 & .273 & .325 & 0.867 & 16.15 & $-$10.85 & 13.75 & \cellcolor{bestcell}\checkmark \\
& Qwen3.7-Max       & .781 & .463 & \cellcolor{bestcell}\textbf{.303} & .190 & .273 & .323 & 0.733 & 15.41 & $-$13.20 & 15.79 & \cellcolor{bestcell}\checkmark \\
& Kimi-K2.6         & .790 & .514 & .293 & .199 & .136 & .304 & 0.890 & 19.34 & $-$13.08 & 16.99 & \cellcolor{bestcell}\checkmark \\
& DS-V4-Pro         & .783 & .463 & .303 & .250 & .014 & .278 & 0.653 & 10.34 & \cellcolor{bestcell}\textbf{$-$6.63} & \cellcolor{bestcell}\textbf{9.65} & \cellcolor{bestcell}\checkmark \\
& HY3-Preview       & \cellcolor{bestcell}\textbf{.807} & \cellcolor{bestcell}\textbf{.574} & .292 & .132 & .000 & .274 & 0.673 & 11.78 & $-$8.74 & 11.61 & \cellcolor{bestcell}\checkmark \\
& DS-V4-Flash       & .795 & .475 & .298 & .195 & .009 & .268 & 0.640 & 10.84 & $-$7.40 & 10.72 & \cellcolor{bestcell}\checkmark \\
\midrule
\multirow{6}{*}{\rotatebox[origin=c]{90}{\textbf{Baselines}}}
& EqW               & --- & --- & --- & --- & --- & --- & \cellcolor{bestcell}\textbf{0.710} & \cellcolor{bestcell}\textbf{10.75} & $-$6.16 & 9.36 & --- \\
& 60/40             & --- & --- & --- & --- & --- & --- & 0.592 & 9.56 & $-$5.98 & 9.42 & --- \\
& RiskPar           & --- & --- & --- & --- & --- & --- & $-$0.130 & 3.34 & $-$3.50 & 4.68 & --- \\
& CovRiskPar        & --- & --- & --- & --- & --- & --- & $-$0.914 & 0.28 & $-$4.16 & 3.98 & --- \\
& MinVar            & --- & --- & --- & --- & --- & --- & $-$3.761 & $-$0.82 & \cellcolor{bestcell}\textbf{$-$1.15} & \cellcolor{bestcell}\textbf{1.28} & --- \\
& BL                & --- & --- & --- & --- & --- & --- & $-$0.929 & $-$3.51 & $-$9.62 & 7.82 & --- \\
\bottomrule
\end{tabular}
\end{adjustbox}
\caption{Main pipeline results. \colorbox{bestcell}{Green} = column best. ``Gate'' = pass all three stress scenarios under that profile. LLMs ranked by CEPS within each profile.}
\label{tab:pipeline_main}
\end{table*}

\textbf{Similar allocation scores conceal downstream differences.}
Table~\ref{tab:pipeline_main} ranks GLM-5.1 first and Qwen3.6-Plus second by CEPS under all three profiles. Across the 30 model--profile combinations, S1 ranges from .742 to .807 and S3 from only .264 to .310, while turnover agreement (S4) spans .121--.320 and risk-state agreement (S5) .000--.409. Under the conservative profile, Qwen3.6-Plus outperforms GLM-5.1 at four of five stages, but GLM-5.1's S4 score is nearly twice as high (.304 vs. .153), yielding a slightly higher CEPS (.299 vs. .292). \textbf{Similar allocation scores can therefore mask differences in turnover and risk-state agreement that affect overall pipeline rankings.}

\begin{figure}[t]
    \centering
    \includegraphics[width=\columnwidth]{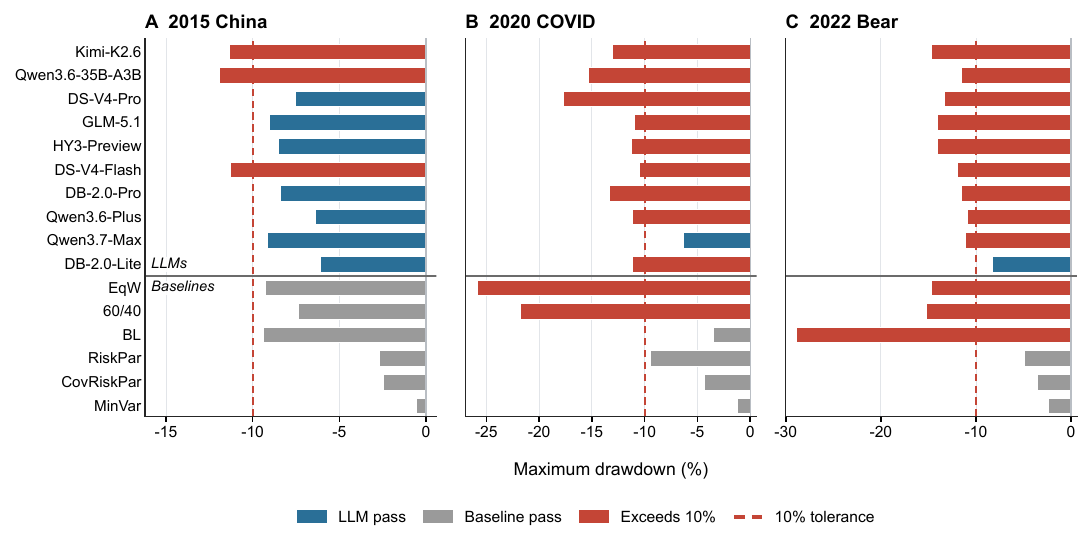}
    \caption{Maximum drawdown under the conservative profile in each stress regime.}
    \label{fig:stress_gate}
\end{figure}

\textbf{Normal-period drawdown rankings can reverse under stress.}
In the normal period, 15 of the 30 model--profile cells outperform EqW on Sharpe, and all 30 remain within their respective drawdown limits. Figure~\ref{fig:stress_gate} shows that all ten models fail at least one conservative-profile scenario; nine pass all three balanced-profile scenarios and all ten pass the aggressive-profile scenarios. Under the balanced profile, DS-V4-Flash moves from the smallest normal-period drawdown (6.56\%) to the largest COVID-window drawdown (22.19\%), becoming the only model to fail that profile's stress gate. Low normal-market drawdown therefore does not establish stress resilience. Details are in Appendix~\ref{app:stress_eval}.

\subsection{QA--Pipeline Rank Inversion}
\label{sec:rank_dissociation}

\textbf{QA and pipeline rankings differ sharply.} GLM-5.1 moves from seventh in QA to first in balanced-profile CEPS, Kimi-K2.6 from last to fourth, and DS-V4-Flash from tied first to tenth ($\rho=-.32$). Direct QA--Sharpe correlations across ten models range from $-.67$ to $+.83$ over the twelve period--profile settings. Positive associations appear in conservative and balanced 2015--16 settings, while the other ten intervals include zero (Appendix~\ref{app:qa_financial}). The observed QA rankings do not consistently carry over to portfolio outcomes.

Dynamic allocation requires combining market interpretation, signals, and profile constraints over the portfolio. T4/T5 test allocation with supplied statistics. Appendix~\ref{app:qa_details} examines task scores through information ablations and supplementary structured decisions. \textbf{Dynamic evaluation tests how local answers are used in portfolio construction, which standalone QA does not establish.}

\subsection{Cross-Period Evaluation}
\label{sec:cross_regime}

\begin{figure*}[t]
    \centering
    \includegraphics[width=\textwidth]{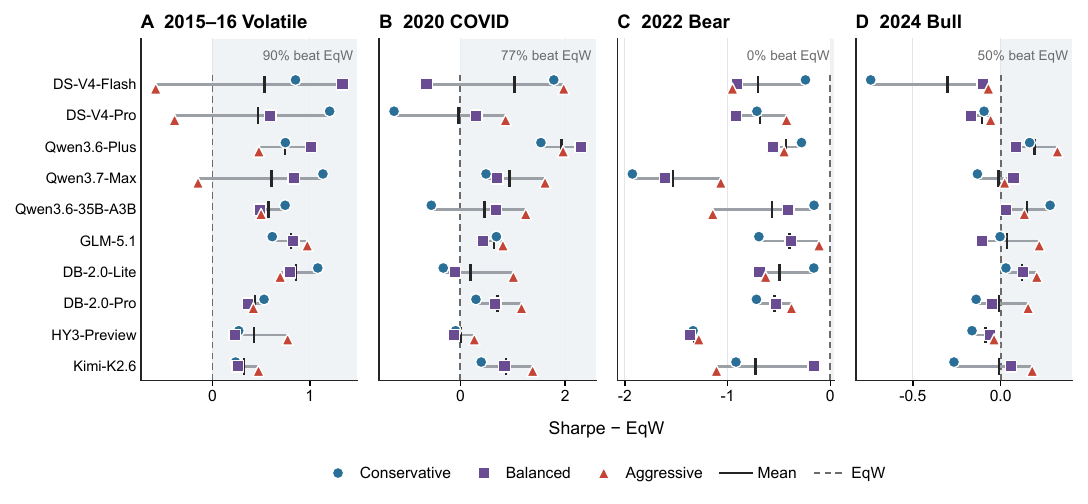}
    \caption{Sharpe relative to EqW across four market periods. Light shading marks $\Delta{>}0$ (beat EqW). Panel annotations report the fraction of model--profile pairs that beat EqW in each period; (d) uses a narrower axis scale.}
    \label{fig:cross_period}
\end{figure*}

To probe performance under diverse market conditions, we reuse the three stress windows and report period Sharpe under three investor profiles: 2015--16 (volatile), 2020 (COVID crash and rebound), and 2022 (bear), alongside 2024 (bull). Across the 120 evaluated model--profile--period cases, LLMs beat EqW in 65 cases (54.2\%). Figure~\ref{fig:cross_period} plots Sharpe minus EqW for every model--profile pair. Beat rates vary sharply: 77\% in the 2020 crash-and-rebound window, 90\% in 2015--16, 50\% in the 2024 bull market, and 0\% in the prolonged 2022 bear market. EqW Sharpe is -0.602 in 2020, where the mean model gap is $+0.68$, but reaches 0.710 in 2024; in 2022, every model--profile pair trails EqW, with mean $\Delta$Sharpe $-0.74$. \textbf{LLM portfolios show no consistent Sharpe advantage over EqW across these four windows.} Across fixed model--profile pairs, the mean range of $\Delta$Sharpe over the four periods is 1.84, compared with .71 across profiles within the same model and period, showing that relative performance varies more across these market-window and decision-frequency settings than across investor profiles. Appendices~\ref{app:cross_regime} and~\ref{app:nav_trajectories} report results and NAV trajectories.

\subsection{Real-Time Evaluation}
\label{sec:live_eval}

Evaluating historical markets risks leakage from prices, headlines, or outcomes seen during LLM pretraining~\citep{time-travel-is-cheating-neurips25,stockbench-arxiv25}. \portbench therefore supports real-time evaluation with point-in-time inputs: at each decision date, the model receives only data available through the previous close and runs the five-stage pipeline. Over 12 trading days (2026-07-16--07-31), we collect 72 decisions from six models with daily rebalancing. Mean S1 scores range from .78 to .82, while S4 turnover agreement remains low at .047--.163. This initial run demonstrates stage-level evaluation on newly arriving data and more frequent portfolio decisions. Appendix~\ref{app:live_eval} reports daily stage scores and lookback CEPS.

\section{Deep Analysis}
\label{sec:analysis}

\subsection{Local Gains Can Fade or Reverse}
\label{sec:analysis_composition}

Paired interventions under the balanced profile test how replacing one stage with its reference affects later decisions. In this auxiliary diagnostic, S4/S5 score the model's execution and risk-management plans. Replacing S1 with its reference output raises S2 by $+.626$ in normal episodes; both references follow the same trailing-return rule. Its mean effect on S3 is only $+.017$ and is negative in 40 of 110 episodes. Replacing S2 raises S3 by only $+.023$, with 39 negative cases. Under stress, S2 and S3 replacement reduce S5 by $-.059$ and $-.056$, although most episode-level S5 changes are zero. \textbf{Large gains at the next stage become much smaller or turn negative at later stages.} Case traces reveal execution and risk changes missed by stage scores; see Appendix~\ref{app:paired_details} for details.

\subsection{Broad Holdings Conceal Correlated Risk}
\label{sec:analysis_baselines}

The 330 saved 2024-normal allocations hold 73 assets on average, yet still concentrate weight in correlated assets. For each decision, we construct a matched allocation by equalizing weights among held assets within each class while preserving asset selection, class weights, and uninvested cash. This isolates the risk effect of within-class weights.

\begin{figure*}[t]
\centering
\includegraphics[width=\textwidth]{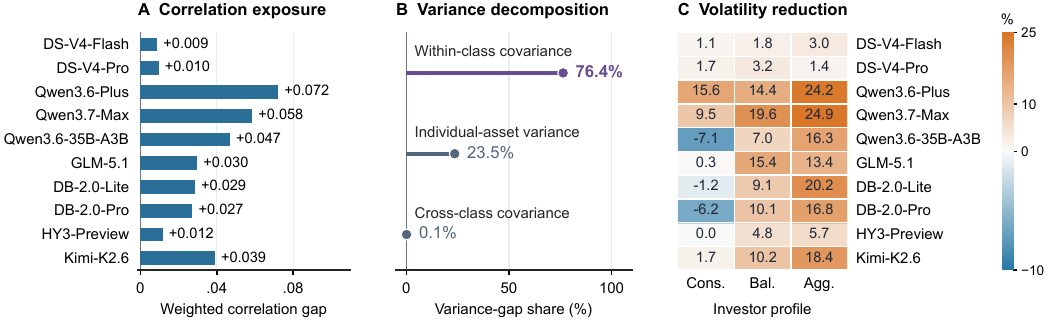}
\caption{Correlation risk at fixed holdings and asset-class weights. \textbf{A:} Mean difference in volatility-weighted pairwise correlation. \textbf{B:} Contributions to the mean portfolio-variance difference across all 330 decisions; percentages are shares of this difference. Differences in A and B are model minus within-class equal weighting over the same held assets. \textbf{C:} Mean relative volatility reduction (\%) from equalization; negative values indicate increased volatility.}
\label{fig:correlation_risk}
\end{figure*}

Across ten models, three profiles, and eleven decision dates, this equalization lowers volatility estimated from lookback data by 8.5\% on average, with reductions in 271/330 decisions. Figure~\ref{fig:correlation_risk}A shows higher volatility-weighted pairwise correlation under model weights for all models. Panel B attributes 76.4\% of the mean variance difference to within-class covariance, versus 23.5\% to individual-asset variance terms. Panel C shows larger mean volatility reductions under more aggressive profiles: 1.6\%, 9.6\%, and 14.4\% for conservative, balanced, and aggressive allocations. \textbf{Diversification depends on how capital is distributed among correlated holdings, even when asset selections and class exposures are unchanged.} Appendix~\ref{app:correlation_risk} gives the covariance decomposition, model--profile results, and sensitivity analyses.

\begin{figure*}[!t]
\centering
\includegraphics[width=\textwidth]{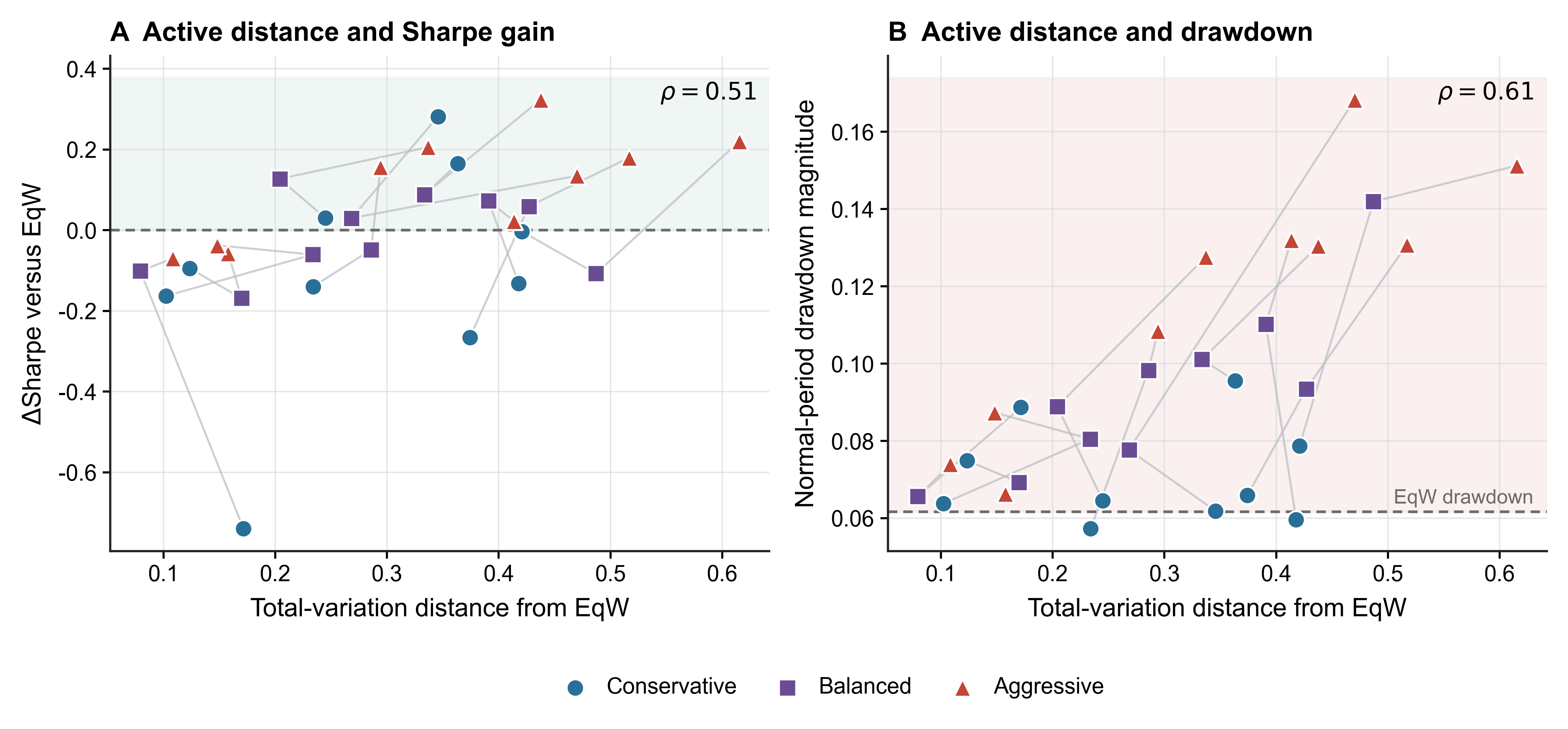}
\caption{Active-allocation diagnostics across 30 normal-period model--profile cells, with gray paths connecting the three profiles of each model. \textbf{A:} Total-variation distance from EqW versus $\Delta$Sharpe. \textbf{B:} The same distance versus drawdown magnitude. Shaded regions mark $\Delta$Sharpe $>0$ in A and drawdowns larger than EqW in B.}
\label{fig:composed_decision_gap}
\end{figure*}

\paragraph{Active allocation and realized risk.}
Across the 30 cells, the mean total-variation distance is 0.306 from EqW and 0.890 from the profile-constrained reference. Figure~\ref{fig:composed_decision_gap} relates distance from EqW to realized outcomes. Larger deviations from EqW are associated with higher volatility ($\rho=.65$), drawdown magnitude ($\rho=.61$), and $\Delta$Sharpe ($\rho=.51$) in the normal period. The models therefore take meaningful active positions, with higher relative Sharpe accompanied by greater volatility and drawdowns~\mbox{\citep{demiguel2009optimal}}. Moving from conservative to aggressive increases distance from EqW by .070 on matched allocations.

\paragraph{Allocation-to-execution gap.}
Under the balanced profile, every model produces less turnover than the reference; overall, the mean model-to-reference turnover ratio is .203. Reference allocations hold 8--14 securities, with maximum weights of .24--.72, whereas model allocations spread weight more broadly. Total-variation distance from the reference averages .890, giving mean S3 reference-weight similarity of .110. Mean model HHI is .0207, whose reciprocal is about 48 effective positions. The allocations differ from the reference in both concentration and turnover, explaining why partial credit for class exposure can coexist with low S4 scores. Appendix~\ref{app:active_tilt} reports the diagnostics; Appendix~\ref{app:showcase_pipeline} gives full traces.

\subsection{Performance Reverses across Regimes}
\label{sec:analysis_stress_necessity}

\textbf{In 2020, 23 of 30 cases beat EqW, whereas none do so in 2022.} Both windows use weekly decisions, contrasting a sharp crash and recovery with a prolonged decline. EqW's Sharpe is similar in these windows ($-.602$ versus $-.540$), whereas mean model Sharpe falls from $.074$ to $-1.281$. The reversal therefore reflects changes in model-portfolio performance, rather than simply a stronger benchmark. \textbf{The mean relative Sharpe advantage is largest in the crash-and-rebound window.} The mean Sharpe advantage of aggressive over conservative profiles is largest in 2020 at $+.935$, compared with $-.039$ in 2022 and $+.214$ in 2024 (Appendix~\ref{app:cross_regime}); details are in Appendix~\ref{app:process_outcome}.

\subsection{Risk Profiles Shift Exposure}
\label{sec:analysis_profiles}

On matched model--date pairs, shifting from conservative to aggressive raises risk-asset exposure by .175 and lowers bond-plus-cash exposure by .162. Normal-period volatility rises by .050 for nine of ten models and drawdown magnitude by .047 for eight. Mean Sharpe increases by .214, with nine improving and one declining. Under stress, the shift raises volatility for all ten in 2020 and 2022. Yet total return rises for all ten in the 2020 rebound and falls for nine in the 2022 decline. In 2022, five models achieve higher Sharpe under the aggressive profile, yet four have lower returns and deeper drawdowns. With negative excess returns, higher volatility can make Sharpe less negative, highlighting the need to assess returns and drawdowns alongside Sharpe. \textbf{Profiles change exposure, risk, and normal-period returns, while the return advantage varies across periods.} Figure~\ref{fig:correlation_risk}C shows more scope to reduce estimated volatility in aggressive portfolios through within-class reweighting without changing asset selections or class exposures. Appendix~\ref{app:profile_allocations} reports matched changes.

\section{Related Work}
\label{sec:related}

Financial LLM benchmarks have progressed from knowledge retrieval~\citep{finqa-emnlp21, convfinqa-emnlp22, pixiu-neurips23} to investment decision-making~\citep{finben-neurips24, xfinbench-acl25, finmme-acl25}, yet portfolio management evaluations remain limited to static QA, single-asset-class backtests, or asset-by-asset backtests~\citep{finrl-neurips22, cryptotrade-emnlp24, stockbench-arxiv25, investorbench-acl25}. StockBench~\citep{stockbench-arxiv25} introduces process-level analysis but lacks cross-asset correlation scoring and investor-profile adaptation; LLM agents for PM~\citep{Fincon-neurips24, MASS-arxiv25, hedgeagents-www25} rely on proprietary backtests that assess only terminal outcomes~\citep{bench-prioritize-risk-arxiv25, finsaber-kdd26}. Despite robust portfolio construction relying on covariance structures~\citep{portfolio-selection-jof, risk-parity-panagora} and non-LLM methods exploiting them~\citep{smartfolio-ijcai25}, cross-asset correlation risk remains underexamined in LLM portfolio evaluation~\citep{risk-first-eval-icml25}. \portbench examines how multi-asset allocation decisions translate into portfolio risk across market regimes and investor profiles. Full discussion is in Appendix~\ref{app:related_work}.

\section{Conclusion}
\label{sec:conclusion}

We presented \portbench, a benchmark for LLM-driven portfolio management with market data spanning six asset classes from 2015 to 2025. \portbench provides a multi-asset market dataset of 183 market series with prices, macro series, and time-aligned news, alongside a dual-layer evaluation framework combining static PM QA with a five-stage evaluation of historical portfolios. The framework uses a class-level correlation metric, asset-level covariance diagnostics, and \ceps, and supports stress-window, investor-profile, and real-time evaluation. Despite high financial QA scores, the evaluated LLMs fail to beat equal weighting on Sharpe in 55 of 120 evaluations across four historical periods. Their portfolios hold many assets, yet within-class weights concentrate exposure in correlated holdings, limiting diversification. These findings expose a gap between financial QA and portfolio performance: LLMs concentrate correlated exposures and show no consistent Sharpe advantage over equal weighting across these periods.
\section*{Limitations}
\label{sec:limitations}

Although \portbench provides a comprehensive evaluation framework, several limitations remain: (1) The sandbox environment relies on deterministic transaction costs applied to historical price data, so it does not capture real-world microstructure dynamics, liquidity constraints, and order-flow impact. Integrating generative market simulators like MarS~\citep{mars-iclr25} to model tokenized order flow and support shock injection would provide more realistic execution feedback. (2) Historical evaluation uses monthly decisions in the normal period and weekly decisions in three selected stress windows. The daily real-time window covers twelve trading days (Appendix~\ref{app:live_eval}); longer real-time evaluation would help test whether the results hold across periods. (3) The model produces S1--S3 through a fixed set of isolated LLM prompts without persistent memory, external tool access, or multi-agent coordination; the sandbox then computes the S4/S5 execution and risk scores used in the ranking from the S3 allocation~\citep{tradingagents-aaai25,hedgeagents-www25}. Future work should adopt agentic methods that incorporate tool calling, long-horizon memory, and inter-agent communication to improve portfolio management.

\section*{Ethical Statement}
\label{sec:ethics}

\portbench is designed as a research benchmark for evaluating LLM capabilities in portfolio management and is not intended as financial advice or as a decision-support tool for real investment. All evaluations use publicly available historical market data; no proprietary, private, or personally identifiable information is used. The benchmark does not involve human subjects, and no crowd-sourced annotations were collected. We caution against deploying LLM-generated portfolio allocations in live trading without rigorous human oversight. As our experiments demonstrate, the evaluated models fail to consistently outperform simple heuristic baselines and exhibit fragile behavior under stress conditions. The benchmark's stress-test evaluation is specifically designed to surface such failure modes before deployment, but passing the stress gate does not certify safety in real-world use.

\section*{LLM Statement}
We used LLM-based tools to polish the writing and refine the language of this paper.

\bibliography{reference}

\clearpage
\appendix
\onecolumn
\addtocontents{toc}{\protect\setcounter{tocdepth}{2}}
\renewcommand{\contentsname}{Appendix Contents}
\tableofcontents
\clearpage
\twocolumn

\section{Additional Related Work}
\label{app:related_work}

\subsection{Financial LLM Benchmarks}
\label{app:related_benchmarks}

Financial LLM benchmarks span knowledge retrieval, numerical reasoning, and quantitative tasks~\citep{finqa-emnlp21,convfinqa-emnlp22,flue-flang-emnlp22,financebench-arxiv23,pixiu-neurips23,financereasoning-acl25,findabench-coling25,finben-neurips24,xfinbench-acl25,finmme-acl25,mme-finance-mm25}. Most QA benchmarks stop before allocation under realized markets. Dynamic evaluations commonly restrict the asset universe to equities or cryptocurrencies~\citep{finrl-neurips22,finben-neurips24,cryptotrade-emnlp24,stockbench-arxiv25,democratizing-alpha-icaif25}, or score products independently instead of a joint portfolio~\citep{investorbench-acl25}. StockBench~\citep{stockbench-arxiv25} evaluates multiple trading steps for 20 DJIA equities, but omits cross-asset allocation and investor profiles. Table~\ref{tab:positioning} compares \portbench on financial QA, joint multi-asset allocation, linked decisions, realized risk, and investor constraints.
\subsection{LLMs in Financial Decision-Making}
\label{app:related_agents}

LLM agents now support market analysis, signal generation, and portfolio construction~\citep{finagent-kdd24,tradingagents-aaai25,agents-web26,agent-quant-neurips26,agents-for-investment-management-icaif25}. FinCon~\citep{Fincon-neurips24} uses a manager--analyst hierarchy, MASS~\citep{MASS-arxiv25} scales multi-agent portfolio simulation, HedgeAgents~\citep{hedgeagents-www25} assigns hedging specialists across asset classes, and a cryptocurrency framework~\citep{mas-crypto-arxiv25} coordinates agents over the top-30 cryptocurrencies. These systems are evaluated in separate backtesting settings. \portbench supplies one six-class market interface and scores each stage of a shared five-stage process, enabling stage-level comparison alongside portfolio outcomes.

\subsection{Portfolio Risk Evaluation}
\label{app:related_portfolio}

Portfolio theory makes covariance central to allocation: modern portfolio theory~\citep{portfolio-selection-jof} and risk parity~\citep{risk-parity-panagora} optimize portfolio-level variance or risk contributions. Data-driven methods also use cross-asset co-movement and temporal structure~\citep{mst-eurphysb,deeptrader-aaai21,frequant-kdd24}; SmartFolio~\citep{smartfolio-ijcai25}, for example, penalizes positive intra-class correlation and rewards inter-class hedging. Risk-first evaluation further shows why average returns alone miss failures under market stress~\citep{bench-prioritize-risk-arxiv25,risk-first-eval-icml25}. \portbench addresses both requirements with correlation-aware allocation scoring, three stress windows, and three investor profiles.

\begin{table*}[t]
\centering
\begin{adjustbox}{max width=\textwidth}
\begin{tabular}{lcccccc}
\toprule
\textbf{Benchmark} & \textbf{Multi-Asset} & \textbf{Fin.\ QA} & \textbf{Alloc.\ Quality} & \textbf{Seq.\ Process} & \textbf{Risk/Stress} & \textbf{Profile Align.} \\
\midrule
FinQA~\citep{finqa-emnlp21}                      &            & $\checkmark$ &            &            &            &            \\
ConvFinQA~\citep{convfinqa-emnlp22}              &            & $\checkmark$ &            &            &            &            \\
PIXIU~\citep{pixiu-neurips23}                    &            & $\checkmark$ &            &            &            &            \\
FinanceBench~\citep{financebench-arxiv23}        &            & $\checkmark$ &            &            &            &            \\
CryptoTrade~\citep{cryptotrade-emnlp24}          &            &              & $\circ$    & $\circ$    & $\checkmark$ &            \\
FinBen~\citep{finben-neurips24}                  & $\circ$    & $\checkmark$ & $\circ$    & $\circ$    & $\circ$    &            \\
FinEval~\citep{fineval-naacl25}                  &            & $\checkmark$ &            & $\circ$    &            & $\circ$    \\
XFinBench~\citep{xfinbench-acl25}               &            & $\checkmark$ &            &            & $\circ$    &            \\
InvestorBench~\citep{investorbench-acl25}        & $\circ$    &              & $\circ$    & $\circ$    & $\checkmark$ & $\circ$    \\
FinanceReasoning~\citep{financereasoning-acl25}  &            & $\checkmark$ &            &            &            &            \\
FinDABench~\citep{findabench-coling25}           &            &              &            &            & $\circ$    &            \\
StockBench~\citep{stockbench-arxiv25}            & $\circ$    &              & $\circ$    & $\checkmark$ & $\checkmark$ &            \\
\midrule
\textbf{\portbench (Ours)}                       & $\checkmark$ & $\checkmark$ & $\checkmark$ & $\checkmark$ & $\checkmark$ & $\checkmark$ \\
\bottomrule
\end{tabular}
\end{adjustbox}
\caption{Comparison of representative financial LLM benchmarks. Multi-Asset covers joint allocation across asset classes; Fin.\ QA tests financial knowledge or reasoning; Alloc.\ Quality scores portfolio weights or their outcomes; Seq.\ Process evaluates multiple linked decisions; Risk/Stress evaluates realized risk or stress periods; Profile Align.\ conditions decisions on investor constraints. $\checkmark$ denotes an explicit evaluation dimension, $\circ$ denotes a restricted setting or indirect measure, and a blank denotes no reported evaluation.}
\label{tab:positioning}
\end{table*}
\section{Dataset and Market Context}
\label{app:market_base_dataset}

\begin{figure}[!h]
\centering
\includegraphics[width=\columnwidth]{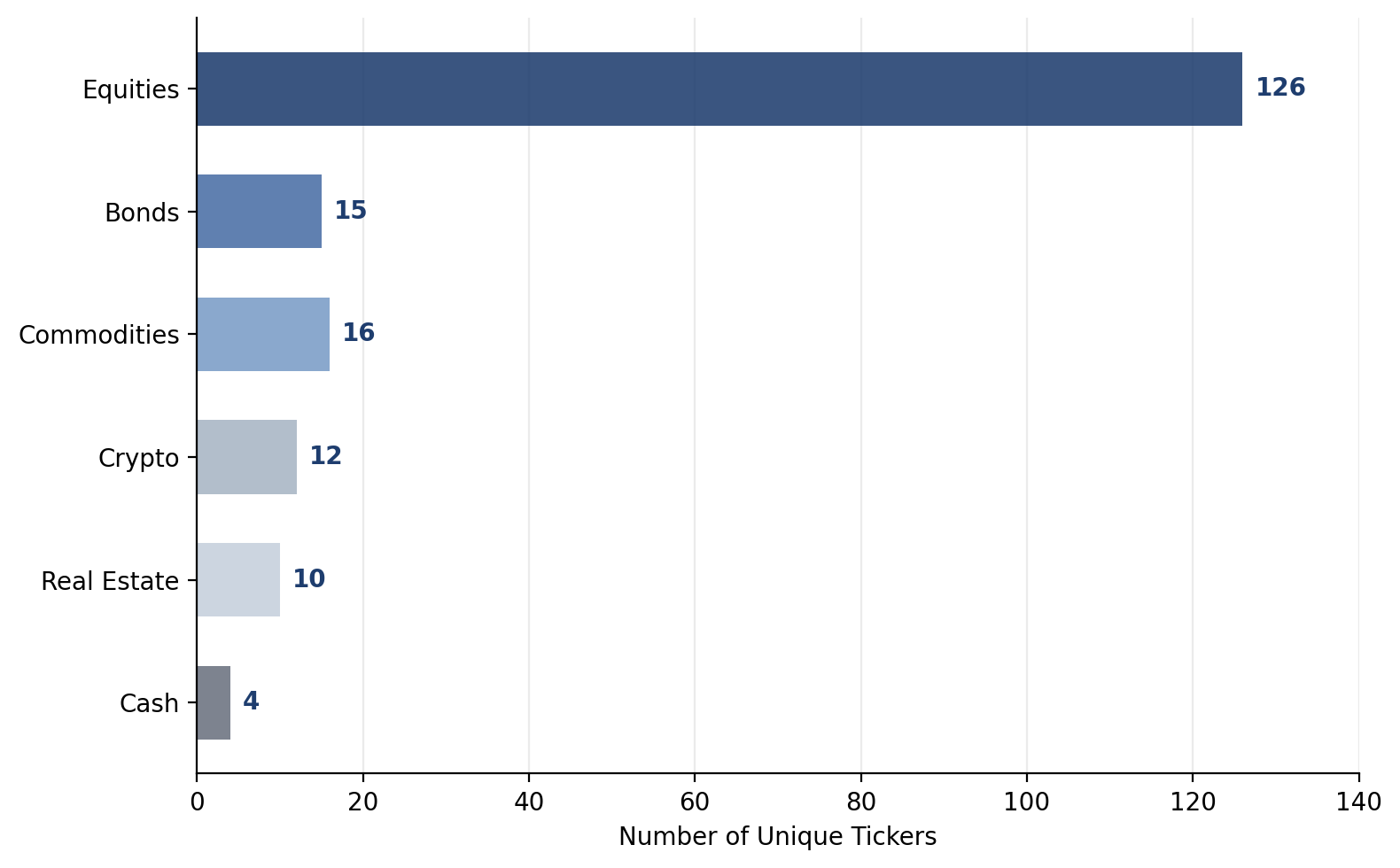}
\caption{Number of unique instruments or series per asset class in \portbench.}
\label{fig:app_asset_class_tickers}
\end{figure}

\portbench covers 183 market series spanning 2015--2025 across six asset classes, collected from Yahoo Finance (price/return series), FRED (macroeconomic indicators), and Kaggle (supplementary equity and cryptocurrency series). Figure~\ref{fig:app_asset_class_tickers} shows instrument counts by asset class. Equities exhibit the broadest coverage (126 series), including individual stocks, broad-market, sector, and factor ETFs, and index context. Commodities (16 tickers) and bonds (15 series) provide representative cross-class hedging opportunities, cryptocurrency (12 tickers) captures major and mid-cap digital assets, and real estate (10 series) and cash equivalents (4 series) extend the market coverage.

\paragraph{Code and source data.}
The benchmark implementation is licensed under Apache~2.0. Market data are obtained from the providers listed above, with reuse subject to the respective source terms. Historical allocation results use the evaluation subset defined in Appendix~\ref{app:evaluation_details}.

\paragraph{Market context at decision time.}
At each decision date, the model receives a point-in-time market context containing a 60-trading-day lookback window of price history and daily returns for all assets in scope; macro indicators (Fed funds rate, VIX, yield curve slope); an intra-class correlation matrix for each asset class and an inter-class correlation matrix, both recomputed from the lookback window at each decision date; any available news text or earnings filings preceding the decision date; the current portfolio weights; and the current portfolio NAV. The intra- and inter-class correlation matrices are formatted as structured tables and injected directly into the S1 and S3 prompts, giving models explicit access to the correlation information required for correlation-aware allocation. The correlation interface is restricted to the asset classes present in each snapshot.

\subsection{Price Trajectories}
\label{app:dataset_overview}

\begin{figure*}[!ht]
\centering
\begin{subfigure}[b]{0.48\textwidth}
\includegraphics[width=\textwidth]{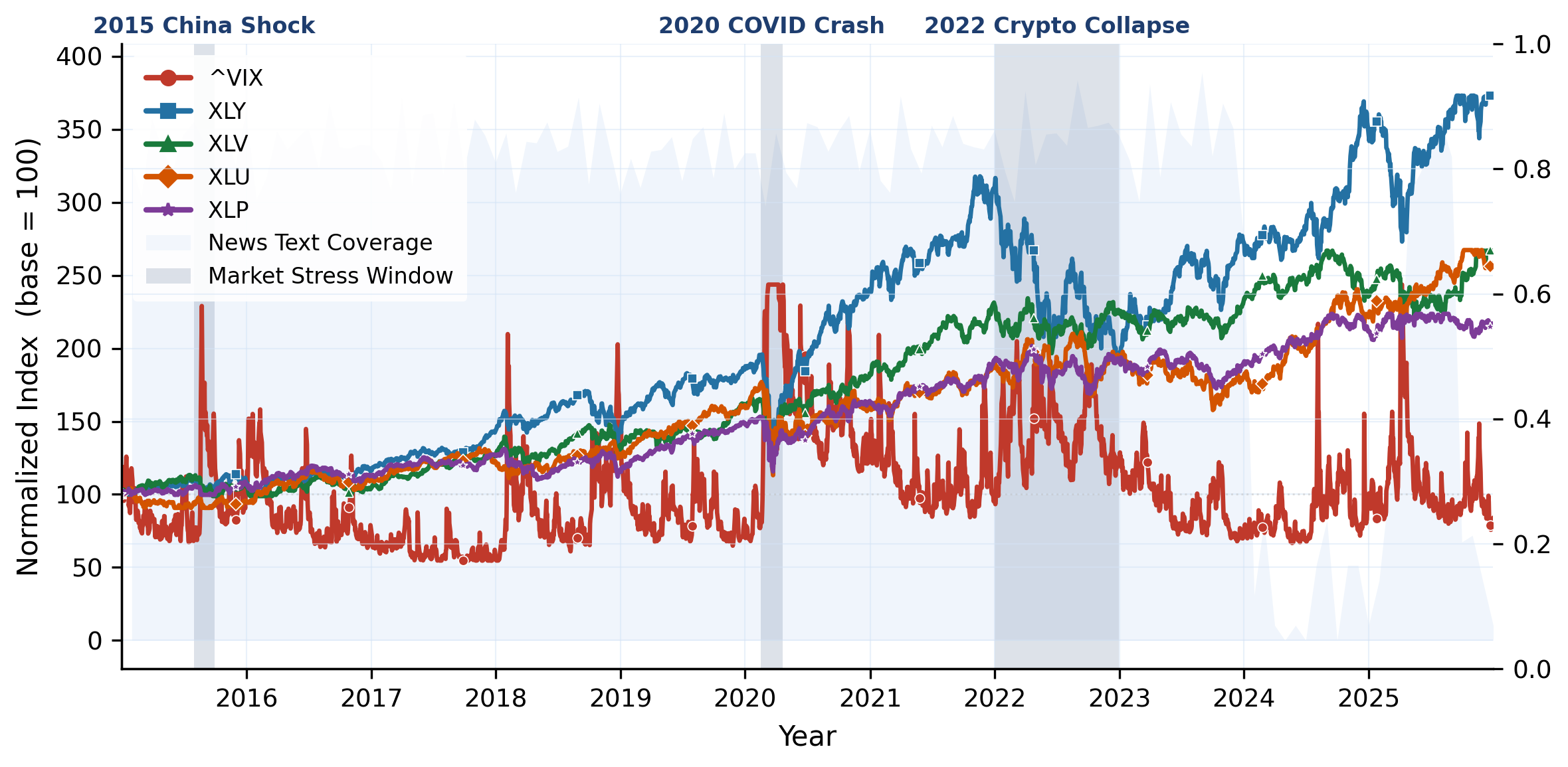}
\caption{Equities (representative)}
\label{fig:dataset_overview_equities}
\end{subfigure}
\hfill
\begin{subfigure}[b]{0.48\textwidth}
\includegraphics[width=\textwidth]{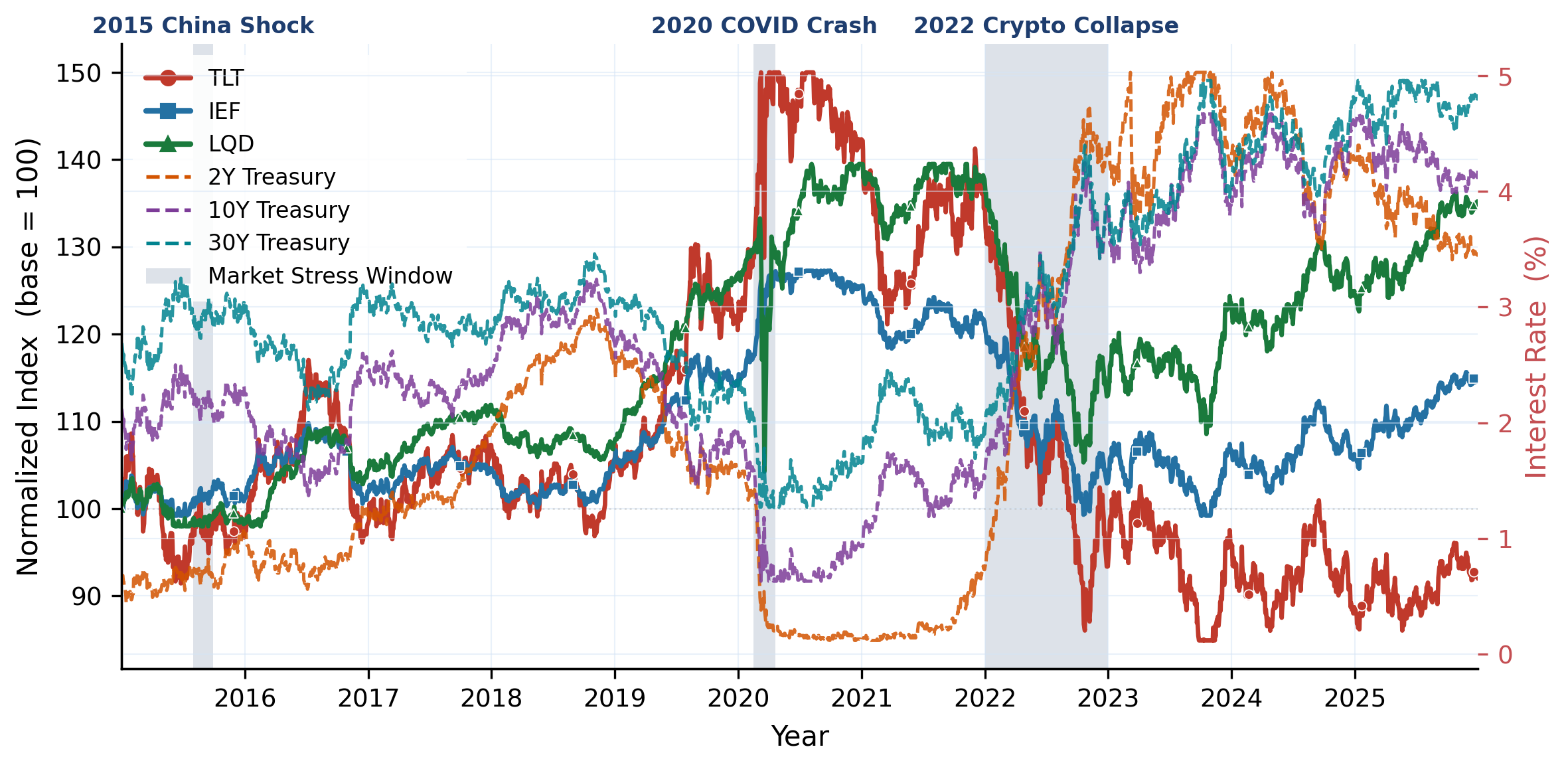}
\caption{Bonds (representative)}
\label{fig:dataset_overview_bonds}
\end{subfigure}

\vspace{4pt}

\begin{subfigure}[b]{0.48\textwidth}
\includegraphics[width=\textwidth]{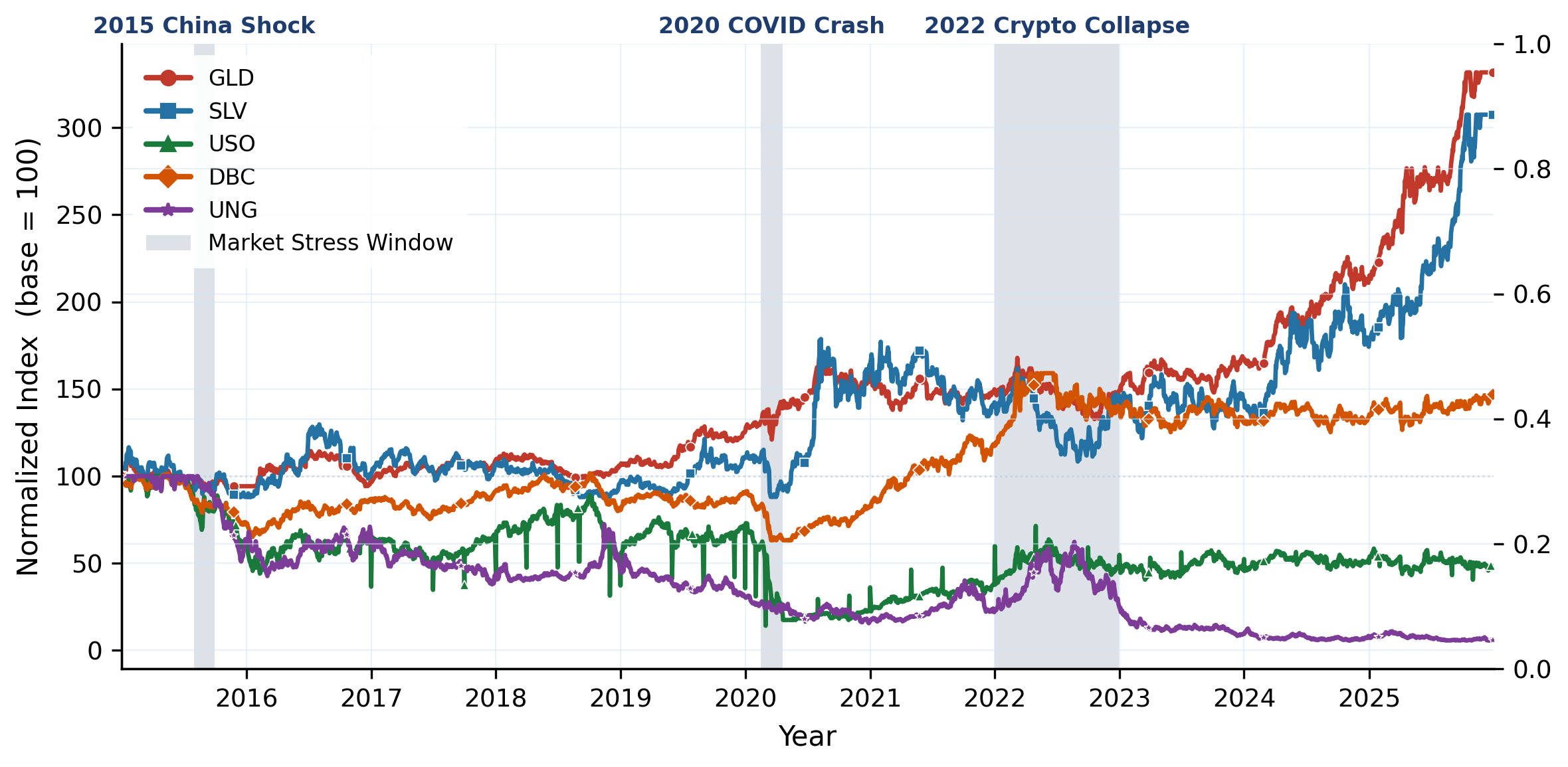}
\caption{Commodities (representative)}
\label{fig:dataset_overview_commodities}
\end{subfigure}
\hfill
\begin{subfigure}[b]{0.48\textwidth}
\includegraphics[width=\textwidth]{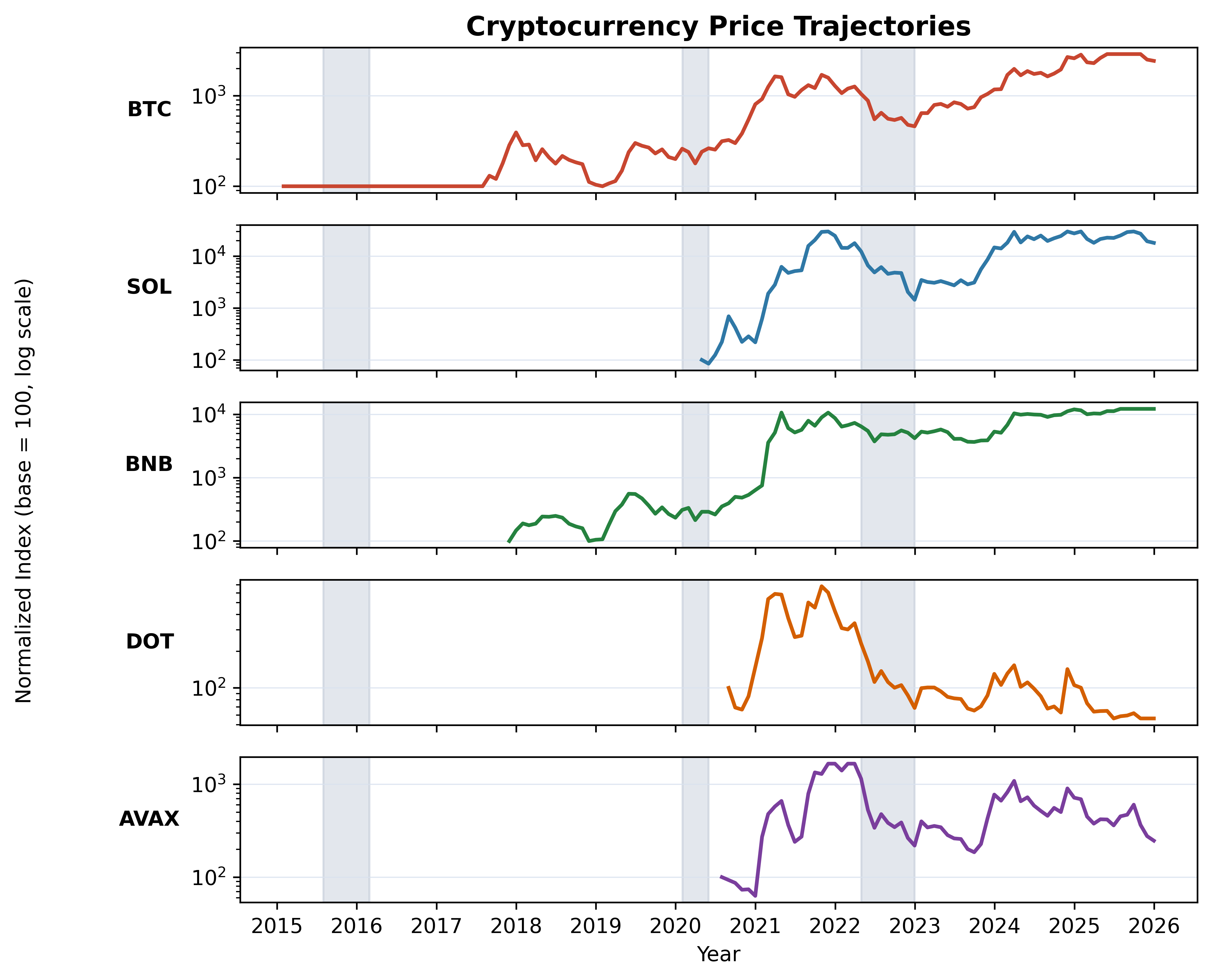}
\caption{Cryptocurrency (representative)}
\label{fig:dataset_overview_crypto}
\end{subfigure}

\vspace{4pt}

\begin{subfigure}[b]{0.48\textwidth}
\includegraphics[width=\textwidth]{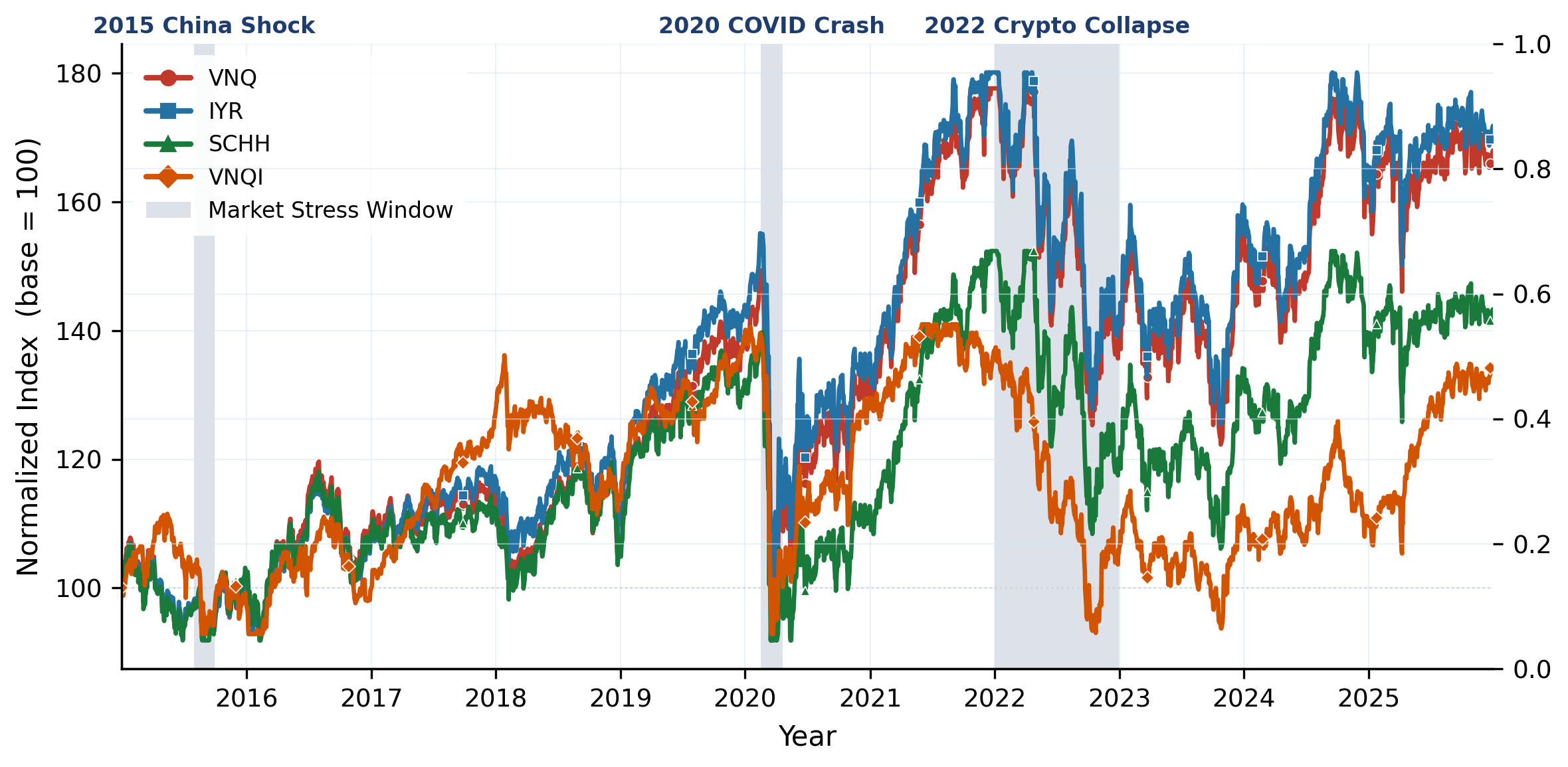}
\caption{Real Estate (representative)}
\label{fig:dataset_overview_re}
\end{subfigure}
\hfill
\begin{subfigure}[b]{0.48\textwidth}
\includegraphics[width=\textwidth]{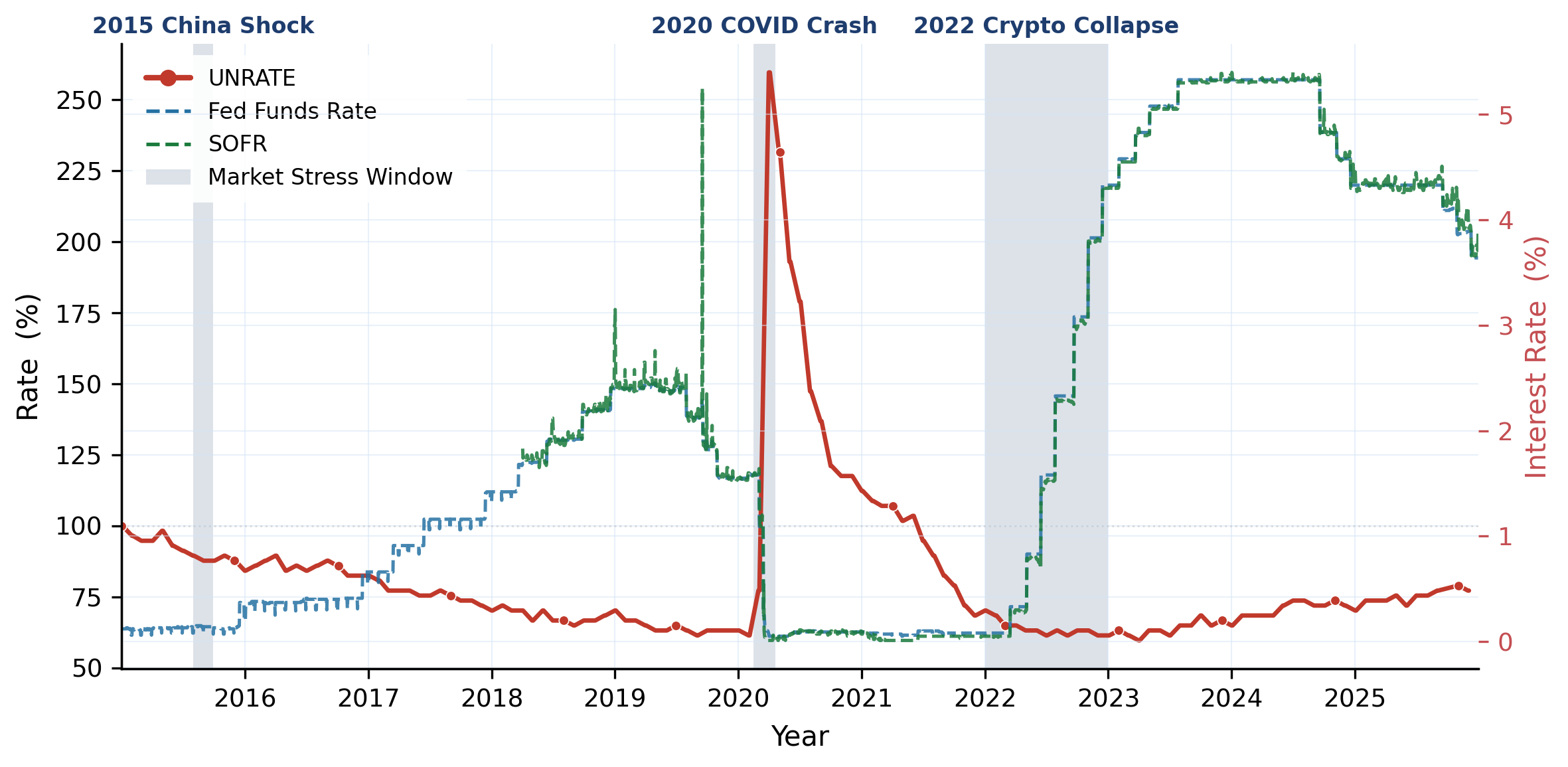}
\caption{Cash-related macro series (representative)}
\label{fig:dataset_overview_cash}
\end{subfigure}

\caption{Representative market-series trajectories from each asset class. Price series are indexed to 100 at their first available observation; the cryptocurrency panel uses a logarithmic index axis. Interest-rate series retain percentage units on secondary axes.}
\label{fig:dataset_overview_b}
\end{figure*}

Figure~\ref{fig:dataset_overview_b} displays representative market series from the six asset classes over 2015--2025. Coverage ranges from 126 equity instruments to four cash-related series. Within-class paths also differ: natural gas and crude oil fluctuate more sharply than gold, while BTC, SOL, BNB, DOT, and AVAX follow distinct cryptocurrency cycles and listing histories. These series differ in history length, volatility, and drawdown patterns.
\textbf{The dataset combines long-history traditional assets with later-listed cryptocurrencies, so history lengths differ across assets in the dataset.}

\subsection{Data Preprocessing}
\label{app:preprocessing}

\paragraph{Calendar alignment.}
All price and return series are aligned to a common business-day calendar.
For realized portfolio accounting, short price gaps of up to five consecutive trading days are forward-filled before computing returns; daily returns are not carried forward.
Gaps exceeding five days are retained as missing values and excluded from correlation estimation using pairwise complete observations, so that assets with non-overlapping listing histories (particularly cryptocurrency) do not reduce the effective sample for other asset pairs.

\paragraph{Market regime labeling.}
Each asset class is assigned one of four market regime labels, bull, bear, sideways, or crisis, on a rolling basis.
A crisis window begins when the maximum drawdown from the trailing 252-trading-day peak exceeds 15\%.
Bull and bear periods are identified using a dual moving-average crossover rule (50-day and 200-day); periods where neither condition is satisfied are labeled sideways.
Regime labels group QA samples and evaluation results by market regime.

\paragraph{Data splits.}
The dataset is divided into three non-overlapping splits with year-end boundaries:
\begin{itemize}
  \item \textbf{Train}: 2015--2022 (eight years; used for correlation matrix estimation and QA generation)
  \item \textbf{Validation}: 2023 (one year; used for hyperparameter selection and QA validation)
  \item \textbf{Test}: 2024--2025 (two years; held out for reported QA evaluation)
\end{itemize}
Pipeline evaluation uses calendar windows that sit inside this coverage but serve a different role: the primary normal-period backtest is the 2024 bull window, with 2015--16, 2020, and 2022 reserved for stress-gate and cross-period Sharpe comparisons (Section~\ref{sec:experiments}). These pipeline windows are therefore evaluation regimes, not a redefinition of the QA train/val/test split.

\paragraph{Correlation matrix estimation.}
Figures~\ref{fig:correlation_heatmap} and~\ref{fig:inter_class_correlation} use a dataset-level Pearson matrix computed from 2015--2022 daily returns with pairwise complete observations. Each pipeline episode recomputes its correlation matrix from the trailing 60 trading days available at the decision date. The profile-constrained S3 reference, CovRiskPar, MinVar, and Black--Litterman use covariance estimated from the same decision-time snapshot; covariance-based baselines apply 252-day annualization. Pearson correlation maps directly to the intra- and inter-class scoring formulas in Appendix~\ref{app:metrics}.

\subsection{Correlation Structure}
\label{app:correlation}

\begin{figure*}[!ht]
\centering
\includegraphics[width=\textwidth]{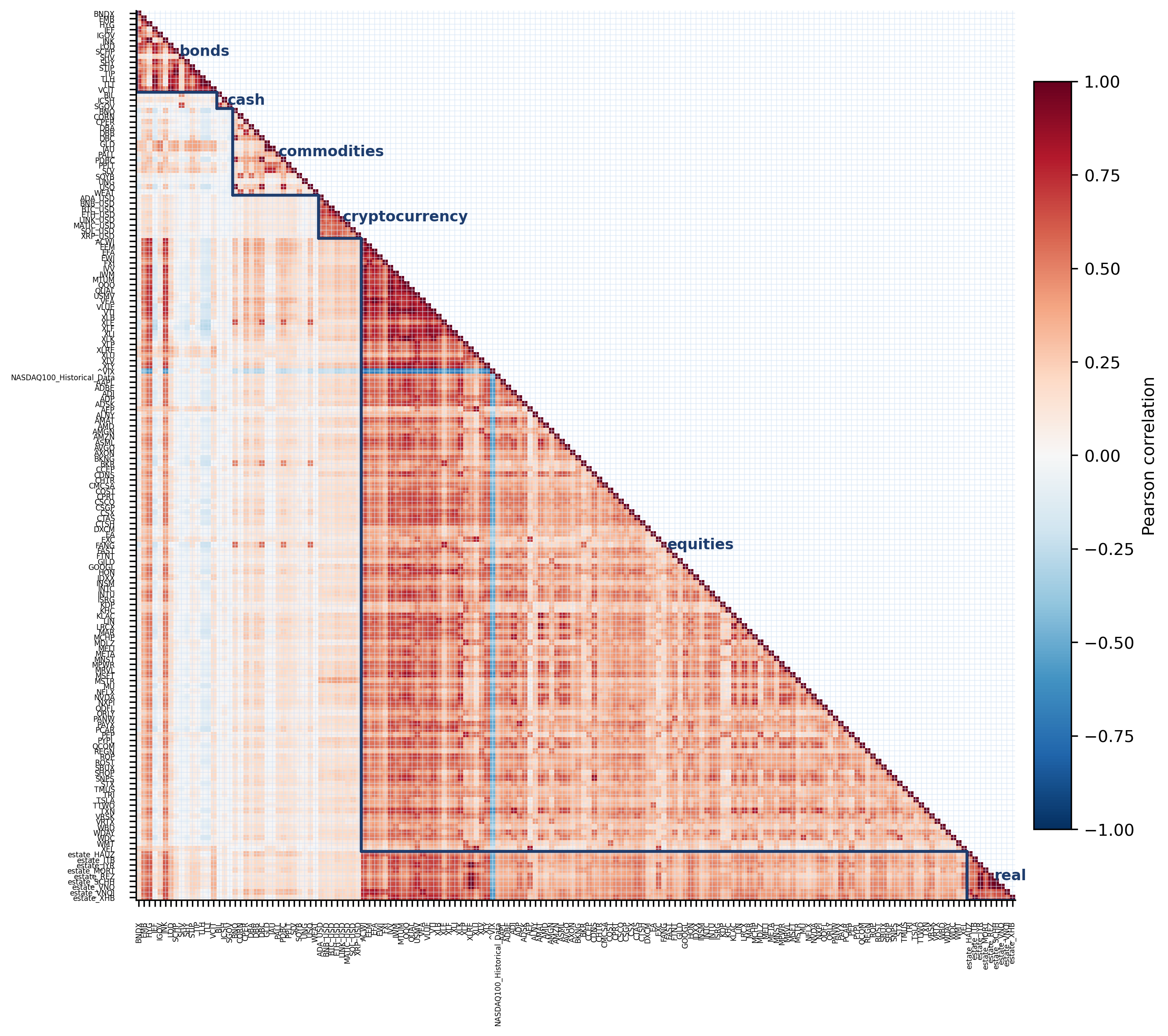}
\caption{Pairwise Pearson correlation matrix across all six asset classes, computed from daily returns over the full training period (2015--2022). Rows and columns are ordered by asset class.}
\label{fig:correlation_heatmap}
\end{figure*}

\begin{figure*}[!ht]
\centering
\includegraphics[width=\textwidth]{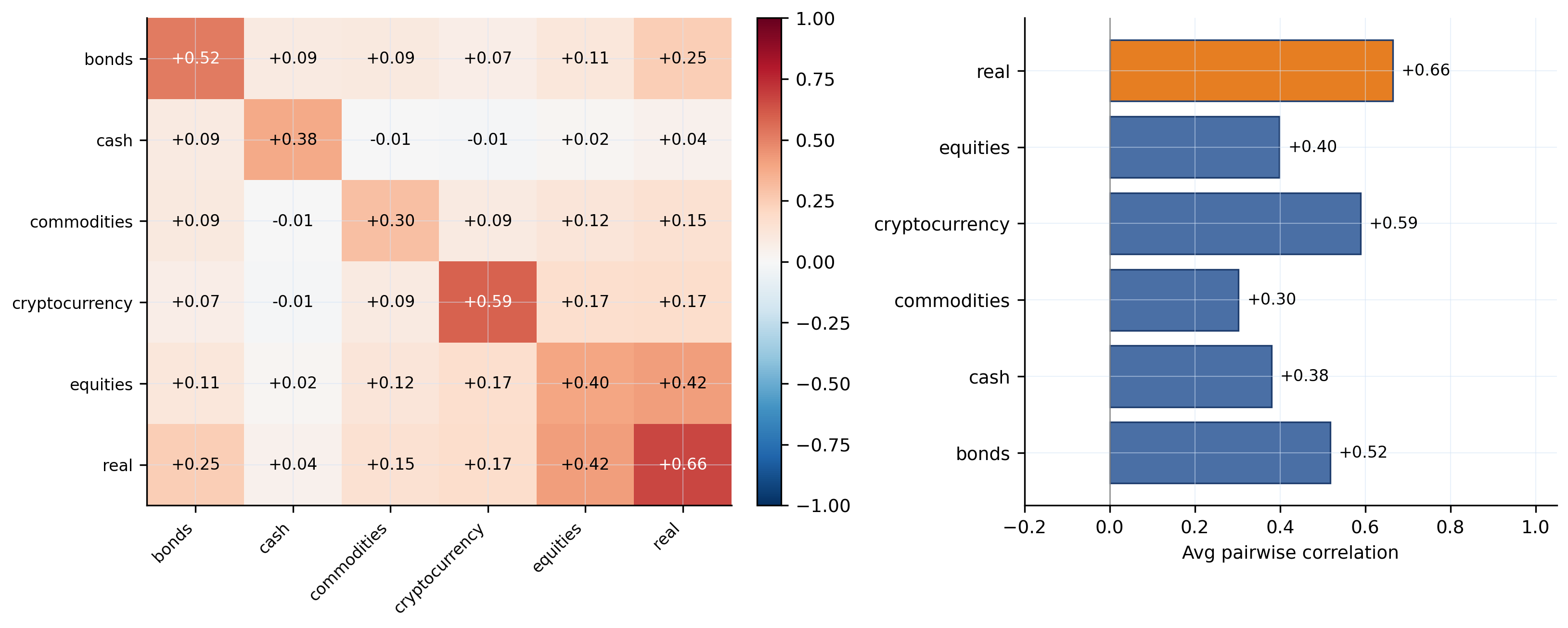}
\caption{Mean pairwise return correlations within and across asset classes. Left: the class-level correlation matrix, with diagonal entries denoting within-class means. Right: mean within-class correlation for each asset class.}
\label{fig:inter_class_correlation}
\end{figure*}

Figures~\ref{fig:correlation_heatmap} and~\ref{fig:inter_class_correlation} show low average correlation for several cross-class pairs and stronger positive correlation within equities and real estate. \textbf{Diversification depends on which asset classes receive weight: spreading capital inside a highly correlated class provides less hedging than allocating across weakly correlated classes.}

\section{Evaluation Protocol}
\label{app:evaluation}

\subsection{QA Task Templates}
\label{app:qa_templates}

QA targets are generated automatically from the \mbdataset and explicit task definitions, without manual item-level annotation. They include realized outcomes (T1), numerical calculations and decision rules (T2--T6), and a predefined regime-to-allocation policy (T7).

QA results report mean task scores in $[0,1]$. T1 uses exact direction-label agreement. T2/T3 use $\max(0,1-|\hat y-y|/|y|)$ for nonzero numeric targets; a zero target receives one only when $|\hat y|<10^{-6}$. T4/T5 use cosine similarity between predicted and target weight vectors. T6/T7 use the component scores defined below. Missing or unparseable answers receive zero, and the overall QA score averages the seven template means equally.

\noindent\textbf{T1 (Return direction).}
The ground truth is the sign of the realized $h$-day forward return, where $h$ is the forecast horizon specified in the question (default $h{=}21$ trading days). Direction labels are positive, negative, or flat (within $\pm 1\%$).

\noindent\textbf{T2 (VaR estimation).}
Historical simulation VaR at confidence level $p$ (e.g., $p{=}0.95$ or $0.99$; distinct from the S3 mixing weight $\alpha$):
\[
  \text{VaR}_p = \text{quantile}(r_{1:L},\; 1 - p)
\]
where $r_{1:L}$ is the trailing lookback return series provided in the question (default $L{=}60$ trading days). The scored answer is $\text{VaR}_p$; the question specifies the requested confidence level.

\noindent\textbf{T3 (Position sizing).}
The fixed-fractional Kelly-inspired formula~\citep{kelly1956new}:
\[
  f^* = \min\!\left(1.0,\;\frac{\delta_\text{max}}{|\text{VaR}_{99\%}|}\right)
\]
where $\delta_\text{max}$ is the question's drawdown limit.

\noindent\textbf{T4 (Long-only minimum-variance pairwise allocation).}
The prompt includes individual daily volatilities $\varsigma_1$, $\varsigma_2$ (distinct from pipeline stage scores $\sigma_t$), annualized mean returns $\mu_1$, $\mu_2$, and a minimum annualized portfolio return $\mu_{\min}$. The \emph{full} condition also provides the pairwise daily covariance and correlation; the \emph{restricted} condition in Section~\ref{app:info_level} removes these two statistics. The target minimizes two-asset variance subject to the stated return floor:
\begin{align*}
  \min_{w_1,w_2}\quad &w_1^2\varsigma_1^2+w_2^2\varsigma_2^2+2w_1w_2\varsigma_{12} \\
  \text{s.t.}\quad &w_1+w_2=1,\quad w_1,w_2\geq0, \\
  &w_1\mu_1+w_2\mu_2\geq\mu_{\min},
\end{align*}
where $\varsigma_{12}$ is the lookback sample covariance. Without a binding return floor, the long-only minimum-variance target is $w_1=\operatorname{clip}((\varsigma_2^2-\varsigma_{12})/(\varsigma_1^2+\varsigma_2^2-2\varsigma_{12}),0,1)$ and $w_2=1-w_1$; a vanishing denominator uses equal weights. When the floor binds, the weights satisfy $w_1\mu_1+w_2\mu_2=\mu_{\min}$, giving $w_1=(\mu_{\min}-\mu_2)/(\mu_1-\mu_2)$ and $w_2=1-w_1$.

\noindent\textbf{T5 (Maximum-Sharpe allocation).}
The long-only maximum-Sharpe portfolio for three or more assets, solved numerically via constrained optimization:
\[
  \max_{\mathbf{w}} \frac{\mathbf{w}^\top \boldsymbol{\mu} - r_f}{\sqrt{\mathbf{w}^\top \boldsymbol{\Sigma} \mathbf{w}}}
  \quad \text{s.t.} \quad \textstyle\sum_i w_i = 1,\; w_i \geq 0
\]
with a risk-free rate of $r_f = 4\%$ per annum and annualized $\boldsymbol{\mu}$, $\boldsymbol{\Sigma}$ estimated from the 60-day lookback window. The pipeline S3 reference in Appendix~\ref{app:metrics} uses the raw sample mean and covariance without $r_f$. Equal weight is used as a fallback if the optimizer does not converge. Under the \emph{full} condition the prompt includes the full mean vector $\boldsymbol{\mu}$ and covariance matrix $\boldsymbol{\Sigma}$; under the \emph{restricted} condition the covariance matrix and its header row are stripped (see Section~\ref{app:info_level}).

\noindent\textbf{T6 (Rebalancing decision with trade specification).}
The model is presented with a holdings table of current weights and target weights; pre-computed deviations are withheld. The ground truth is determined as follows. Let $i^* = \operatorname{arg\,max}_i |w_i^\text{current} - w_i^\text{target}|$ be the most off-target asset. The rebalancing flag is:
\[
  \text{rebalance} = \mathbf{1}\!\left[|w_{i^*}^\text{current} - w_{i^*}^\text{target}| > \delta_\text{reb}\right],
\]
with default threshold $\delta_\text{reb} = 0.05$ (distinct from the profile drawdown tolerance $\delta_\text{dd}$). Classes are balanced by construction, with half of all instances requiring rebalancing and half not, yielding exactly 50\% positive / 50\% negative labels.

The answer has two parts. \textbf{Part~A}: a yes/no rebalancing decision. \textbf{Part~B} (required when Part~A is yes): the corrective trade, expressed as ``\textit{sell} X.XXXX \textit{of} ASSET'' or ``\textit{buy} X.XXXX \textit{of} ASSET'', where ASSET $= i^*$ and the trade size is $|w_{i^*}^\text{current} - w_{i^*}^\text{target}|$ (reported but not scored).

Scoring decomposes as follows. If the ground truth is \textit{no}: score $= 1$ if the model answers no, else $0$. If the ground truth is \textit{yes}: score $= 0.40 \times d_\text{yes} + 0.60 \times (0.50 \times c_\text{dir} + 0.50 \times c_\text{asset})$, where $d_\text{yes} = \mathbf{1}[\text{model answers yes}]$, $c_\text{dir}$ indicates correct trade direction (buy versus sell), and $c_\text{asset}$ indicates correct asset identification.

\noindent\textbf{T7 (Regime detection and allocation).}
The target regime is the label assigned to the decision date by the regime classifier. The allocation target uses a predefined mapping from each regime to an increase, decrease, or hold direction for each asset class. This mapping specifies the benchmark's reference policy:
\begin{center}
\small
\begin{adjustbox}{max width=\columnwidth}
\begin{tabular}{lcccccc}
\toprule
\textbf{Regime} & EQ & BO & CO & RE & CR & CA \\
\midrule
Bull    & $\uparrow$ & $\downarrow$ & $\sim$ & $\uparrow$ & $\uparrow$ & $\downarrow$ \\
Bear    & $\downarrow$ & $\uparrow$ & $\sim$ & $\downarrow$ & $\downarrow$ & $\uparrow$ \\
Sideways & $\sim$ & $\sim$ & $\sim$ & $\sim$ & $\downarrow$ & $\uparrow$ \\
Crisis  & $\downarrow$ & $\uparrow$ & $\uparrow$ & $\downarrow$ & $\downarrow$ & $\uparrow$ \\
\bottomrule
\end{tabular}
\end{adjustbox}
\end{center}
EQ = equities, BO = bonds, CO = commodities, RE = real estate, CR = cryptocurrency, CA = cash. $\uparrow$ = increase, $\downarrow$ = decrease, $\sim$ = hold.
Scoring is $\text{score} = 0.5\, d_\text{reg} + 0.5\, \frac{1}{|\mathcal{C}|}\sum_{c \in \mathcal{C}} \mathbf{1}[\hat{a}_c = a_c^*]$, where $d_\text{reg} = \mathbf{1}[\widehat{\text{regime}} = \text{regime}^*]$, $\mathcal{C}$ is the set of six asset classes, and $\hat{a}_c$, $a_c^*$ are the predicted and ground-truth allocation directions for class $c$.

\subsection{Pipeline Metrics}
\label{app:metrics}

\paragraph{Stage scores and CEPS.}
Each stage has an explicit target. S1/S2 use trailing-return rules, S3 uses a profile-constrained numerical allocation and class exposure, and S4/S5 measure turnover and lookback risk-state agreement. For $\boldsymbol\sigma\in[0,1]^5$, the Cross-Stage Error Propagation Score is
\begin{align}
 \bar\sigma&=\tfrac15\sum_{t=1}^5\sigma_t,\label{eq:app_avg}\\
 \Delta_{\mathrm{cascade}}&=\sum_{t=1}^4\max(\sigma_t-\sigma_{t+1},0),\label{eq:app_cascade}\\
 \mathrm{CEPS}&=\operatorname{clip}(\bar\sigma-.1\Delta_{\mathrm{cascade}},0,1).\label{eq:app_ceps}
\end{align}
CEPS combines the mean reference-based stage score with a penalty for downstream score drops. The five stage scores receive equal weight in the mean. Paired interventions measure downstream responses to a reference replacement (Appendix~\ref{app:paired_details}).

\paragraph{S1: market interpretation.}
For eligible asset $i$, let $R_i^{\mathrm{lb}}$ be its cumulative return over the last 60 available observations within the preceding 90 calendar days, through the decision date. The reference view and agreement score are
\begin{align}
 v_i^*&=\operatorname{clip}(R_i^{\mathrm{lb}}/.10,-1,1),\label{eq:app_s1_gt}\\
 \sigma_1&=1-\frac{1}{2n}\sum_{i=1}^n|\hat v_i-v_i^*|.\label{eq:app_s1}
\end{align}
Here $\hat v_i\in[-1,1]$ is the model view and $n$ is the eligible security count; an absent view is neutral. The target specifies a reproducible momentum interpretation of the lookback data.

\paragraph{S2: signal generation.}
The reference maps $v_i^*>.2$ to buy, $v_i^*<-.2$ to sell, and the remaining values to hold. For model signals $\hat s_i$, agreement is
\begin{equation}
 \sigma_2=\frac1n\sum_{i=1}^n\mathbf1[\hat s_i=s_i^*].\label{eq:app_s2}
\end{equation}
Missing signals receive no agreement credit.

\paragraph{Profile-constrained lookback reference.}
Let $\mathcal B$ contain reference-buy assets, $\mathcal D$ the eligible bond and cash securities, and $\mathcal E$ equity and cryptocurrency assets. For profile $p$, the reference numerically maximizes
\begin{equation}
\begin{aligned}
 &\frac{\mathbf w^\top\boldsymbol\mu}{\sqrt{\mathbf w^\top\boldsymbol\Sigma\mathbf w+\epsilon}}\\
 \text{s.t. }&\sum_iw_i=1,\quad w_i\geq0,\\
 &w_i=0\quad(i\notin\mathcal B\cup\mathcal D),\\
 &\sum_{i\in\mathcal E}w_i\leq\alpha_p,\qquad
 \sum_{i\in\mathcal D}w_i\geq\beta_p.
\end{aligned}\label{eq:app_wstar}
\end{equation}
The sample mean $\boldsymbol\mu$ and covariance $\boldsymbol\Sigma$ use common observed dates in the trailing window, with no risk-free adjustment; $\epsilon=10^{-10}$ stabilizes the denominator. SLSQP starts from a feasible allocation, with tolerance $10^{-9}$ and at most 500 iterations. A solution is accepted only if the solver succeeds, satisfies the constraints, and matches or improves the initial objective; otherwise the feasible start is retained. Defensive securities remain eligible regardless of signal to implement the profile floor. Real estate follows its separate asset class and is not included in the equity cap.

We denote the resulting numerical reference by $\mathbf w_p^*$. It gives all models a common allocation target for each date and profile, using only past returns. The objective evaluates return relative to volatility within the lookback sample; subsequent net returns, drawdown, and the stress gate assess investment outcomes. Historical stage scores use this lookback reference. Appendix~\ref{app:reference_perspectives} compares it with an ex-post reference constructed from subsequent realized returns.

\paragraph{S3: allocation agreement and class exposure.}
For normalized long-only portfolios, including any uninvested cash, reference similarity is
\begin{equation}
 s_{\mathrm{acc}}=1-\tfrac12\sum_i|w_i-w_{p,i}^*|.\label{eq:app_sacc}
\end{equation}
It equals one for identical allocations and zero for disjoint allocations. S3 combines this term and class exposure with equal weight. Let $W_c$ be total exposure to class $c$, $\bar\rho_c$ the mean within-class off-diagonal correlation, and $\rho(c,c')$ the mean correlation between its eligible security pairs:
\begin{align}
 s_{\mathrm{intra}}&=\operatorname{clip}\!\left(1-\sum_cW_c\max(\bar\rho_c,0),0,1\right),\label{eq:app_sintra}\\
 \bar\rho_{\mathrm{cross}}&=\frac{\sum_{c\ne c'}W_cW_{c'}\rho(c,c')}{\sum_{c\ne c'}W_cW_{c'}},\label{eq:app_rho_cross}\\
 s_{\mathrm{inter}}&=\operatorname{clip}\!\left((1-\bar\rho_{\mathrm{cross}})/2,0,1\right),\label{eq:app_sinter}\\
 s_{\mathrm{corr}}&=\tfrac12(s_{\mathrm{intra}}+s_{\mathrm{inter}}).\label{eq:app_scorr}
\end{align}
With no positive cross-class weight product, $s_{\mathrm{inter}}=.5$. Uninvested cash contributes to cash-class exposure; class correlations use the eligible securities. These scores describe the class allocation. Redistributing weights inside a class leaves them unchanged, and with exactly two positive classes the cross-class weight product cancels.

\paragraph{Asset-level correlation risk.}
\label{app:correlation_scope}
Portfolio variance $V(\mathbf w)=\mathbf w^\top\boldsymbol\Sigma\mathbf w$ and the weighted pairwise correlation in Appendix~\ref{app:correlation_risk} use the actual security weights. The matched within-class allocation provides a controlled test: selected securities and class totals are identical, while covariance exposure can change. The class score describes how capital is split across asset classes; these asset-level measures assess the risk associated with the weights within each class.

\paragraph{S4: turnover agreement.}
Let $\tau(\mathbf u,\mathbf w)=\sum_{i\ne\mathrm{cash}}|w_i-u_i|$ be traded security weight from pre-trade weights $\mathbf u$. Both the proposed and reference targets are compared from the same pre-trade state. With $\tau=\tau(\mathbf u,\mathbf w)$ and $\tau^*=\tau(\mathbf u,\mathbf w_p^*)$,
\begin{equation}
 \sigma_4=1-\frac{|\tau-\tau^*|}{\max(\tau,\tau^*,10^{-4})}.\label{eq:app_s4}
\end{equation}
Agreement peaks at matching turnover. Trading costs are reported separately as a financial outcome.

\paragraph{S5: lookback risk-state agreement.}
Apply the target weights to common lookback returns and compute the 5th-percentile daily VaR and maximum drawdown, including initial wealth in the running peak. The rebalance flag is true when drawdown breaches the profile tolerance or the maximum deviation from security equal weighting exceeds .05. For model and reference flags $b,b^*$ and risk statistics,
\begin{align}
 e_V&=\frac{|V-V^*|}{\max(|V^*|,10^{-6})},\label{eq:app_s5_err_var}\\
 e_D&=\frac{|D-D^*|}{\max(|D^*|,10^{-6})},\label{eq:app_s5_err}\\
 \sigma_5&=\tfrac12\mathbf1[b=b^*]+\tfrac12\operatorname{clip}\!\left(1-\tfrac{e_V+e_D}{2},0,1\right).\label{eq:app_s5}
\end{align}
Here $V$ denotes VaR and $D$ drawdown in this equation. Smaller and larger risk statistics both reduce reference agreement; realized volatility, drawdown, and the stress gate independently assess portfolio risk. The profile VaR threshold is available for risk alerts and does not set the rebalance flag.

\subsection{Stress Windows and Profiles}
\label{app:stress_profiles}

\begin{table}[t]
\centering\small
\begin{adjustbox}{max width=\columnwidth}
\begin{tabular}{lll r}
\toprule
Window & Start & End & Decisions\\
\midrule
China shock & 2015-08-01 & 2016-02-29 & 30\\
COVID crash & 2020-02-01 & 2020-05-31 & 17\\
2022 decline & 2022-05-01 & 2022-12-31 & 35\\
2024 normal & 2024-01-01 & 2024-12-31 & 11\\
\bottomrule
\end{tabular}\end{adjustbox}
\caption{Evaluation windows and saved decisions per model--profile pair. Stress decisions are weekly; normal decisions are monthly, February--December.}
\label{tab:app_stress}
\end{table}

The weekly stress schedule resolves changes during rapidly evolving market conditions; monthly normal-period evaluation measures longer holding intervals. All comparison strategies, including EqW, share the schedule within each window, so their relative performance is compared at the same decision frequency. All three stress windows use weekly decisions, so the 2020--2022 reversal is observed at the same decision frequency. Comparisons with monthly decisions in 2024 also reflect a change in decision frequency. The stress gate passes when realized maximum drawdown magnitude is at most $\delta_p$ in every stress window.

\begin{table}[t]
\centering\small
\begin{tabular}{lrrrr}
\toprule
Profile & $\alpha_p$ & $\beta_p$ & $\delta_p$ & VaR limit\\
\midrule
Conservative & .40 & .40 & .10 & $-.010$\\
Balanced & .65 & .20 & .20 & $-.020$\\
Aggressive & .90 & .05 & .35 & $-.040$\\
\bottomrule
\end{tabular}
\caption{Profile parameters: equity-plus-crypto cap $\alpha_p$, bond-plus-cash floor $\beta_p$, and drawdown tolerance $\delta_p$. Uninvested cash counts toward the defensive floor.}
\label{tab:app_profiles}
\end{table}

Profiles appear in the S1/S2 prompts; S3 uses the resulting signals and market statistics. Exposure compliance evaluates both target-weight inequalities in Table~\ref{tab:app_profiles}.

\subsection{Baselines and Common Portfolio Replay}
\label{app:evaluation_details}

\paragraph{Tradable securities.}
The benchmark comprises 183 market series spanning six asset classes, supporting portfolio QA, macroeconomic and news context, and broad market coverage. Portfolio evaluation supplies 99 series with matched price histories: 43 equities, 15 bonds, 16 commodities, 10 real-estate assets, 10 cryptocurrencies, 4 cash instruments, and VIX. VIX is a market indicator, and any weight assigned to it is held as uninvested cash. The other 98 series are the investable assets. The remaining 84 corpus series (82 equities and 2 cryptocurrencies) are excluded from portfolio allocation. At each decision date, an investable asset is eligible when its price is positive and at least 20 return observations are available over the preceding 90 calendar days. Weight assigned to an ineligible asset is held as uninvested cash and is not redistributed. All compared portfolios use this eligibility rule.

\paragraph{Execution and return accounting.}
Every portfolio starts with unit NAV in cash and replays its archived target sequence. A decision is executed at the first observed market close strictly after its date. Existing positions accrue that date's return before trading, and new positions earn subsequent returns. The calendar uses observed SPY trading dates. Outcomes use archived adjusted prices: Yahoo series for ETFs and the adjusted-close column of the NASDAQ-100 source for individual equities. Short price gaps are forward-filled for at most five trading dates before computing returns. Each dated return is applied once. Commission of 5\,bps and slippage of 10\,bps are charged on the absolute changes in security weights; transferring the cash balance has no additional fee. Holdings drift between trades. Costs are relative to initial NAV.

Sharpe uses $(252\bar r-.04)/(\sqrt{252}s_r)$ from the net daily portfolio returns, with a fixed 4\% annual risk-free rate across windows. Volatility is $\sqrt{252}s_r$, total return is terminal NAV minus one, and maximum drawdown includes initial NAV in the running maximum. Each window contains all ten models and three profiles. The 2,790 saved decisions remain in coverage and allocation records; 60 terminal decisions have no following execution close inside their windows, leaving 2,730 executed decisions for stage-score summaries. All 120 portfolio cells remain in financial comparisons. The 330 normal-period decisions all execute. For MinVar in 2024, the risk-free adjustment contributes $-3.13$ to Sharpe at 1.28\% annualized volatility; cumulative trading costs are 0.76\% of initial NAV.

\paragraph{Classical strategies.}
EqW distributes capital across the eligible securities. The 60/40 baseline equally weights eligible equities within its 60\% sleeve and eligible bonds within its 40\% sleeve. RiskPar uses inverse volatility; CovRiskPar uses full-covariance equal risk contribution; MinVar uses long-only minimum variance; BL combines a return prior and directional views. Covariance-based strategies use the common observed trailing returns with 252-day annualization. These six strategies retain their standard allocation rules, providing fixed strategy comparators across investor profiles.

\subsection{Statistical Analysis}
\label{app:postprocessing}

Per-template QA means retain all 50 evaluated items for each model. The seven-template QA aggregate weights each template equally. QA--Sharpe correlations use ten paired model observations separately for each period and profile. CEPS--outcome associations likewise compare models within each profile. Percentile intervals use 10,000 paired model-bootstrap draws. For the covariance comparisons, the model is the resampling cluster, retaining its profiles and dates together. The seed is 20261006. Intervals describe these evaluated model sets and are not adjusted for multiple comparisons. The 65/120 beat rate describes the fixed evaluation grid; its cells share models and market windows. Allocation and profile-change intervals resample models with their dates and profiles retained. Table~\ref{tab:app_active_tilt} applies Benjamini--Hochberg correction to its three Spearman association tests.

\section{QA Dataset and Results}
\label{app:qa_details}

\subsection{Dataset Statistics}
\label{app:qa_stats}

\begin{figure}[!h]
\centering
\includegraphics[width=\columnwidth]{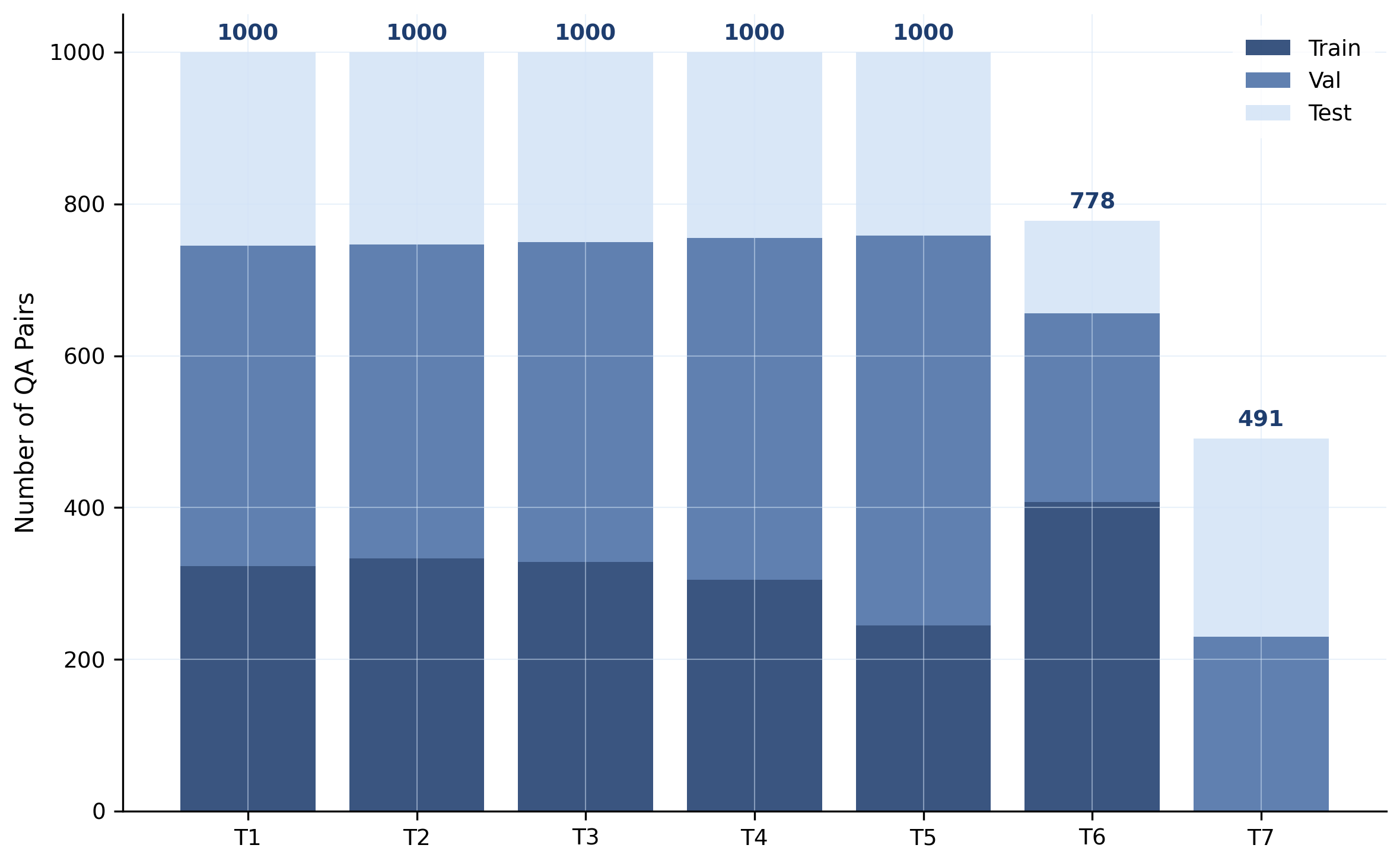}
\caption{QA sample counts by template and data split.}
\label{fig:app_template_by_split}
\end{figure}

The QA dataset contains 6,269 pairs across seven templates (Figure~\ref{fig:app_template_by_split}): T1--T5 each have 1,000 samples, while T6 and T7 have 778 and 491. T7 has no training split because its adaptive allocation labels are reserved for validation and testing. T6 has balanced labels: 50/50 rebalance versus hold.

\begin{figure}[!h]
\centering
\includegraphics[width=\columnwidth]{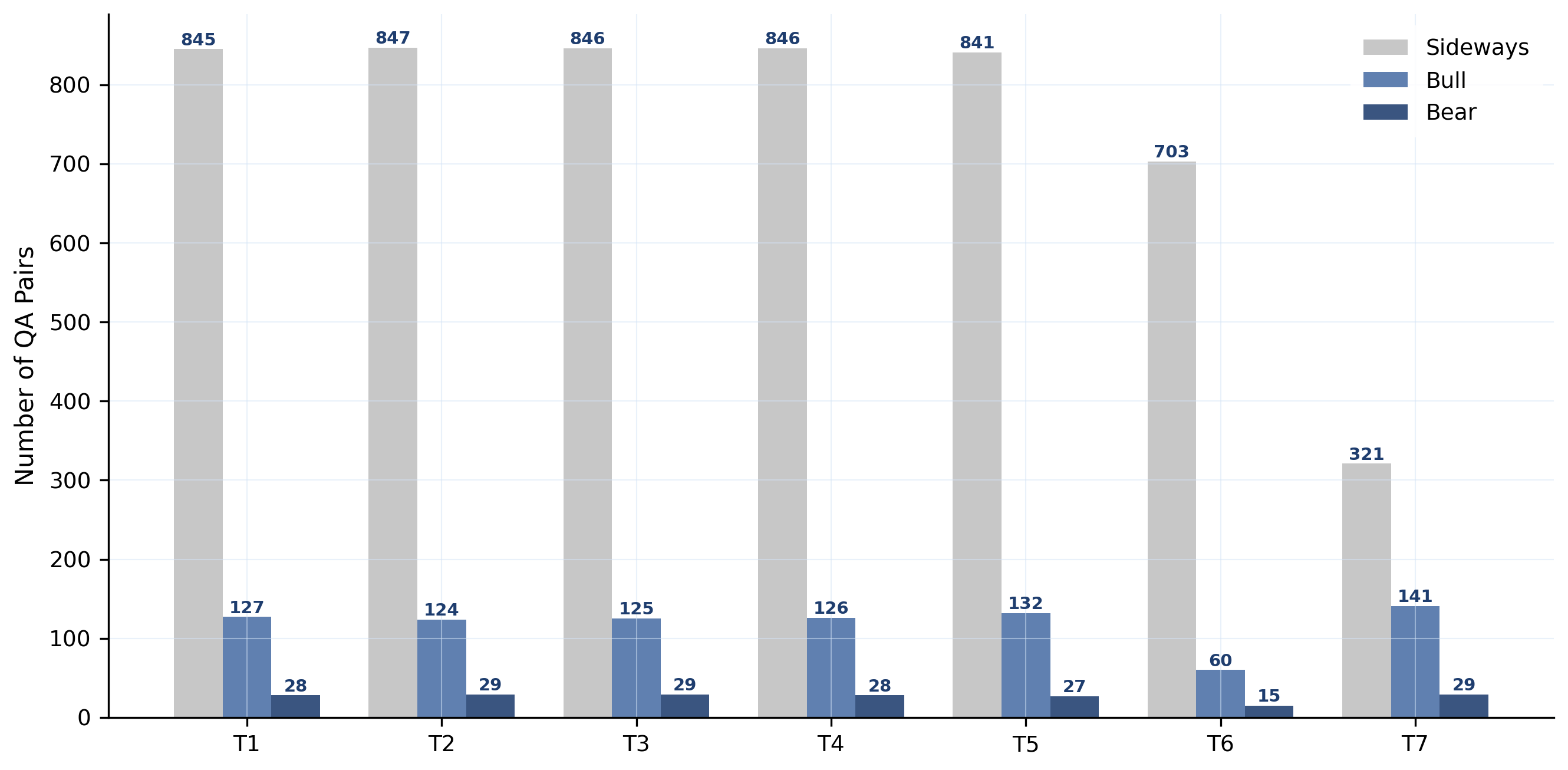}
\caption{QA counts by template and market regime.}
\label{fig:app_template_by_regime}
\end{figure}

Figure~\ref{fig:app_template_by_regime} shows the regime composition. Sideways markets dominate every template ($>$65\%), matching the empirical prevalence of range-bound periods. T1--T5 share nearly identical regime proportions because they sample from the same date pool; T7 raises the bull-market share to 29\% so that its adaptive allocation task has adequate coverage of non-sideways regimes.

\begin{figure}[!h]
\centering
\includegraphics[width=\columnwidth]{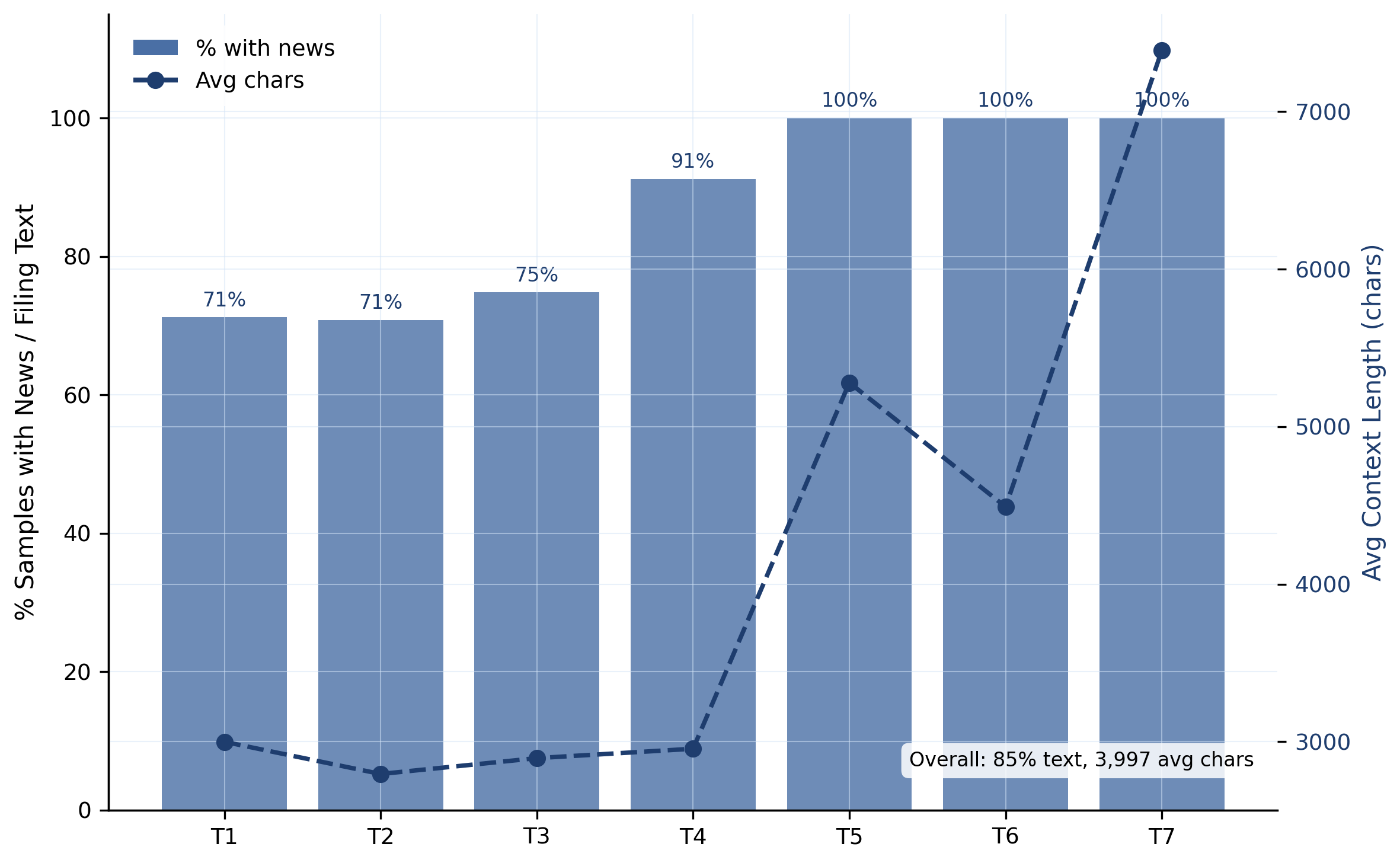}
\caption{Text richness by template: news or filing coverage (bars, left axis), mean context length (dashed line, right axis), and global averages (inset).}
\label{fig:app_text_richness}
\end{figure}

The amount of textual context increases with template complexity (Figure~\ref{fig:app_text_richness}). L1 templates (T1--T3) include news or SEC filings in 71--75\% of samples at roughly 2,800--3,000 characters; L2 (T4) is intermediate; and L3--L4 (T5--T7) reach 100\% coverage at 4,500--7,388 characters. T7 provides the longest contexts (7,388 chars), combining news and macro signals for regime-aware allocation. Overall, 85.3\% of pairs carry textual context, with a mean length of 3,997 characters.
\textbf{Context coverage and length rise with complexity: partial coverage in T1--T3, complete in T5--T7.}

\subsection{Full QA Results}
\label{app:full_qa}

Table~\ref{tab:app_full_qa} reports QA results: per-template mean scores with full information (T1--T7), the restricted condition without the covariance matrix (T4\textsubscript{r}, T5\textsubscript{r}), formula-focused allocation versus broader-task averages (F = mean of T4+T5; B = mean of T1+T2+T6+T7; T3 excluded as eight of ten models exceed 0.94), and mean scores by market regime.

\begin{table*}[!ht]
\centering
\small
\begin{adjustbox}{max width=\textwidth,center}
\begin{tabular}{l rrrrrrr r rr rr rrr}
\toprule
& \multicolumn{7}{c}{\textbf{Per-Template (Full)}} & & \multicolumn{2}{c}{\textbf{Restricted}} & \multicolumn{2}{c}{\textbf{Task Type}} & \multicolumn{3}{c}{\textbf{Market Regime}} \\
\cmidrule(lr){2-8} \cmidrule(lr){10-11} \cmidrule(lr){12-13} \cmidrule(lr){14-16}
\textbf{Model} & \textbf{T1} & \textbf{T2} & \textbf{T3} & \textbf{T4} & \textbf{T5} & \textbf{T6} & \textbf{T7} & \textbf{Mean} & \textbf{T4\textsubscript{r}} & \textbf{T5\textsubscript{r}} & \textbf{F} & \textbf{B} & \textbf{Bull} & \textbf{Bear} & \textbf{Side.} \\
\midrule
DS-V4-Flash      & \cellcolor{bestcell}\textbf{.520} & .843 & .945 & \cellcolor{bestcell}\textbf{1.00} & .932 & .652 & \cellcolor{bestcell}\textbf{.843} & \cellcolor{bestcell}\textbf{.819} & .975 & .860 & .966 & .715 & .827 & .823 & \cellcolor{bestcell}\textbf{.812} \\
Qwen3.7-Max      & .500 & \cellcolor{bestcell}\textbf{.859} & .951 & \cellcolor{bestcell}\textbf{1.00} & .954 & .724 & .742 & \cellcolor{bestcell}\textbf{.819} & \cellcolor{bestcell}\textbf{1.00} & \cellcolor{bestcell}\textbf{.990} & .977 & .706 & .814 & \cellcolor{bestcell}\textbf{.863} & .810 \\
DS-V4-Pro        & \cellcolor{bestcell}\textbf{.520} & .837 & .963 & \cellcolor{bestcell}\textbf{1.00} & \cellcolor{bestcell}\textbf{.992} & .652 & .760 & .818 & \cellcolor{bestcell}\textbf{1.00} & .660 & \cellcolor{bestcell}\textbf{.996} & .692 & \cellcolor{bestcell}\textbf{.844} & .846 & .802 \\
DB-2.0-Lite      & .460 & .798 & .957 & .956 & .897 & .810 & .747 & .804 & .961 & .940 & .927 & .704 & .780 & .846 & .806 \\
DB-2.0-Pro       & .440 & .847 & .963 & .991 & .912 & .824 & .530 & .787 & .979 & .923 & .952 & .660 & .764 & .806 & .792 \\
Qwen3.6-Plus     & .440 & .858 & \cellcolor{bestcell}\textbf{.968} & \cellcolor{bestcell}\textbf{1.00} & .804 & .640 & .768 & .783 & \cellcolor{bestcell}\textbf{1.00} & .810 & .902 & .677 & .799 & .801 & .771 \\
GLM-5.1          & .440 & .855 & .964 & \cellcolor{bestcell}\textbf{1.00} & .421 & \cellcolor{bestcell}\textbf{.882} & .738 & .757 & \cellcolor{bestcell}\textbf{1.00} & .531 & .711 & \cellcolor{bestcell}\textbf{.729} & .778 & .765 & .746 \\
Qwen3.6-35B-A3B  & .460 & .808 & .961 & \cellcolor{bestcell}\textbf{1.00} & .230 & .564 & .763 & .684 & \cellcolor{bestcell}\textbf{1.00} & .320 & .615 & .649 & .714 & .729 & .662 \\
\rowcolor{worstcell}
HY3-Preview      & .460 & .386 & .336 & .975 & .958 & .468 & .783 & .624 & .982 & .974 & .967 & .524 & .664 & .663 & .597 \\
\rowcolor{worstcell}
Kimi-K2.6        & .420 & .422 & .493 & .956 & .280 & .684 & .320 & .511 & .978 & .710 & .618 & .462 & .556 & .531 & .487 \\
\bottomrule
\end{tabular}
\end{adjustbox}
\caption{Complete QA evaluation results. Per-template mean scores under the full information condition (T1--T7), restricted condition without the covariance matrix (T4\textsubscript{r}, T5\textsubscript{r}), formula-focused allocation and broader-task averages (F, B), and mean scores by market regime. Models are ranked by mean score. \colorbox{bestcell}{Green} = column best; \colorbox{worstcell}{pink rows} = bottom two models with lower mean scores (Mean $< 0.65$).}
\label{tab:app_full_qa}
\end{table*}

The overall QA score ranges from .511 to .819. The widest template spreads occur on T3 (.336--.968), T5 (.230--.992), and T7 (.320--.843), while T4 remains near ceiling for every model. \textbf{Aggregate QA rankings therefore combine tasks that differ substantially in both difficulty and the size of score differences between models.}

We also recompute the QA average without T3 and T4. The resulting ranking has Spearman $\rho=.988$ with the seven-template ranking, and no model moves by more than one place.

\subsection{T3/T4 Numeric and Structured Decisions}
\label{app:qa_saturation}

Table~\ref{tab:app_full_qa} shows near-ceiling numeric scores: eight of ten models exceed 0.94 on T3 and every model scores at least 0.956 on T4. Both templates provide quantitative inputs for sizing or pairwise allocation.

We further evaluate four models on a separate set of 50 T3 and 50 T4 structured-decision items per model. Each output gives a plan, feasible set, and binding constraints. Field score averages correctness across these fields; exact match requires all fields to be correct. Table~\ref{tab:app_t34_model_diag} shows high field accuracy for DB-2.0-Pro and the two DS-V4 models, with lower exact-match rates. HY3-Preview produces valid JSON for all 100 items but no exact matches; 39 T3 and 44 T4 outputs receive partial credit. Its errors span constraint propagation, arithmetic, table reading, and multi-field consistency.

\begin{table}[!h]
\centering
\small
\begin{adjustbox}{max width=\columnwidth,center}
\begin{tabular}{lrrrr}
\toprule
& \multicolumn{2}{c}{\textbf{T3}} & \multicolumn{2}{c}{\textbf{T4}} \\
\cmidrule(lr){2-3}\cmidrule(lr){4-5}
\textbf{Model} & \textbf{Field} & \textbf{Exact} & \textbf{Field} & \textbf{Exact} \\
\midrule
DB-2.0-Pro  & .956 & .820 & .965 & .840 \\
DS-V4-Pro   & .953 & .800 & .989 & .940 \\
DS-V4-Flash & .931 & .740 & .922 & .680 \\
HY3-Preview & .238 & .000 & .350 & .000 \\
\bottomrule
\end{tabular}
\end{adjustbox}
\caption{Supplementary T3/T4 structured-decision results on separate items. Field score averages correctness across the selected plan, feasible set, and binding constraints; exact match requires all fields to be correct. Main QA results use the original tasks in Table~\ref{tab:app_full_qa}.}
\label{tab:app_t34_model_diag}
\end{table}

For DB-2.0-Pro and the two DS-V4 models, mean field/exact scores are .947/.787 on T3 and .959/.820 on T4. Their 95\% CIs are [.924, .968]/[.707, .860] and [.938, .976]/[.747, .880], respectively, from 10,000 item-bootstrap samples retaining all three outputs per sampled item. \textbf{The supplementary tasks distinguish component accuracy from complete consistency across the selected plan, feasible set, and binding constraints.}

\subsection{Task Diagnostics}
\label{app:info_level}
\label{app:formula_judgment}

\paragraph{Information ablation.}

T4 and T5 prompts expose the statistics used by their allocation formulas: T4 includes pairwise covariance and correlation, and T5 includes the mean vector and covariance matrix. The restricted prompts remove covariance and correlation from T4 and the covariance matrix and its header from T5, retaining the T5 mean vector. The difference $\Delta=\text{score}_\text{full}-\text{score}_\text{restricted}$ measures the change in mean task score when covariance information is supplied. Results: Table~\ref{tab:app_full_qa}.

On T5, only the two DS-V4 models have $\Delta>0$: DS-V4-Pro falls by 0.332 and DS-V4-Flash by 0.073 when the matrix is removed. The mean T5 score rises for the other eight models, from 0.006 for Qwen3.6-Plus to 0.430 for Kimi-K2.6; Kimi-K2.6 rises from 0.280 to 0.710. T4 remains near ceiling in both conditions, ranging from 0.956 to 1.000 with full information and from 0.961 to 1.000 without the covariance statistics. The T5 pattern shows that the supplied matrix can hinder answer generation, although the ablation cannot separate table parsing from numerical reasoning.

Pairing the archived full and restricted responses by model and item shows that 460/500 T4 scores (92\%) and 360/500 T5 scores (72\%) are unchanged. The aggregate ablation differences are therefore concentrated in a subset of responses. These counts retain every evaluated item, including zero scores.

\paragraph{Supplied-statistic allocation versus broader tasks.}

We compare formula-focused allocation tasks (T4: minimum-variance allocation; T5: maximum-Sharpe optimization, both with covariance statistics supplied) with the broader set of return prediction, VaR estimation, rebalancing, and regime detection tasks (T1, T2, T6, and T7). T3 is reported separately because eight of ten models score above 0.94. The F and B columns of Table~\ref{tab:app_full_qa} report these descriptive averages for all models.

The mean T4--T5 score is 0.863, compared with 0.652 for T1--T2 and T6--T7. This 0.211 difference favors T4--T5 for eight of ten models. GLM-5.1 and Qwen3.6-35B-A3B reverse the aggregate pattern, with T6 scores of 0.882 and 0.564 versus T5 scores of 0.421 and 0.230. These averages reflect different information and scoring conditions.

\subsection{QA and Portfolio Outcomes}
\label{app:qa_financial}

We relate the seven-template QA macro-average to portfolio Sharpe across four market windows and three investor profiles. The comparison uses 3,500 responses from ten models, seven templates, and 50 common items per template.

Each period--profile association contains ten paired model observations. The same QA scores recur across settings, so 120 portfolio cells are not 120 independent QA observations. Table~\ref{tab:qa_financial_direct} reports Spearman correlations and percentile intervals from 10,000 paired model-bootstrap draws.
\begin{table}[t]
\centering
\small
\begin{adjustbox}{max width=\columnwidth,center}
\begin{tabular}{llr}
\toprule
Window & Profile & $\rho$ [95\% CI] \\
\midrule
2015--16 & Conservative & $0.83$ [$0.40$, $1.00$] \\
2015--16 & Balanced & $0.71$ [$0.10$, $0.96$] \\
2015--16 & Aggressive & $-0.67$ [$-0.96$, $0.01$] \\
2020 & Conservative & $0.20$ [$-0.61$, $0.79$] \\
2020 & Balanced & $-0.33$ [$-0.90$, $0.52$] \\
2020 & Aggressive & $0.38$ [$-0.45$, $0.90$] \\
2022 & Conservative & $0.12$ [$-0.69$, $0.79$] \\
2022 & Balanced & $-0.59$ [$-0.99$, $0.18$] \\
2022 & Aggressive & $0.36$ [$-0.51$, $0.77$] \\
2024 & Conservative & $-0.10$ [$-0.90$, $0.77$] \\
2024 & Balanced & $-0.08$ [$-0.69$, $0.60$] \\
2024 & Aggressive & $-0.42$ [$-0.89$, $0.37$] \\
\bottomrule
\end{tabular}
\end{adjustbox}
\caption{QA--Sharpe associations across ten paired models per setting. QA uses the macro-average over all seven templates. Intervals use 10,000 paired model-bootstrap draws.}
\label{tab:qa_financial_direct}
\end{table}

Correlations range from $-.67$ to $.83$. QA scores are positively associated with Sharpe in the conservative and balanced 2015--16 settings, whose two unadjusted intervals exclude zero. The other ten intervals include zero. In 2024, the estimates are $-.10$, $-.08$, and $-.42$. The correlations vary across periods and profiles. These descriptive comparisons motivate evaluating candidate systems on portfolio outcomes under the intended profile and decision schedule.

\section{Financial Outcomes}
\label{app:financial_outcomes}

\subsection{Stress-Gate Results}
\label{app:stress_eval}

Table~\ref{tab:app_stress_detail} reports maximum drawdown for every model--profile pair in all three stress windows. Gate passage uses the stated profile tolerance across all windows. No model passes all three at the conservative limit; nine pass at the balanced limit and all ten at the aggressive limit.
\begin{table*}[t]
\centering
\small
\begin{adjustbox}{max width=\textwidth,center}
\begin{tabular}{llrrrc}
\toprule
Profile & Model & 2015--16 DD\% & 2020 DD\% & 2022 DD\% & Gate \\
\midrule
Conservative & DS-V4-Flash & -11.30 & -10.53 & -11.92 & $\times$ \\
 & DS-V4-Pro & -7.51 & -17.69 & -13.32 & $\times$ \\
 & Qwen3.6-Plus & -6.37 & -11.16 & -10.94 & $\times$ \\
 & Qwen3.7-Max & -9.17 & -6.28 & -11.15 & $\times$ \\
 & Qwen3.6-35B-A3B & -11.89 & -15.32 & -11.50 & $\times$ \\
 & GLM-5.1 & -9.01 & -10.93 & -14.06 & $\times$ \\
 & DB-2.0-Lite & -6.09 & -11.18 & -8.28 & $\times$ \\
 & DB-2.0-Pro & -8.42 & -13.38 & -11.53 & $\times$ \\
 & HY3-Preview & -8.48 & -11.28 & -14.04 & $\times$ \\
 & Kimi-K2.6 & -11.32 & -13.02 & -14.63 & $\times$ \\
\midrule
Balanced & DS-V4-Flash & -8.93 & -22.19 & -17.30 & $\times$ \\
 & DS-V4-Pro & -8.65 & -15.81 & -14.13 & $\checkmark$ \\
 & Qwen3.6-Plus & -8.38 & -11.70 & -12.71 & $\checkmark$ \\
 & Qwen3.7-Max & -9.24 & -10.66 & -15.21 & $\checkmark$ \\
 & Qwen3.6-35B-A3B & -9.51 & -16.32 & -13.87 & $\checkmark$ \\
 & GLM-5.1 & -8.50 & -12.99 & -14.73 & $\checkmark$ \\
 & DB-2.0-Lite & -8.45 & -17.86 & -14.14 & $\checkmark$ \\
 & DB-2.0-Pro & -8.55 & -15.46 & -13.02 & $\checkmark$ \\
 & HY3-Preview & -8.13 & -14.60 & -15.45 & $\checkmark$ \\
 & Kimi-K2.6 & -10.52 & -12.57 & -12.90 & $\checkmark$ \\
\midrule
Aggressive & DS-V4-Flash & -11.02 & -14.39 & -19.24 & $\checkmark$ \\
 & DS-V4-Pro & -13.46 & -16.47 & -18.56 & $\checkmark$ \\
 & Qwen3.6-Plus & -10.01 & -14.42 & -15.46 & $\checkmark$ \\
 & Qwen3.7-Max & -17.80 & -11.71 & -16.19 & $\checkmark$ \\
 & Qwen3.6-35B-A3B & -12.43 & -13.77 & -17.53 & $\checkmark$ \\
 & GLM-5.1 & -11.95 & -15.99 & -15.33 & $\checkmark$ \\
 & DB-2.0-Lite & -8.68 & -16.68 & -15.87 & $\checkmark$ \\
 & DB-2.0-Pro & -11.01 & -13.78 & -13.54 & $\checkmark$ \\
 & HY3-Preview & -8.12 & -15.86 & -16.13 & $\checkmark$ \\
 & Kimi-K2.6 & -13.79 & -12.57 & -19.84 & $\checkmark$ \\
\bottomrule
\end{tabular}
\end{adjustbox}
\caption{Stress-window drawdowns and the joint drawdown gate for every model and profile.}
\label{tab:app_stress_detail}
\end{table*}

Table~\ref{tab:app_stress_stages} reports stage scores and drawdown for GLM-5.1, Qwen3.6-Plus (Qwen3.7-Max for the aggressive profile), HY3-Preview, and DS-V4-Flash.

\begin{table*}[!ht]
\centering
\small
\begin{adjustbox}{max width=\textwidth,center}
\begin{tabular}{llrrrrrrrc}
\toprule
\textbf{Model} & \textbf{Scenario} & \textbf{S1} & \textbf{S2} & \textbf{S3} & \textbf{S4} & \textbf{S5} & \textbf{CEPS} & \textbf{MaxDD\%} & \textbf{Pass} \\
\midrule
\multicolumn{10}{l}{\textit{Conservative (10\% MaxDD tolerance)}} \\
\midrule
GLM-5.1 & 2015 China & .790 & .537 & .318 & .170 & .501 & .393 & $-$9.01 & \checkmark \\
 & 2020 COVID & .713 & .528 & .337 & .175 & .420 & .369 & $-$10.93 & \cellcolor{worstcell}$\times$ \\
 & 2022 Decline & .744 & .529 & .305 & .143 & .552 & .390 & $-$14.06 & \cellcolor{worstcell}$\times$ \\
\midrule
Qwen3.6-Plus & 2015 China & .783 & .489 & .334 & .200 & .381 & .369 & $-$6.37 & \checkmark \\
 & 2020 COVID & .709 & .505 & .338 & .220 & .385 & .367 & $-$11.16 & \cellcolor{worstcell}$\times$ \\
 & 2022 Decline & .740 & .512 & .293 & .215 & .528 & .395 & $-$10.94 & \cellcolor{worstcell}$\times$ \\
\midrule
HY3-Preview & 2015 China & .766 & .441 & .348 & .099 & .144 & .282 & $-$8.48 & \checkmark \\
 & 2020 COVID & .712 & .534 & .361 & .174 & .276 & .338 & $-$11.28 & \cellcolor{worstcell}$\times$ \\
 & 2022 Decline & .717 & .430 & .267 & .115 & .198 & .273 & $-$14.04 & \cellcolor{worstcell}$\times$ \\
\midrule
DS-V4-Flash & 2015 China & .783 & .473 & .338 & .460 & .337 & .399 & $-$11.30 & \cellcolor{worstcell}$\times$ \\
 & 2020 COVID & .716 & .386 & .300 & .376 & .312 & .350 & $-$10.53 & \cellcolor{worstcell}$\times$ \\
 & 2022 Decline & .737 & .413 & .257 & .370 & .363 & .354 & $-$11.92 & \cellcolor{worstcell}$\times$ \\
\midrule
\multicolumn{10}{l}{\textit{Balanced (20\% MaxDD tolerance)}} \\
\midrule
GLM-5.1 & 2015 China & .790 & .544 & .325 & .197 & .500 & .403 & $-$8.50 & \checkmark \\
 & 2020 COVID & .716 & .527 & .326 & .192 & .419 & .370 & $-$12.99 & \checkmark \\
 & 2022 Decline & .744 & .529 & .279 & .143 & .483 & .369 & $-$14.73 & \checkmark \\
\midrule
Qwen3.6-Plus & 2015 China & .785 & .526 & .320 & .193 & .422 & .381 & $-$8.38 & \checkmark \\
 & 2020 COVID & .697 & .464 & .288 & .261 & .250 & .326 & $-$11.70 & \checkmark \\
 & 2022 Decline & .737 & .498 & .293 & .224 & .418 & .370 & $-$12.71 & \checkmark \\
\midrule
HY3-Preview & 2015 China & .771 & .446 & .330 & .131 & .107 & .277 & $-$8.13 & \checkmark \\
 & 2020 COVID & .707 & .486 & .322 & .173 & .387 & .346 & $-$14.60 & \checkmark \\
 & 2022 Decline & .716 & .402 & .258 & .130 & .042 & .234 & $-$15.45 & \checkmark \\
\midrule
DS-V4-Flash & 2015 China & .786 & .400 & .313 & .285 & .276 & .333 & $-$8.93 & \checkmark \\
 & 2020 COVID & .711 & .425 & .298 & .368 & .065 & .291 & $-$22.19 & \cellcolor{worstcell}$\times$ \\
 & 2022 Decline & .737 & .468 & .247 & .387 & .285 & .348 & $-$17.30 & \checkmark \\
\midrule
\multicolumn{10}{l}{\textit{Aggressive (35\% MaxDD tolerance)}} \\
\midrule
GLM-5.1 & 2015 China & .794 & .574 & .316 & .207 & .523 & .416 & $-$11.95 & \checkmark \\
 & 2020 COVID & .708 & .535 & .302 & .279 & .406 & .387 & $-$15.99 & \checkmark \\
 & 2022 Decline & .744 & .530 & .286 & .202 & .509 & .393 & $-$15.33 & \checkmark \\
\midrule
Qwen3.7-Max & 2015 China & .789 & .545 & .323 & .258 & .482 & .411 & $-$17.80 & \checkmark \\
 & 2020 COVID & .723 & .565 & .335 & .385 & .417 & .420 & $-$11.71 & \checkmark \\
 & 2022 Decline & .747 & .535 & .314 & .261 & .454 & .395 & $-$16.19 & \checkmark \\
\midrule
HY3-Preview & 2015 China & .774 & .459 & .309 & .144 & .393 & .340 & $-$8.12 & \checkmark \\
 & 2020 COVID & .694 & .483 & .297 & .203 & .098 & .281 & $-$15.86 & \checkmark \\
 & 2022 Decline & .715 & .410 & .251 & .142 & .133 & .255 & $-$16.13 & \checkmark \\
\midrule
DS-V4-Flash & 2015 China & .772 & .443 & .292 & .336 & .102 & .310 & $-$11.02 & \checkmark \\
 & 2020 COVID & .682 & .425 & .270 & .373 & .125 & .302 & $-$14.39 & \checkmark \\
 & 2022 Decline & .742 & .383 & .235 & .461 & .071 & .282 & $-$19.24 & \checkmark \\
\bottomrule
\end{tabular}
\end{adjustbox}
\caption{Stress-window stage scores and drawdown for four models per profile. The gate table covers ten models.}
\label{tab:app_stress_stages}
\end{table*}

\subsection{Cross-Period Performance}
\label{app:cross_regime}

Table~\ref{tab:app_cross_regime} compares balanced-profile Sharpe with EqW under the same universe, schedule, and cost model in each window. Figure~\ref{fig:cross_period} includes all three profiles. Appendix~\ref{app:nav_trajectories} plots daily NAV trajectories.
\begin{table}[t]
\centering
\small
\begin{adjustbox}{max width=\columnwidth,center}
\begin{tabular}{lrrrr}
\toprule
Model & 2015--16 & 2020 & 2022 & 2024 \\
\midrule
EqW & -1.047 & -0.602 & -0.540 & 0.710 \\
DS-V4-Flash & 0.290 & -1.250 & -1.451 & 0.609 \\
DS-V4-Pro & -0.452 & -0.302 & -1.460 & 0.542 \\
Qwen3.6-Plus & -0.034 & 1.696 & -1.095 & 0.798 \\
Qwen3.7-Max & -0.214 & 0.099 & -2.146 & 0.783 \\
Qwen3.6-35B-A3B & -0.564 & 0.078 & -0.948 & 0.740 \\
GLM-5.1 & -0.217 & -0.170 & -0.924 & 0.603 \\
DB-2.0-Lite & -0.255 & -0.705 & -1.229 & 0.838 \\
DB-2.0-Pro & -0.681 & 0.062 & -1.073 & 0.662 \\
HY3-Preview & -0.819 & -0.725 & -1.908 & 0.650 \\
Kimi-K2.6 & -0.790 & 0.239 & -0.702 & 0.769 \\
\bottomrule
\end{tabular}
\end{adjustbox}
\caption{Balanced-profile Sharpe across the four historical windows. Stress windows use weekly decisions; 2024 uses monthly decisions.}
\label{tab:app_cross_regime}
\end{table}

\subsection{NAV Trajectories}
\label{app:nav_trajectories}

\begin{figure}[!h]
\centering
\includegraphics[width=\columnwidth]{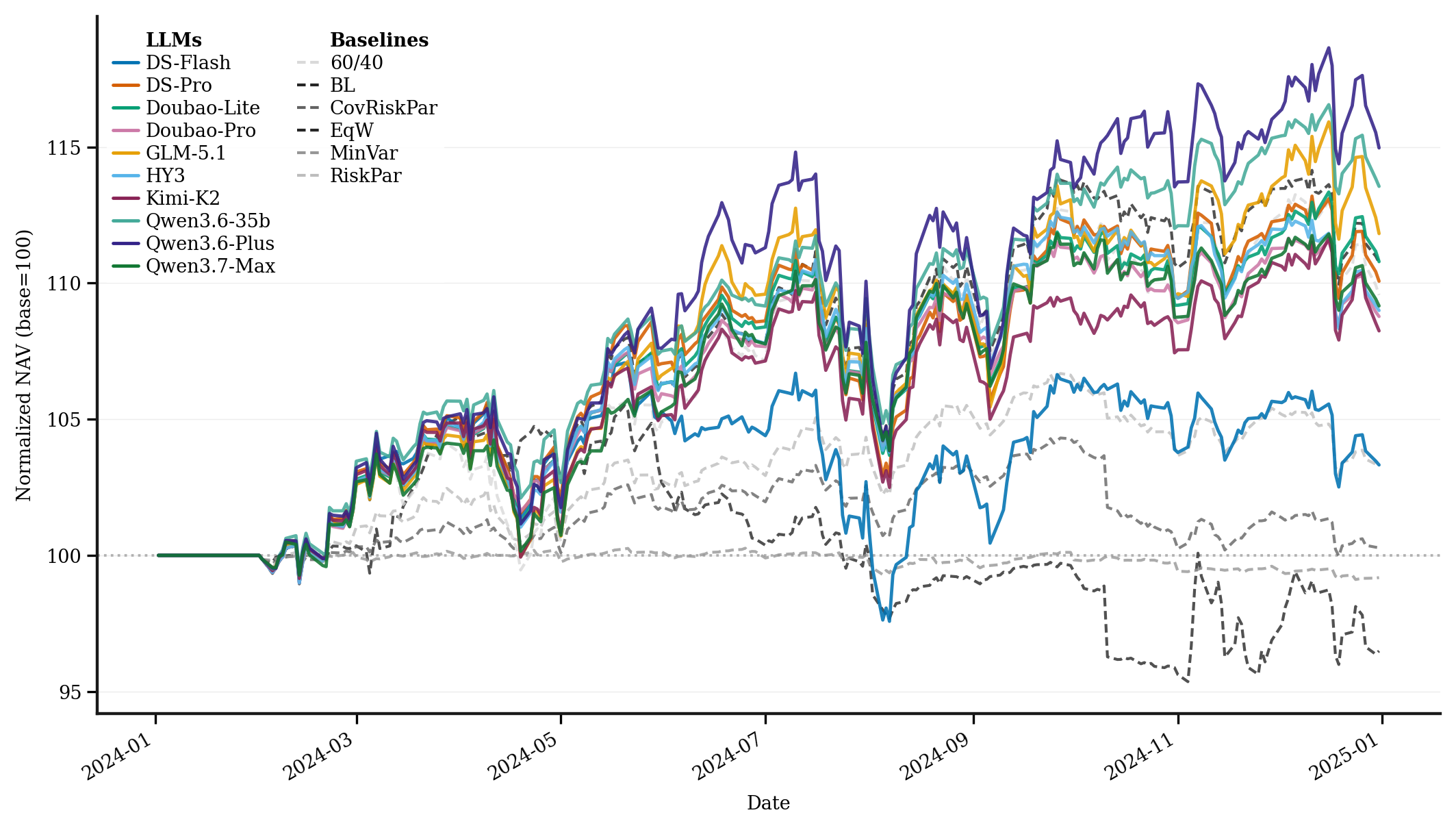}
\caption{Normalized NAV in the 2024 normal window under the conservative profile.}
\label{fig:app_nav_cons}
\end{figure}

\begin{figure}[!h]
\centering
\includegraphics[width=\columnwidth]{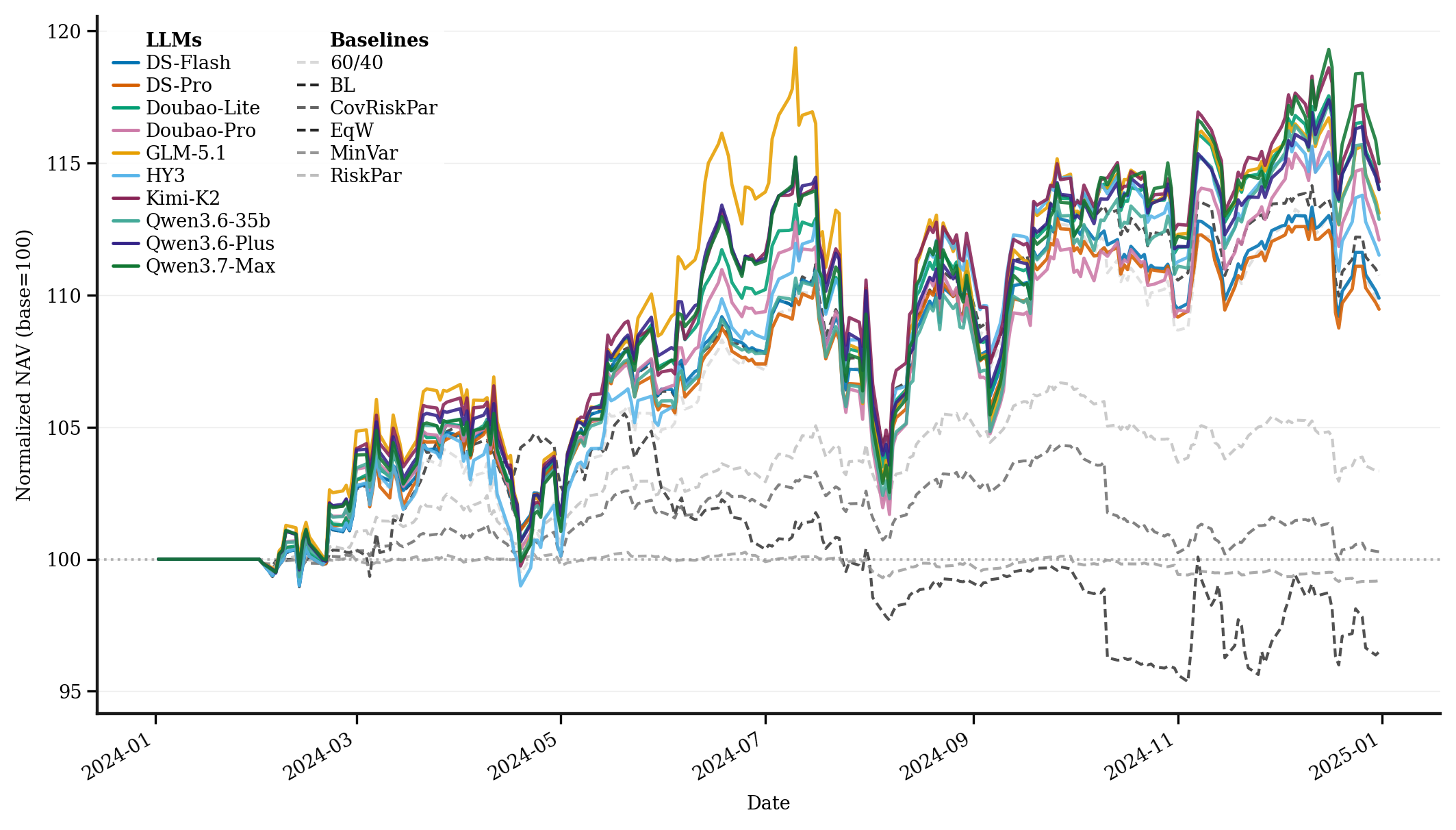}
\caption{Normalized NAV in the 2024 normal window under the balanced profile.}
\label{fig:app_nav_bal}
\end{figure}

\begin{figure}[!h]
\centering
\includegraphics[width=\columnwidth]{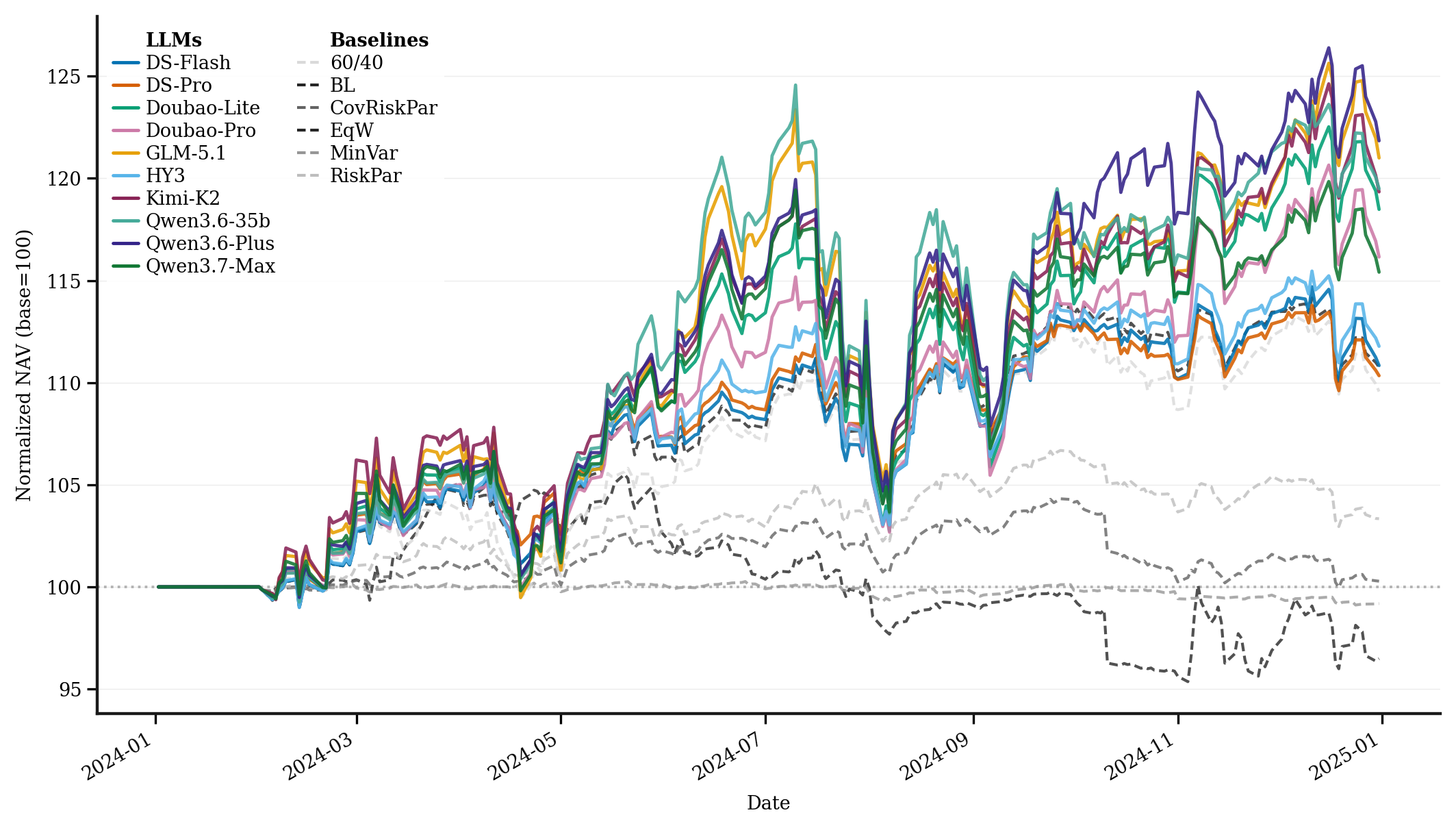}
\caption{Normalized NAV in the 2024 normal window under the aggressive profile.}
\label{fig:app_nav_agg}
\end{figure}

\begin{figure}[!h]
\centering
\includegraphics[width=\columnwidth]{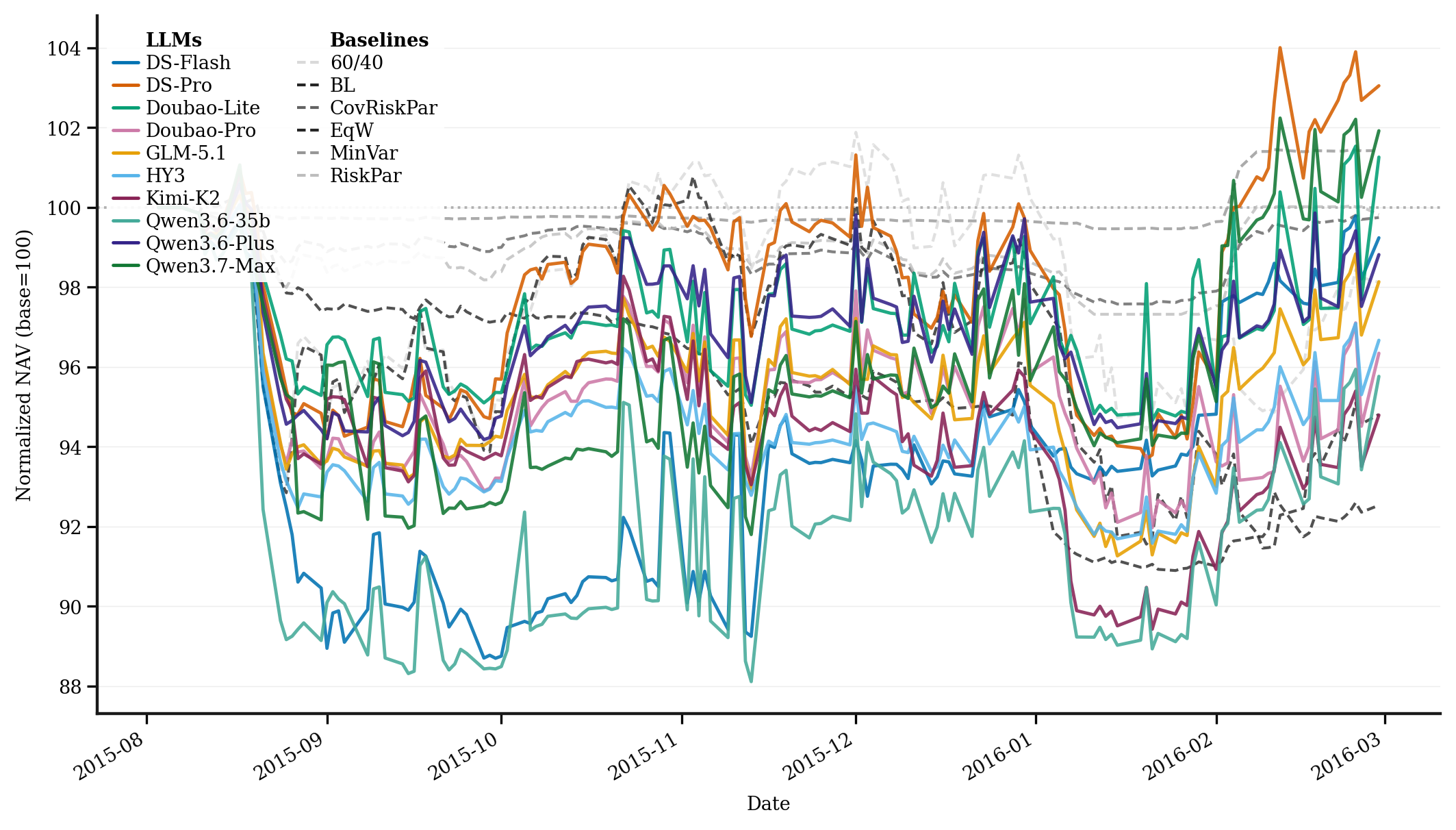}
\caption{Normalized NAV during the 2015 China Shock under the conservative profile.}
\label{fig:app_nav_stress_china}
\end{figure}

\begin{figure}[!h]
\centering
\includegraphics[width=\columnwidth]{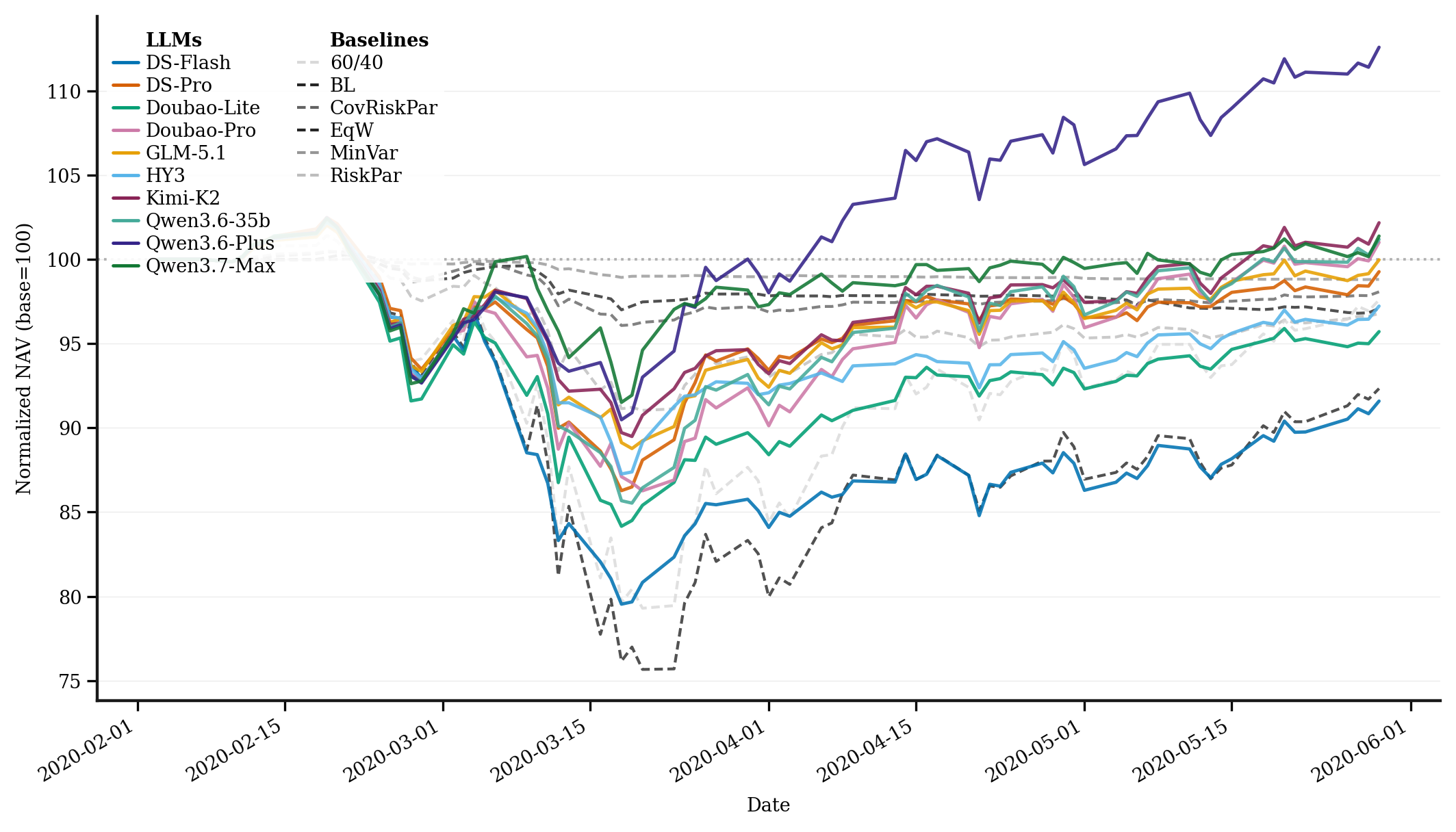}
\caption{Normalized NAV during the 2020 COVID Crash under the balanced profile.}
\label{fig:app_nav_stress_covid}
\end{figure}

Figures~\ref{fig:app_nav_cons}--\ref{fig:app_nav_agg} compare the three profiles in the 2024 normal window. Terminal model NAV ranges from 1.033 to 1.150 for conservative portfolios, 1.095 to 1.150 for balanced portfolios, and 1.103 to 1.218 for aggressive portfolios. Figures~\ref{fig:app_nav_stress_china} and~\ref{fig:app_nav_stress_covid} show the corresponding paths during the 2015 and 2020 stress windows. The trajectories show when portfolios diverge, recover, or retain losses.

\begin{figure*}[t]
\centering
\includegraphics[width=\textwidth]{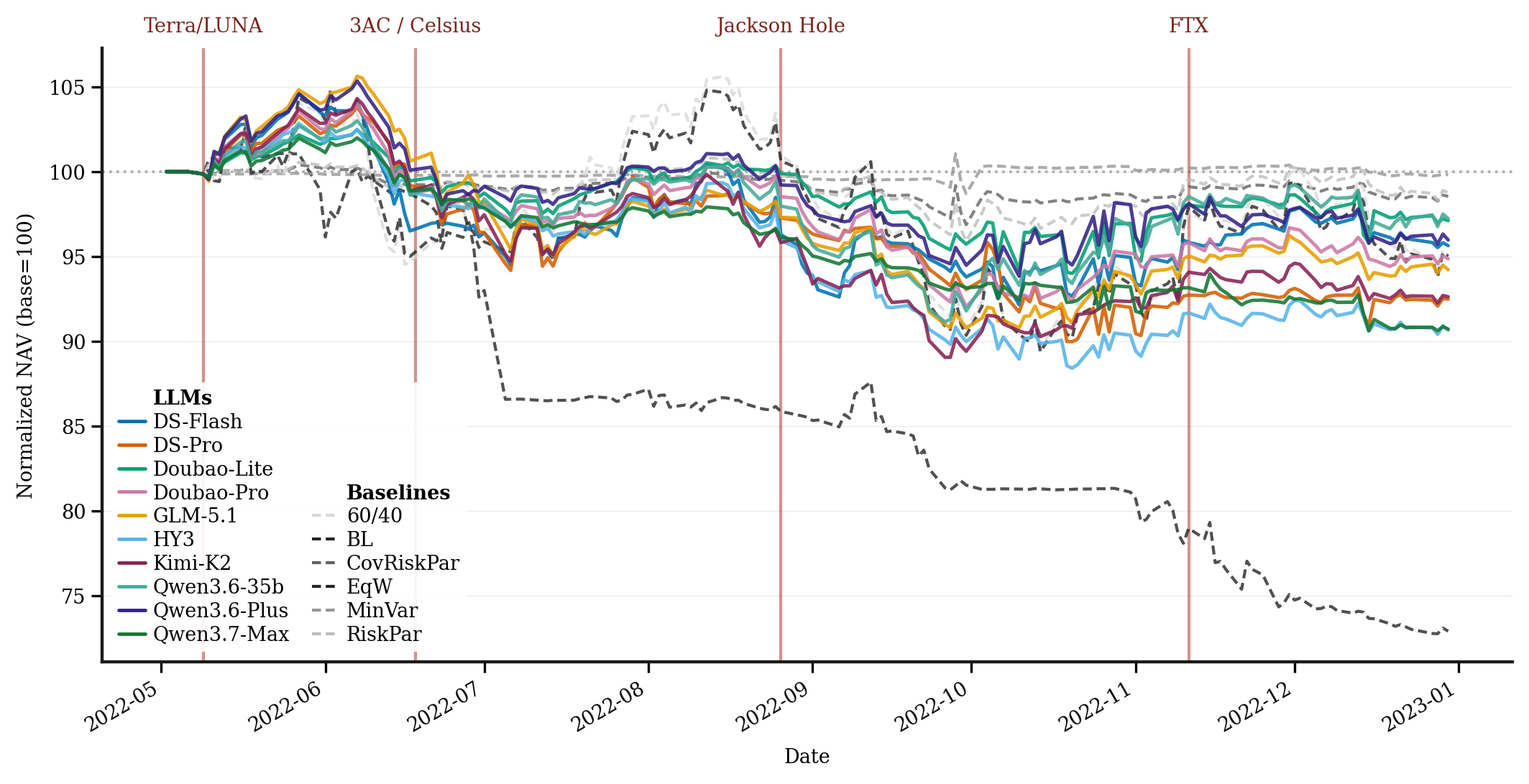}
\caption{Normalized NAV during the 2022 Market Decline under the conservative profile, when no LLM beats EqW. Vertical lines mark events during this period.}
\label{fig:nav_stress_crypto}
\end{figure*}

\paragraph{2022 Market Decline.}
Figure~\ref{fig:nav_stress_crypto} shows the accumulation of losses during the 2022 window. Conservative model portfolios finish with NAV between .907 and .972, below their initial value of one. No model--profile pair beats EqW on Sharpe in this window. The dated event markers provide context for the market path.

\section{Pipeline Diagnostics}
\label{app:pipeline_diagnostics}

\subsection{Stage-Reference Agreement}
\label{app:stage_agreement}

Table~\ref{tab:replay_stages} reports all five stage scores and CEPS for each normal-period model--profile cell. S1/S2 compare outputs with trailing-return rules; S3 measures allocation similarity and class exposure; S4/S5 measure turnover and lookback risk-state agreement. The exposure and outcome columns in Table~\ref{tab:allocation_compliance} assess the corresponding investment requirements directly.
\begin{table*}[t]
\centering
\small
\begin{adjustbox}{max width=\textwidth,center}
\begin{tabular}{llrrrrrr}
\toprule
\textbf{Profile} & \textbf{Model} & \textbf{S1} & \textbf{S2} & \textbf{S3} & \textbf{S4} & \textbf{S5} & \textbf{CEPS} \\
\midrule
\multirow{10}{*}{\textbf{Conservative}} & DS-V4-Flash & 0.742 & 0.315 & 0.298 & 0.320 & 0.022 & 0.253 \\
 & DS-V4-Pro & 0.749 & 0.339 & 0.302 & 0.217 & 0.000 & 0.239 \\
 & Qwen3.6-Plus & 0.774 & 0.491 & 0.310 & 0.153 & 0.136 & 0.292 \\
 & Qwen3.7-Max & 0.746 & 0.385 & 0.291 & 0.151 & 0.000 & 0.232 \\
 & Qwen3.6-35B-A3B & 0.754 & 0.405 & 0.304 & 0.217 & 0.045 & 0.263 \\
 & GLM-5.1 & 0.763 & 0.438 & 0.301 & 0.304 & 0.091 & 0.299 \\
 & DB-2.0-Lite & 0.751 & 0.339 & 0.299 & 0.143 & 0.000 & 0.223 \\
 & DB-2.0-Pro & 0.750 & 0.403 & 0.295 & 0.143 & 0.000 & 0.236 \\
 & HY3-Preview & 0.778 & 0.489 & 0.290 & 0.121 & 0.000 & 0.251 \\
 & Kimi-K2.6 & 0.768 & 0.394 & 0.303 & 0.269 & 0.000 & 0.261 \\
\midrule
\multirow{10}{*}{\textbf{Balanced}} & DS-V4-Flash & 0.774 & 0.321 & 0.290 & 0.172 & 0.000 & 0.226 \\
 & DS-V4-Pro & 0.764 & 0.387 & 0.301 & 0.289 & 0.005 & 0.266 \\
 & Qwen3.6-Plus & 0.793 & 0.531 & 0.298 & 0.209 & 0.091 & 0.300 \\
 & Qwen3.7-Max & 0.774 & 0.451 & 0.297 & 0.158 & 0.182 & 0.292 \\
 & Qwen3.6-35B-A3B & 0.772 & 0.476 & 0.295 & 0.237 & 0.000 & 0.272 \\
 & GLM-5.1 & 0.773 & 0.472 & 0.303 & 0.296 & 0.409 & 0.384 \\
 & DB-2.0-Lite & 0.773 & 0.387 & 0.298 & 0.137 & 0.045 & 0.244 \\
 & DB-2.0-Pro & 0.778 & 0.483 & 0.294 & 0.179 & 0.000 & 0.263 \\
 & HY3-Preview & 0.791 & 0.531 & 0.264 & 0.133 & 0.000 & 0.258 \\
 & Kimi-K2.6 & 0.786 & 0.464 & 0.297 & 0.247 & 0.000 & 0.274 \\
\midrule
\multirow{10}{*}{\textbf{Aggressive}} & DS-V4-Flash & 0.795 & 0.475 & 0.298 & 0.195 & 0.009 & 0.268 \\
 & DS-V4-Pro & 0.783 & 0.463 & 0.303 & 0.250 & 0.014 & 0.278 \\
 & Qwen3.6-Plus & 0.802 & 0.566 & 0.297 & 0.162 & 0.409 & 0.366 \\
 & Qwen3.7-Max & 0.781 & 0.463 & 0.303 & 0.190 & 0.273 & 0.323 \\
 & Qwen3.6-35B-A3B & 0.786 & 0.553 & 0.299 & 0.252 & 0.318 & 0.365 \\
 & GLM-5.1 & 0.779 & 0.510 & 0.295 & 0.310 & 0.409 & 0.393 \\
 & DB-2.0-Lite & 0.787 & 0.478 & 0.300 & 0.138 & 0.364 & 0.331 \\
 & DB-2.0-Pro & 0.787 & 0.494 & 0.300 & 0.168 & 0.273 & 0.325 \\
 & HY3-Preview & 0.807 & 0.574 & 0.292 & 0.132 & 0.000 & 0.274 \\
 & Kimi-K2.6 & 0.790 & 0.514 & 0.293 & 0.199 & 0.136 & 0.304 \\
\bottomrule
\end{tabular}
\end{adjustbox}
\caption{2024 agreement with stage-specific references. S4 measures turnover agreement and S5 risk-state agreement; financial outcomes are reported separately.}
\label{tab:replay_stages}
\end{table*}

\subsection{Lookback and Ex-Post References}
\label{app:reference_perspectives}

Reference allocations depend on the return information used to construct them. We therefore consider two complementary references: a lookback allocation based on information available at the decision date, and an ex-post allocation based on subsequent realized returns. They provide a decision-time view and a hindsight view of the same portfolio decisions.

We compare them on the same 330 decisions in the 2024 normal window, holding model outputs, eligible assets, reference buy signals, and profile constraints fixed. For the ex-post reference, the mean and covariance in Eq.~\ref{eq:app_wstar} use returns after the decision date through the next decision date, or year-end for the final decision. Both references use the same optimization procedure. S3--S5 use the corresponding reference weights; all other scoring definitions, including the lookback risk sample for S5, remain fixed.

GLM-5.1 ranks first under both references, and the same three models occupy the top three places. Across all ten models, the profile-averaged CEPS rankings have Spearman $\rho=.77$, with several other positions changing. Mean S5 rises from .108 to .283, mainly because the reference VaR and drawdown estimates are closer to those of the model portfolios on the unchanged lookback sample. Five of the 33 ex-post references retain the feasible initial allocation under the optimization rule in Appendix~\ref{app:metrics}; these cases account for about half of the mean S5 increase. The two references thus identify the same leading group while showing how reference choice affects individual scores and ranks. The model portfolios' realized returns and risks are unchanged.

\subsection{Allocation Agreement and Exposure Compliance}
\label{app:allocation_compliance}

Table~\ref{tab:allocation_compliance} separates reference-weight similarity, class-exposure score, target-exposure compliance, and realized financial outcomes for all normal-period cells. Exposure compliance checks both profile inequalities at each decision.
\begin{table*}[t]
\centering
\small
\begin{adjustbox}{max width=\textwidth,center}
\begin{tabular}{llrrrrrrrrc}
\toprule
Profile & Model & Ref. & Class & CEPS & Exposure\% & Sharpe & Ret\% & DD\% & Vol\% & Gate \\
\midrule
\multirow{10}{*}{\rotatebox[origin=c]{90}{Conservative}} & DS-V4-Flash & 0.118 & 0.477 & 0.253 & 18.2 & -0.029 & 3.32 & -8.87 & 9.51 & $\times$ \\
 & DS-V4-Pro & 0.117 & 0.487 & 0.239 & 0.0 & 0.616 & 10.07 & -7.50 & 9.87 & $\times$ \\
 & Qwen3.6-Plus & 0.127 & 0.493 & 0.292 & 0.0 & 0.875 & 14.97 & -9.56 & 12.22 & $\times$ \\
 & Qwen3.7-Max & 0.121 & 0.461 & 0.232 & 0.0 & 0.578 & 9.16 & -5.97 & 8.94 & $\times$ \\
 & Qwen3.6-35B-A3B & 0.119 & 0.489 & 0.263 & 0.0 & 0.992 & 13.55 & -6.19 & 9.21 & $\times$ \\
 & GLM-5.1 & 0.115 & 0.486 & 0.299 & 0.0 & 0.707 & 11.82 & -7.87 & 11.00 & $\times$ \\
 & DB-2.0-Lite & 0.123 & 0.476 & 0.223 & 0.0 & 0.740 & 10.79 & -6.46 & 8.98 & $\times$ \\
 & DB-2.0-Pro & 0.112 & 0.477 & 0.236 & 0.0 & 0.571 & 8.77 & -5.73 & 8.34 & $\times$ \\
 & HY3-Preview & 0.115 & 0.464 & 0.251 & 0.0 & 0.547 & 9.00 & -6.38 & 9.21 & $\times$ \\
 & Kimi-K2.6 & 0.119 & 0.487 & 0.261 & 0.0 & 0.445 & 8.25 & -6.59 & 9.94 & $\times$ \\
\midrule
\multirow{10}{*}{\rotatebox[origin=c]{90}{Balanced}} & DS-V4-Flash & 0.109 & 0.471 & 0.226 & 81.8 & 0.609 & 9.89 & -6.56 & 9.68 & $\times$ \\
 & DS-V4-Pro & 0.116 & 0.487 & 0.266 & 54.5 & 0.542 & 9.46 & -6.93 & 10.29 & $\checkmark$ \\
 & Qwen3.6-Plus & 0.112 & 0.484 & 0.300 & 45.5 & 0.798 & 13.99 & -10.11 & 12.35 & $\checkmark$ \\
 & Qwen3.7-Max & 0.109 & 0.484 & 0.292 & 0.0 & 0.783 & 14.97 & -11.02 & 13.95 & $\checkmark$ \\
 & Qwen3.6-35B-A3B & 0.102 & 0.488 & 0.272 & 36.4 & 0.740 & 12.87 & -7.77 & 11.91 & $\checkmark$ \\
 & GLM-5.1 & 0.111 & 0.495 & 0.384 & 0.0 & 0.603 & 13.10 & -14.20 & 15.89 & $\checkmark$ \\
 & DB-2.0-Lite & 0.113 & 0.483 & 0.244 & 9.1 & 0.838 & 14.01 & -8.89 & 11.70 & $\checkmark$ \\
 & DB-2.0-Pro & 0.107 & 0.481 & 0.263 & 45.5 & 0.662 & 12.09 & -9.83 & 12.36 & $\checkmark$ \\
 & HY3-Preview & 0.092 & 0.437 & 0.258 & 9.1 & 0.650 & 11.52 & -8.05 & 11.66 & $\checkmark$ \\
 & Kimi-K2.6 & 0.101 & 0.494 & 0.274 & 0.0 & 0.769 & 14.29 & -9.34 & 13.32 & $\checkmark$ \\
\midrule
\multirow{10}{*}{\rotatebox[origin=c]{90}{Aggressive}} & DS-V4-Flash & 0.115 & 0.481 & 0.268 & 90.9 & 0.640 & 10.84 & -7.40 & 10.72 & $\checkmark$ \\
 & DS-V4-Pro & 0.121 & 0.485 & 0.278 & 100.0 & 0.653 & 10.34 & -6.63 & 9.65 & $\checkmark$ \\
 & Qwen3.6-Plus & 0.099 & 0.496 & 0.366 & 63.6 & 1.034 & 21.84 & -13.05 & 16.57 & $\checkmark$ \\
 & Qwen3.7-Max & 0.109 & 0.497 & 0.323 & 100.0 & 0.733 & 15.41 & -13.20 & 15.79 & $\checkmark$ \\
 & Qwen3.6-35B-A3B & 0.092 & 0.505 & 0.365 & 54.5 & 0.845 & 19.49 & -16.83 & 18.33 & $\checkmark$ \\
 & GLM-5.1 & 0.089 & 0.502 & 0.393 & 63.6 & 0.930 & 21.00 & -15.13 & 17.92 & $\checkmark$ \\
 & DB-2.0-Lite & 0.104 & 0.496 & 0.331 & 100.0 & 0.916 & 18.48 & -12.76 & 15.45 & $\checkmark$ \\
 & DB-2.0-Pro & 0.105 & 0.496 & 0.325 & 100.0 & 0.867 & 16.15 & -10.85 & 13.75 & $\checkmark$ \\
 & HY3-Preview & 0.106 & 0.477 & 0.274 & 63.6 & 0.673 & 11.78 & -8.74 & 11.61 & $\checkmark$ \\
 & Kimi-K2.6 & 0.093 & 0.492 & 0.304 & 18.2 & 0.890 & 19.34 & -13.08 & 16.99 & $\checkmark$ \\
\bottomrule
\end{tabular}
\end{adjustbox}
\caption{2024 results under common execution. Ref. is similarity to the profile-constrained lookback reference; Class is the class-exposure correlation score. Exposure is the percentage of decisions meeting both exposure limits. CEPS combines five-stage scores and penalizes downstream score drops; return, drawdown (DD), and volatility measure financial outcomes. Gate requires all three stress-window drawdowns to meet the profile tolerance. Models follow a fixed order.}
\label{tab:allocation_compliance}
\end{table*}

\subsection{Allocation and Turnover}
\label{app:active_tilt}

The 330 normal allocations have mean total-variation distance .306 from EqW and .890 from the profile-constrained reference. The mean decision-wise ratio of model to reference turnover is .203, using the same pre-trade weights for both targets. Reference allocations hold 8--14 securities with maximum weights .24--.72. Model mean HHI is .0207; its reciprocal is 48.3, while the mean of the individual reciprocal HHIs is 55.9. EqW over the eligible securities has HHI $1/n$ and $n$ effective holdings. Table~\ref{tab:app_active_tilt} reports these allocation diagnostics and their associations with realized outcomes.

\begin{table}[!h]
\centering
\small
\begin{adjustbox}{max width=\columnwidth,center}
\begin{tabular}{lrrr}
\toprule
\textbf{Quantity} & \textbf{Estimate} & \textbf{95\% CI} & \textbf{$q$} \\
\midrule
Mean distance to EqW & .306 & --- & --- \\
Mean distance to S3 reference & .890 & --- & --- \\
Mean S3 weight similarity & .110 & --- & --- \\
Mean S3 correlation score & .484 & --- & --- \\
Mean S3 score & .297 & --- & --- \\
Mean HHI & .0207 & --- & --- \\
Mean turnover ratio & .203 & --- & --- \\
$d_\text{EqW}$ vs. $\Delta$Sharpe & $.514$ & [$.209$, $.736$] & $.004$ \\
$d_\text{EqW}$ vs. volatility & $.655$ & [$.389$, $.803$] & $<.001$ \\
$d_\text{EqW}$ vs. drawdown magnitude & $.612$ & [$.378$, $.776$] & $<.001$ \\
\bottomrule
\end{tabular}
\end{adjustbox}
\caption{Active-tilt diagnostics over 30 normal-period model--profile cells: Spearman correlations with model-cluster bootstrap intervals and BH-FDR $q$-values.}
\label{tab:app_active_tilt}
\end{table}

\subsection{Allocation Changes by Profile}
\label{app:profile_allocations}

Matching each model and decision date, conservative-to-aggressive risk exposure increases by .175 and bond-plus-cash exposure falls by .162; distance from EqW increases by .070. Across the ten model-level financial pairs, volatility rises by .050 on average (nine positive changes), drawdown magnitude by .047 (eight positive changes), and Sharpe by .214 (nine positive changes). The mean aggressive-minus-conservative Sharpe differences are $-.423$ in 2015--16, $+.935$ in 2020, $-.039$ in 2022, and $+.214$ in 2024. Table~\ref{tab:app_profile_changes} gives the paired changes and model-cluster intervals. Risk-asset exposure is the equity-plus-crypto weight used to assess cap compliance; bond-plus-cash exposure includes uninvested cash. The exposure response coexists with period-dependent financial outcomes.

\begin{table}[!h]
\centering
\small
\begin{adjustbox}{max width=\columnwidth,center}
\begin{tabular}{lrrrr}
\toprule
\textbf{Agg.--Cons. change} & \textbf{Mean} & \textbf{95\% CI} & \textbf{+} & \textbf{$-$} \\
\midrule
Equity+crypto exposure & $.175$ & [$.124$, $.221$] & 103 & 6 \\
Bond+cash exposure & $-.162$ & [$-.205$, $-.117$] & 5 & 104 \\
Distance from EqW & $.070$ & [$.026$, $.113$] & 89 & 20 \\
HHI & $.0065$ & [$.0027$, $.0106$] & 91 & 18 \\
Volatility (model means) & $.0496$ & [$.0317$, $.0659$] & 9 & 1 \\
DD magnitude (model means) & $.0465$ & [$.0237$, $.0679$] & 8 & 2 \\
Sharpe (model means) & $.2140$ & [$.0886$, $.3491$] & 9 & 1 \\
\bottomrule
\end{tabular}
\end{adjustbox}
\caption{Matched conservative-to-aggressive allocation and outcome changes. Allocation rows contain 110 model--date pairs; outcome rows contain ten paired model means. The last two columns omit zero counts.}
\label{tab:app_profile_changes}
\end{table}

\subsection{CEPS Parameter Sensitivity}
\label{app:parameter_sensitivity}

\begin{figure*}[t]
\centering
\includegraphics[width=\textwidth]{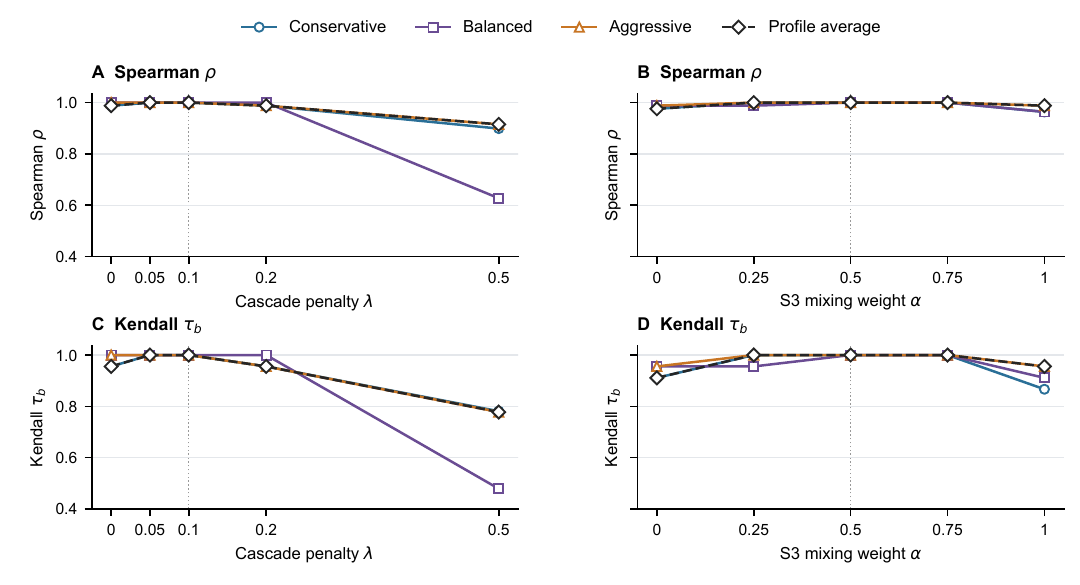}
\caption{CEPS ranking sensitivity over 330 normal-period decisions. Each curve compares ten model means with the corresponding default ranking. The profile-average curve gives each profile equal weight. Dotted vertical lines mark the default values. The other parameter is fixed while each parameter varies.}
\label{fig:ceps_parameter_sensitivity}
\end{figure*}

We vary $\lambda\in\{0,.05,.1,.2,.5\}$ and $\alpha\in\{0,.25,.5,.75,1\}$, changing one parameter at a time while holding the other at its default ($\lambda=.1$, $\alpha=.5$). We recompute CEPS for all 330 normal-period decisions, then average by model within each profile and across the three profiles. Figure~\ref{fig:ceps_parameter_sensitivity} compares each ten-model ranking with its default ranking using Spearman $\rho$ and Kendall $\tau_b$.

Rankings change little across the tested settings with $\lambda\leq.2$: within-profile Spearman correlations range from .988 to 1.000. Across the full $\alpha$ sweep, they range from .964 to 1.000. The three-profile mean ranking is unchanged at $\lambda=.05$ and at $\alpha=.25$ or $.75$. At $\lambda=.5$, clipping sets 253/330 decision scores to zero. In the balanced profile, 93/110 scores become zero, creating tied model means and making the ranking depend on the few remaining positive scores. These results support ranking stability around the default penalty and across the tested S3 weights.

\subsection{Stage Scores and Financial Associations}
\label{app:process_outcome}

Table~\ref{tab:app_process_outcome} links CEPS to realized outcomes within each profile. Under the aggressive profile, higher CEPS is associated with higher Sharpe ($\rho=.85$, 95\% CI $[.41,1.00]$), connecting stage agreement to risk-adjusted performance in this setting. The association is weaker for conservative ($.21$) and balanced ($.09$) portfolios, whose intervals include zero. Positive associations with volatility and drawdown magnitude show that the financial interpretation of stage agreement also depends on the risk taken by the portfolio.
\begin{table*}[t]
\centering
\small
\begin{adjustbox}{max width=\textwidth,center}
\begin{tabular}{lrrr}
\toprule
Profile & Sharpe & Volatility & Drawdown magnitude \\
\midrule
Conservative & $0.21$ [$-0.66$, $0.65$] & $0.82$ [$0.37$, $0.96$] & $0.58$ [$-0.08$, $0.86$] \\
Balanced & $0.09$ [$-0.61$, $1.00$] & $0.79$ [$0.31$, $1.00$] & $0.76$ [$0.14$, $0.99$] \\
Aggressive & $0.85$ [$0.41$, $1.00$] & $0.76$ [$0.21$, $0.96$] & $0.72$ [$0.19$, $0.96$] \\
\bottomrule
\end{tabular}
\end{adjustbox}
\caption{CEPS--financial outcome associations by profile (ten models). Paired model-bootstrap intervals.}
\label{tab:app_process_outcome}
\end{table*}

Table~\ref{tab:app_process_pooled} reports pooled associations and stress diagnostics. Across profiles, CEPS correlates positively with Sharpe ($\rho=.587$). Under the balanced profile, higher normal-period CEPS is also associated with a lower worst-window drawdown relative to the profile tolerance ($\rho=-.818$, 95\% CI $[-.975,-.346]$). These profile-specific associations support assessing stage agreement alongside risk-adjusted returns and stress resilience.
\begin{table}[!h]
\centering
\small
\begin{adjustbox}{max width=\columnwidth,center}
\begin{tabular}{llrr}
\toprule
\textbf{Scope} & \textbf{Statistic} & \textbf{Estimate} & \textbf{95\% CI} \\
\midrule
All profiles & Sharpe ($\rho$) & $.587$ & [$.337$, $.809$] \\
All profiles & Volatility ($\rho$) & $.855$ & [$.683$, $.921$] \\
All profiles & DD magnitude ($\rho$) & $.835$ & [$.618$, $.921$] \\
Aggressive & Stress ratio ($\rho$) & $-.600$ & [$-.925$, $.038$] \\
Balanced & Stress ratio ($\rho$) & $-.818$ & [$-.975$, $-.346$] \\
Conservative & Stress ratio ($\rho$) & $.333$ & [$-.459$, $.887$] \\
\bottomrule
\end{tabular}
\end{adjustbox}
\caption{Pooled outcome associations and profile-specific stress diagnostics. Stress ratio is the largest scenario drawdown divided by the profile tolerance. Intervals resample models.}
\label{tab:app_process_pooled}
\end{table}

\paragraph{Full and core CEPS.}
Table~\ref{tab:app_ceps_core} compares full S1--S5 CEPS with S1--S3 core CEPS, averaging across profiles. Spearman $\rho$ is .66. GLM-5.1 moves from core rank~4 to full rank~1, while HY3-Preview moves from 2 to 8. S4 turnover agreement and S5 risk-state agreement change the ordering of models with similar upstream scores.

\begin{table}[!h]
\centering
\small
\begin{adjustbox}{max width=\columnwidth,center}
\begin{tabular}{lcc cc}
\toprule
\textbf{Model} & \textbf{Full} & \textbf{R} & \textbf{Core} & \textbf{R} \\
\midrule
GLM-5.1 & .359 & 1 & .468 & 4 \\
Qwen3.6-Plus & .319 & 2 & .491 & 1 \\
Qwen3.6-35B-A3B & .300 & 3 & .468 & 3 \\
Qwen3.7-Max & .282 & 4 & .451 & 7 \\
Kimi-K2.6 & .280 & 5 & .463 & 5 \\
DB-2.0-Pro & .275 & 6 & .462 & 6 \\
DB-2.0-Lite & .266 & 7 & .442 & 8 \\
HY3-Preview & .261 & 8 & .484 & 2 \\
DS-V4-Pro & .261 & 9 & .441 & 9 \\
DS-V4-Flash & .249 & 10 & .428 & 10 \\
\bottomrule
\end{tabular}
\end{adjustbox}
\caption{CEPS-full (S1--S5) vs. CEPS-core (S1--S3): ranks (R), three-profile means, Spearman $\rho=.66$.}
\label{tab:app_ceps_core}
\end{table}

\paragraph{Stress-gate threshold.}
Rescaling each profile tolerance by $.8$, $.9$, $1.1$, and $1.2$ changes 27 of 360 scenario-level tolerance checks (7.5\%). Most checks remain unchanged.

\subsection{Paired Stage Interventions}
\label{app:paired_details}

\paragraph{Protocol and scoring.}
\label{app:agentic_s4_s5}
Callable S4 selects order type, urgency, and slippage limits for a bound S3 target. Its score weights legality, target tracking, direction consistency, and cost awareness by .20, .35, .25, and .20; the runtime binds the target quantities and directions. Callable S5 selects a risk action from executed weights and risk statistics. Its reference chooses a 0.8 scale-down for a VaR or drawdown breach, equal-weight rebalancing for a drift breach, and hold otherwise. These scores assess plan choices under the bound task interfaces.

The balanced-profile intervention diagnostic uses callable S4/S5 plan tasks, so its downstream changes measure responses in generated execution and risk plans. The main ranking evaluates turnover and lookback risk-state agreement. The normal diagnostic covers 99 securities. The financial replay uses the tradable set defined in Appendix~\ref{app:evaluation_details}. The replaced stage receives its reference output and scores one; downstream scores use the model outputs for those stages.

The paired analysis contains 300 intervention episodes: 110 normal and 190 stress. Replacing each stage with its reference output yields 1,500 branches and 4,500 stage-score changes. All 3,000 calls to subsequent model-dependent stages meet the required output formats. Confidence intervals for the complete matrices use 10,000 circular moving-block bootstrap samples within model--window units, and $p$-values use Benjamini--Hochberg FDR control within each split. For this paired intervention analysis, S3 scores are recomputed from the stage allocations. Ten terminal-period S3 scores use only the finite correlation pairs, with the corresponding weights renormalized.

\begin{figure*}[!ht]
\centering
\includegraphics[width=.92\textwidth]{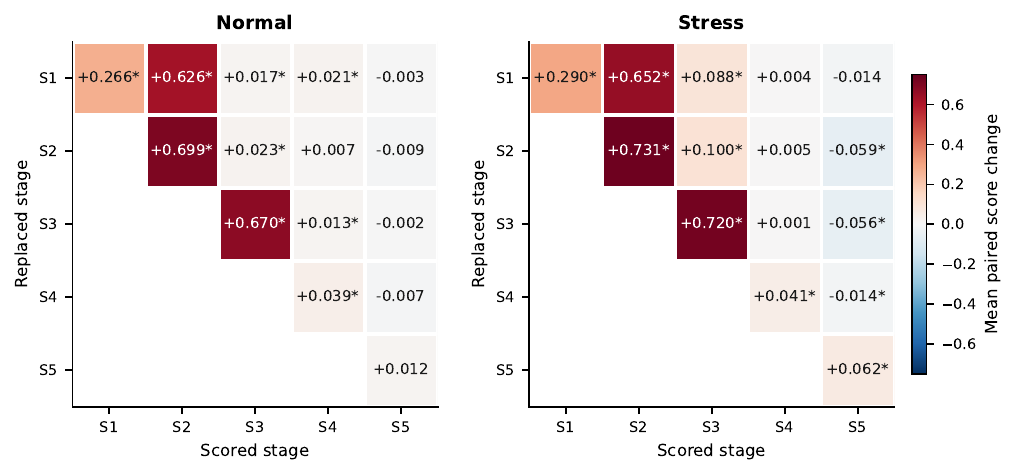}
\caption{Paired influence matrix for normal and stress episodes. Each cell is the measured-stage score change after replacing the indicated stage with its reference output. Asterisks mark BH-FDR $q<0.05$; black outlines identify the stage pairs discussed in the main text.}
\label{fig:app_paired_heatmap}
\end{figure*}

Figure~\ref{fig:app_paired_heatmap} shows the stage-specific changes. Replacing S1 raises S2 by $+.626$ in normal episodes, yet its mean effect on S3 is only $+.017$ and is negative in 40 episodes. Replacing S2 raises S3 by $+.023$ in normal episodes, with 39 negative cases. Under stress, replacing S2 or S3 lowers S5 by $-.059$ and $-.056$, respectively. Most changes at later measured stages are zero. \textbf{Reference replacements improve some downstream stage scores, with effects that vary by stage and market condition.}

Tables~\ref{tab:app_influence_normal} and~\ref{tab:app_influence_stress} report the complete cell-level means, confidence intervals, sign counts, and FDR-adjusted tests for the normal and stress episodes.

\begin{table*}[!ht]
\centering
\small
\begin{adjustbox}{max width=\textwidth,center}
\begin{tabular}{lrrrrrrr}
\toprule
\textbf{Cell} & \textbf{Mean} & \textbf{95\% CI} & \textbf{+} & \textbf{$-$} & \textbf{0} & \textbf{$p$} & \textbf{$q$} \\
\midrule
S1$\rightarrow$S1 & +0.266\textsuperscript{*} & [+0.258, +0.274] & 110 & 0 & 0 & $<.001$ & $<.001$ \\
S1$\rightarrow$S2 & +0.626\textsuperscript{*} & [+0.607, +0.644] & 110 & 0 & 0 & $<.001$ & $<.001$ \\
S1$\rightarrow$S3 & +0.017\textsuperscript{*} & [+0.004, +0.032] & 70 & 40 & 0 & .011 & .019 \\
S1$\rightarrow$S4 & +0.021\textsuperscript{*} & [+0.012, +0.031] & 25 & 4 & 81 & $<.001$ & $<.001$ \\
S1$\rightarrow$S5 & -0.003 & [-0.022, +0.014] & 3 & 6 & 101 & .732 & .785 \\
S2$\rightarrow$S2 & +0.699\textsuperscript{*} & [+0.684, +0.713] & 110 & 0 & 0 & $<.001$ & $<.001$ \\
S2$\rightarrow$S3 & +0.023\textsuperscript{*} & [+0.009, +0.037] & 71 & 39 & 0 & .001 & .002 \\
S2$\rightarrow$S4 & +0.007 & [-0.002, +0.017] & 24 & 13 & 73 & .098 & .134 \\
S2$\rightarrow$S5 & -0.009 & [-0.026, +0.007] & 2 & 9 & 99 & .277 & .324 \\
S3$\rightarrow$S3 & +0.670\textsuperscript{*} & [+0.659, +0.681] & 110 & 0 & 0 & $<.001$ & $<.001$ \\
S3$\rightarrow$S4 & +0.013\textsuperscript{*} & [+0.003, +0.023] & 24 & 9 & 77 & .009 & .015 \\
S3$\rightarrow$S5 & -0.002 & [-0.015, +0.013] & 3 & 4 & 103 & .812 & .812 \\
S4$\rightarrow$S4 & +0.039\textsuperscript{*} & [+0.031, +0.047] & 34 & 0 & 76 & $<.001$ & $<.001$ \\
S4$\rightarrow$S5 & -0.007 & [-0.021, +0.006] & 2 & 4 & 104 & .280 & .324 \\
S5$\rightarrow$S5 & +0.012 & [+0.003, +0.022] & 3 & 0 & 107 & .040 & .061 \\
\bottomrule
\end{tabular}
\end{adjustbox}
\caption{Full normal paired influence matrix. Asterisks mark Benjamini--Hochberg $q<0.05$. Sign counts are episode-level paired changes; $n=110$ for every cell.}
\label{tab:app_influence_normal}
\end{table*}

\begin{table*}[!ht]
\centering
\small
\begin{adjustbox}{max width=\textwidth,center}
\begin{tabular}{lrrrrrrr}
\toprule
\textbf{Cell} & \textbf{Mean} & \textbf{95\% CI} & \textbf{+} & \textbf{$-$} & \textbf{0} & \textbf{$p$} & \textbf{$q$} \\
\midrule
S1$\rightarrow$S1 & +0.290\textsuperscript{*} & [+0.283, +0.298] & 190 & 0 & 0 & $<.001$ & $<.001$ \\
S1$\rightarrow$S2 & +0.652\textsuperscript{*} & [+0.637, +0.668] & 190 & 0 & 0 & $<.001$ & $<.001$ \\
S1$\rightarrow$S3 & +0.088\textsuperscript{*} & [+0.081, +0.096] & 172 & 18 & 0 & $<.001$ & $<.001$ \\
S1$\rightarrow$S4 & +0.004 & [-0.004, +0.012] & 34 & 25 & 131 & .361 & .387 \\
S1$\rightarrow$S5 & -0.014 & [-0.040, +0.011] & 28 & 21 & 141 & .283 & .327 \\
S2$\rightarrow$S2 & +0.731\textsuperscript{*} & [+0.722, +0.740] & 190 & 0 & 0 & $<.001$ & $<.001$ \\
S2$\rightarrow$S3 & +0.100\textsuperscript{*} & [+0.093, +0.108] & 175 & 15 & 0 & $<.001$ & $<.001$ \\
S2$\rightarrow$S4 & +0.005 & [-0.003, +0.012] & 30 & 25 & 135 & .233 & .291 \\
S2$\rightarrow$S5 & -0.059\textsuperscript{*} & [-0.087, -0.031] & 22 & 37 & 131 & $<.001$ & $<.001$ \\
S3$\rightarrow$S3 & +0.720\textsuperscript{*} & [+0.714, +0.727] & 190 & 0 & 0 & $<.001$ & $<.001$ \\
S3$\rightarrow$S4 & +0.001 & [-0.008, +0.010] & 32 & 29 & 129 & .873 & .873 \\
S3$\rightarrow$S5 & -0.056\textsuperscript{*} & [-0.079, -0.032] & 23 & 45 & 122 & $<.001$ & $<.001$ \\
S4$\rightarrow$S4 & +0.041\textsuperscript{*} & [+0.037, +0.046] & 55 & 0 & 135 & $<.001$ & $<.001$ \\
S4$\rightarrow$S5 & -0.014\textsuperscript{*} & [-0.025, -0.003] & 9 & 9 & 172 & .017 & .023 \\
S5$\rightarrow$S5 & +0.062\textsuperscript{*} & [+0.044, +0.080] & 35 & 0 & 155 & $<.001$ & $<.001$ \\
\bottomrule
\end{tabular}
\end{adjustbox}
\caption{Full stress paired influence matrix. Asterisks mark Benjamini--Hochberg $q<0.05$. Sign counts are episode-level paired changes; $n=190$ for every cell.}
\label{tab:app_influence_stress}
\end{table*}

\paragraph{Callable-stage results.}
Table~\ref{tab:app_agentic_s4_s5} reports plan scores, outcome scores, and their components over 330 normal and 570 stress episodes across all three profiles. Runtime binding fixes S4 legality, target tracking, and direction, which carry 80\% of the plan score; model variation is concentrated in cost awareness. Under stress, S5 action choice falls from .927 to .884 and adjustment scale from .917 to .857. These experiments evaluate model-selected execution settings and risk actions under fixed order and portfolio inputs.

\begin{table*}[!ht]
\centering
\small
\caption{Callable S4/S5 plan and outcome scores over 330 normal and 570 stress episodes across all three profiles.}
\label{tab:app_agentic_s4_s5}
\begin{adjustbox}{max width=\textwidth,center}
\begin{tabular}{llrr}
\toprule
\textbf{Stage} & \textbf{Score or subscore} & \textbf{Normal} & \textbf{Stress} \\
\midrule
S4 & Plan score & .954 & .959 \\
S4 & Outcome score & .985 & .963 \\
S4 & Order legality / target tracking & 1.000 / 1.000 & 1.000 / 1.000 \\
S4 & Direction consistency / cost awareness & .999 / .774 & .999 / .794 \\
S4 & Shortfall / cost / turnover / filled-weight match & .979 / .979 / .991 / .992 & .948 / .948 / .975 / .980 \\
\midrule
S5 & Plan score & .952 & .918 \\
S5 & Outcome score & .959 & .916 \\
S5 & Alert identification / action choice & .976 / .927 & .961 / .884 \\
S5 & Corrective compliance / adjustment scale & .972 / .917 & .943 / .857 \\
S5 & CVaR / drawdown / violation / period return & .983 / .976 / .927 / .950 & .945 / .945 / .882 / .891 \\
\bottomrule
\end{tabular}
\end{adjustbox}
\end{table*}

\paragraph{Downstream gains and representative cases.}
Table~\ref{tab:repair_signs} excludes the directly replaced stage and retains positive, unchanged, and negative downstream means. S1 replacement improves the downstream mean in 109/110 normal and 186/190 stress episodes; later replacements have more mixed effects. To show what the measured changes mean, Table~\ref{tab:repair_cases} selects the normal-period branch nearest the median within each sign group over S1--S3 replacements. Selection uses the same fixed rule for all models.

\begin{table}[t]
\centering
\small
\begin{adjustbox}{max width=\columnwidth}
\begin{tabular}{llrrrrr}
\toprule
Split & Replaced stage & $n$ & $+$ & $0$ & $-$ & Mean \\
\midrule
Normal & S1 & 110 & 109 & 0 & 1 & $0.165$ \\
Normal & S2 & 110 & 68 & 0 & 42 & $0.007$ \\
Normal & S3 & 110 & 24 & 73 & 13 & $0.006$ \\
Stress & S1 & 190 & 186 & 0 & 4 & $0.181$ \\
Stress & S2 & 190 & 139 & 0 & 51 & $0.016$ \\
Stress & S3 & 190 & 45 & 80 & 65 & $-0.042$ \\
\bottomrule
\end{tabular}
\end{adjustbox}
\caption{Mean score change strictly after the replaced stage in the paired diagnostic. Counts classify each branch's downstream mean; the direct replacement gain is excluded. Zero means unchanged measured scores, not necessarily unchanged decisions.}
\label{tab:repair_signs}
\end{table}

\begin{table*}[t]
\centering
\small
\begin{adjustbox}{max width=\textwidth}
\begin{tabular}{llllrrrrrr}
\toprule
Group & Model & Date & Replaced stage & $\Delta$S1 & $\Delta$S2 & $\Delta$S3 & $\Delta$S4 & $\Delta$S5 & Downstream \\
\midrule
Improvement & HY3-Preview & 2024-09-02 & S1 & $0.270$ & $0.566$ & $-0.116$ & $0.100$ & $0.000$ & $0.138$ \\
Unchanged & DS-V4-Flash & 2024-04-01 & S3 & $0.000$ & $0.000$ & $0.696$ & $0.000$ & $0.000$ & $0.000$ \\
Deterioration & DS-V4-Flash & 2024-10-01 & S2 & $0.000$ & $0.747$ & $0.017$ & $-0.100$ & $0.000$ & $-0.028$ \\
\bottomrule
\end{tabular}
\end{adjustbox}
\caption{Median normal-period cases from positive, zero, and negative downstream-mean groups over S1--S3 replacements. Ties are resolved by model, date, then stage. Columns report score changes; upstream stages are unchanged. These callable S4/S5 diagnostics are separate from the ranked pipeline.}
\label{tab:repair_cases}
\end{table*}

For HY3-Preview on September 2, replacing S1 raises S2 by .566 and changes S4 from a market to a limit order, gaining .100, despite a .116 decline at S3. Its S5 action changes from rebalance to hold while remaining correct for the respective input state. For DS-V4-Flash on April 1, replacing S3 gains .696 locally, but both downstream plan scores remain at their ceiling; the order type remains limit and the risk action remains rebalance. For the same model on October 1, replacing S2 gains .747 locally and .017 at S3, but S4 changes from limit to market and loses .100; S5 changes from hold to rebalance without changing its score. These cases show why both stage scores and action traces matter: changed decisions need not change scores, and large local score gains need not raise the downstream mean.

\subsection{Within-Class Correlation Risk}
\label{app:correlation_risk}

\paragraph{Decisions and covariance estimates.}
We use all 330 normal-period allocations: ten models, three profiles, and eleven monthly decisions from February through December 2024. We use the financial replay's eligible securities, with excluded weight held in uninvested cash. For each security, we take its last 60 available return observations in the 90 calendar days up to the decision date. The primary sample covariance uses dates observed for all eligible securities, giving each matched pair the same positive-semidefinite matrix. Annualized estimated volatility is $\hat\sigma(\mathbf w)=\sqrt{252\mathbf w^\top\hat\Sigma\mathbf w}$. Uninvested cash has zero variance.

\paragraph{Matched allocation.}
Let $H_c=\{i:c(i)=c,\ w_i>10^{-10}\}$ contain the held securities in class $c$, and $W_c^{\mathrm{sec}}=\sum_{i\in H_c}w_i$ their combined weight. The control assigns $\widetilde w_i=W_c^{\mathrm{sec}}/|H_c|$ for $i\in H_c$, with all other security weights zero and uninvested cash unchanged. This preserves selected securities and total class exposure, including the cash-equivalent sleeve. The relative volatility reduction for decision $e$ is
\begin{equation}
 r_e=1-\hat\sigma(\widetilde{\mathbf w}_e)/\hat\sigma(\mathbf w_e).
\end{equation}
We report the arithmetic mean across 330 pairs and a 95\% interval from 10,000 model-cluster bootstrap samples. The uniform weights are determined by the held securities and their class totals.

\paragraph{Variance decomposition.}
With $c(i)$ denoting security $i$'s class,
\begin{align}
 V(\mathbf w)&=D(\mathbf w)+I(\mathbf w)+C(\mathbf w),\\
 D(\mathbf w)&=\sum_iw_i^2\hat\Sigma_{ii},\\
 I(\mathbf w)&=2\!\!\sum_{\substack{i<j\\c(i)=c(j)}}w_iw_j\hat\Sigma_{ij},\\
 C(\mathbf w)&=2\!\!\sum_{\substack{i<j\\c(i)\ne c(j)}}w_iw_j\hat\Sigma_{ij}.
\end{align}
These terms represent individual-security variance, within-class covariance, and cross-class covariance. Taking model minus control and averaging gives $\overline{\Delta V}=1.3599\times10^{-5}$, comprising $1.0386\times10^{-5}$ from $I$, $3.2017\times10^{-6}$ from $D$, and $1.0762\times10^{-8}$ from $C$. Within-class covariance accounts for \WithinCovarianceShare\% of the mean difference, individual-security variance for 23.5\%, and cross-class terms for 0.1\%.

For $s_i=\sqrt{\hat\Sigma_{ii}}$, volatility-weighted pairwise correlation is
\begin{equation}
 \bar\rho(\mathbf w)=\frac{\sum_{i<j}w_iw_js_is_j\hat\rho_{ij}}{\sum_{i<j}w_iw_js_is_j}.
\end{equation}
Unlike a class-average correlation, this quantity changes with the allocation among correlated securities inside a class. Its mean model-minus-control difference is .033 (95\% model-cluster interval [.021, .046]), positive in 289/330 decisions.

\paragraph{Results and estimator sensitivity.}
Mean volatility reduction is \WithinReduction\% (95\% interval [5.3\%, 12.0\%]), with \WithinLowerCount/330 positive reductions. Table~\ref{tab:correlation_models} reports every model--profile cell. The median reduction is 7.1\%; mean reductions are positive in 27/30 model--profile cells and on all eleven decision dates. The paired design holds the class score fixed.
\begin{table}[t]
\centering
\small
\begin{adjustbox}{max width=\columnwidth,center}
\begin{tabular}{lrrrrr}
\toprule
Model & Cons. & Bal. & Agg. & All & Lower \\
\midrule
DS-V4-Flash & 1.1 & 1.8 & 3.0 & 2.0 & 26/33 \\
DS-V4-Pro & 1.7 & 3.2 & 1.4 & 2.1 & 26/33 \\
Qwen3.6-Plus & 15.6 & 14.4 & 24.2 & 18.1 & 33/33 \\
Qwen3.7-Max & 9.5 & 19.6 & 24.9 & 18.0 & 32/33 \\
Qwen3.6-35B-A3B & -7.1 & 7.0 & 16.3 & 5.4 & 23/33 \\
GLM-5.1 & 0.3 & 15.4 & 13.4 & 9.7 & 28/33 \\
DB-2.0-Lite & -1.2 & 9.1 & 20.2 & 9.4 & 25/33 \\
DB-2.0-Pro & -6.2 & 10.1 & 16.8 & 6.9 & 22/33 \\
HY3-Preview & 0.0 & 4.8 & 5.7 & 3.5 & 27/33 \\
Kimi-K2.6 & 1.7 & 10.2 & 18.4 & 10.1 & 29/33 \\
\midrule
All models & 1.6 & 9.6 & 14.4 & 8.5 & 271/330 \\
\bottomrule
\end{tabular}
\end{adjustbox}
\caption{Estimated volatility reduction (\%) from equalizing held securities within each class at fixed class totals and uninvested cash. Each cell contains eleven decisions; Lower counts positive reductions.}
\label{tab:correlation_models}
\end{table}

Pairwise covariance estimation uses each pair's overlapping observations in the same lookback window. It gives an 8.1\% mean reduction (95\% interval [5.0\%, 11.5\%]), with 270/330 positive reductions. Table~\ref{tab:correlation_sensitivity} also reports controls that equalize across all eligible securities within each class, or across all eligible securities. All controls retain the original uninvested cash. Expanding within-class holdings gives a 9.6\% mean reduction under the common-date estimator; equal weighting across all eligible securities gives 15.5\%. The primary held-asset control isolates weighting at fixed asset selection and class exposure. The main analysis uses common-date covariance to remain positive semidefinite.
\begin{table}[t]
\centering
\small
\begin{adjustbox}{max width=\columnwidth,center}
\begin{tabular}{lrr}
\toprule
Control / covariance & Reduction [95\% CI] & Lower \\
\midrule
Held assets / common & 8.5 [5.3, 12.0] & 271/330 \\
Held assets / pairwise & 8.1 [5.0, 11.5] & 270/330 \\
All assets / fixed classes & 9.6 [5.5, 13.8] & 266/330 \\
Full-support equal weight & 15.5 [9.8, 21.0] & 259/330 \\
\bottomrule
\end{tabular}
\end{adjustbox}
\caption{Volatility reductions (\%) under alternative covariance estimators and allocation controls. Expanded-support controls use common-date covariance; all controls preserve uninvested cash. Intervals resample models with all profiles and dates retained.}
\label{tab:correlation_sensitivity}
\end{table}

\paragraph{Representative allocation.}
DB-2.0-Pro's balanced allocation on November 1, 2024 holds 82 securities. Equalizing within-class weights at fixed holdings and class exposures lowers estimated volatility by 8.0\% and reduces $\bar\rho$ by .074. Its largest positive pairwise covariance contribution is AVGO--NVDA, with correlation .765 and 2.00\% of total portfolio variance. NVDA--XLK contributes a further 1.07\%, with correlation .887. Both pairs contribute correlated risk within this broad portfolio.

\section{Real-Time Evaluation}
\label{app:live_eval}

As discussed in Section~\ref{sec:live_eval}, historical evaluation can overlap with market information in model pretraining~\citep{time-travel-is-cheating-neurips25,stockbench-arxiv25}. \portbench therefore includes a real-time evaluation mode: at each decision date $D$, the model receives only information available through the previous market close.

\begin{figure}[!h]
\centering
\includegraphics[width=\columnwidth]{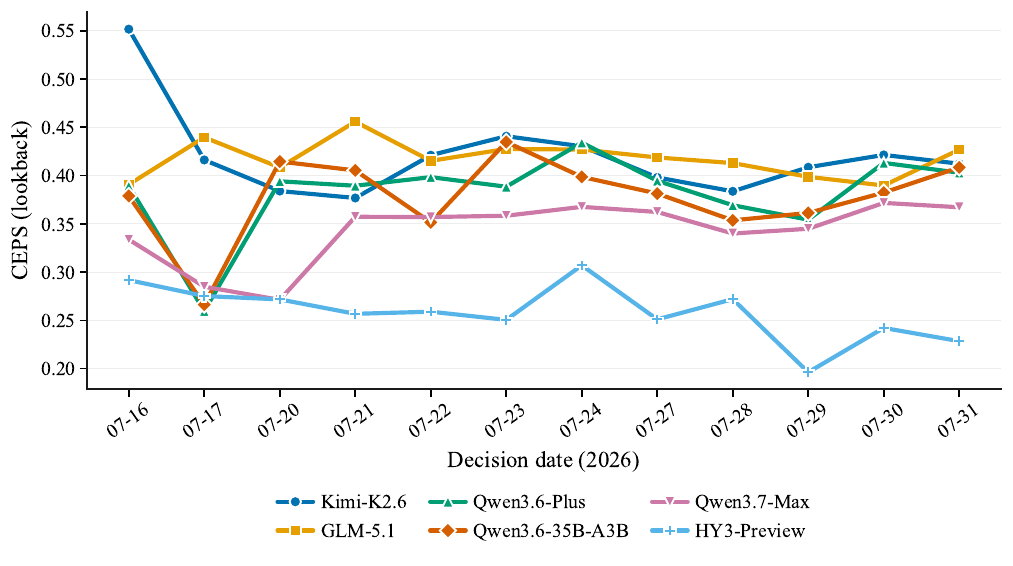}
\caption{Per-decision lookback CEPS over the 12-day real-time window (balanced profile).}
\label{fig:live_ceps_daily}
\end{figure}

\paragraph{Setup.}
We run real-time evaluation with daily rebalancing from 2026-07-16 to 2026-07-31 (12 trading days) under the balanced profile for six models: HY3-Preview, Qwen3.7-Max, Qwen3.6-Plus, Qwen3.6-35B-A3B, GLM-5.1, and Kimi-K2.6. DS-V4 and DB-2.0 models were unavailable through their APIs during the run. Temperature and decoding settings match Section~\ref{sec:setup}. Table~\ref{tab:live_ceps} reports mean \ceps under the lookback S3 reference, and Figure~\ref{fig:live_ceps_daily} plots daily scores.

\begin{table}[!h]
\centering
\small
\begin{tabular}{lc}
\toprule
\textbf{Model} & \textbf{CEPS\textsubscript{lookback}} \\
\midrule
Kimi-K2.6        & \textbf{.420} \\
GLM-5.1          & .418 \\
Qwen3.6-Plus     & .382 \\
Qwen3.6-35B-A3B  & .378 \\
Qwen3.7-Max      & .343 \\
HY3-Preview      & .258 \\
\bottomrule
\end{tabular}
\caption{Mean CEPS over the daily real-time window (2026-07-16--07-31, balanced profile, 12 decision dates). Models are ranked by lookback CEPS.}
\label{tab:live_ceps}
\end{table}

\begin{figure}[!h]
\centering
\includegraphics[width=\columnwidth]{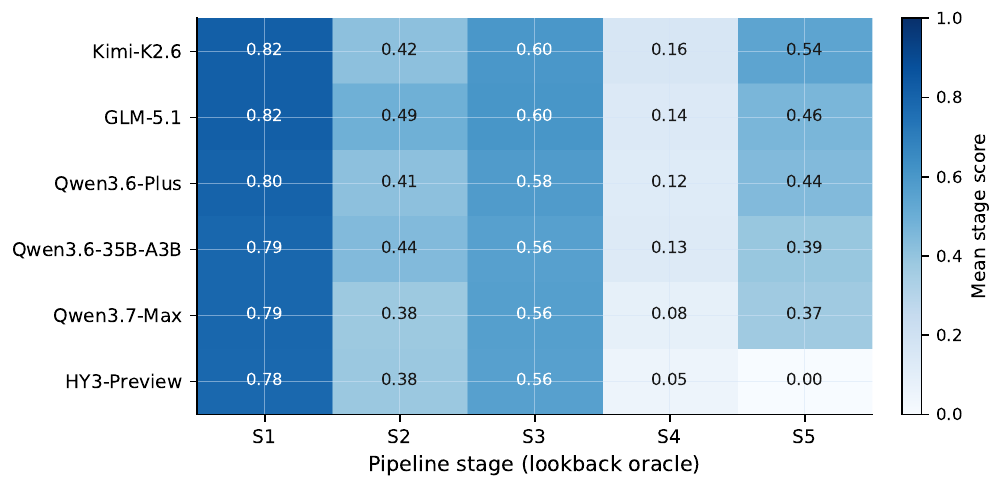}
\caption{Mean lookback stage scores (S1--S5) over the 12 real-time decision dates.}
\label{fig:live_stage_scores}
\end{figure}

Kimi-K2.6 leads with lookback CEPS .420, closely followed by GLM-5.1 at .418. Figure~\ref{fig:live_stage_scores} shows high S1 (.78--.82), lookback S3 between .561 and .604, and low S4 (.047--.163). HY3-Preview has mean lookback S5 of 0. Figure~\ref{fig:live_ceps_daily} shows sharp drops for Qwen3.6-Plus and Qwen3.6-35B-A3B on 2026-07-17, while GLM-5.1 has the lowest day-level variation (std.\ .020). \textbf{Daily scores vary across models and dates, while turnover agreement remains low.}

\paragraph{Scoring.}
The lookback reference uses trailing observations available at the decision date. The daily results use a signal-constrained reference and total-variation weight scoring for S3. Historical portfolio comparisons use the profile-constrained lookback policy in Eq.~\ref{eq:app_wstar}. The twelve-day period, daily frequency, and six-model coverage complement the historical evaluation.

\section{Qualitative Examples}
\label{app:showcase}

This appendix provides detailed examples of three core components of \portbench: the market base dataset (\S\ref{app:showcase_snapshot}), the QA dataset (\S\ref{app:showcase_qa}), and the five-stage evaluation pipeline (\S\ref{app:showcase_pipeline}).

\subsection{Market Snapshot Examples}
\label{app:showcase_snapshot}

\begin{figure*}[!ht]
\centering
\small
\begin{tcolorbox}[colback=white, colframe=deepslate, title={\small\textbf{MarketSnapshot} {\scriptsize (input constructed from trailing window at each rebalance date)} \hfill 2024-03-01 $\mid$ Balanced Profile}, fonttitle=\small\bfseries, boxrule=0.4pt, left=3pt, right=3pt, top=2pt, bottom=2pt]

\begin{tcolorbox}[colback=nearwhite, colframe=coolblue, boxrule=0.3pt, left=2pt, right=2pt, top=1pt, bottom=1pt, title={\scriptsize \textbf{Per-Asset Summary} (representative rows from the 183-series dataset)}]
\scriptsize
\begin{tabular}{llrrrrrrl}
\textbf{Class} & \textbf{Ticker} & \textbf{Close} & \textbf{20d Ret.} & \textbf{60d Ret.} & \textbf{Ann.\ Vol} & \textbf{MaxDD\textsubscript{60d}} & \textbf{Sharpe\textsubscript{60d}} & \textbf{Regime} \\
\midrule
Equities & SPY & 499.11 & $+$4.2\% & $+$8.4\% & 11.6\% & $-$1.7\% & 2.31 & bull \\
         & QQQ & 440.19 & $+$5.8\% & $+$15.3\% & 14.9\% & $-$2.4\% & 3.28 & bull \\
         & XLE & 89.74 & $+$7.1\% & $+$10.2\% & 18.3\% & $-$3.1\% & 1.78 & bull \\
Bonds    & TLT & 94.47 & $-$1.8\% & $-$3.0\% & 14.3\% & $-$4.9\% & $-$0.67 & bear \\
         & IEF & 93.49 & $-$0.5\% & $-$0.9\% & 7.2\% & $-$2.1\% & $-$0.40 & sideways \\
Commod.  & GLD & 202.17 & $+$3.1\% & $+$5.9\% & 13.3\% & $-$1.5\% & 1.41 & bull \\
         & USO & 76.83 & $+$2.4\% & $+$4.1\% & 26.7\% & $-$5.8\% & 0.49 & sideways \\
Crypto   & BTC & 61,148 & $+$22.3\% & $+$46.5\% & 50.1\% & $-$8.3\% & 2.96 & bull \\
         & ETH & 3,412 & $+$18.7\% & $+$38.2\% & 58.4\% & $-$11.6\% & 2.09 & bull \\
RE       & VNQ & 83.11 & $+$0.3\% & $+$1.8\% & 17.0\% & $-$4.2\% & 0.34 & sideways \\
Cash     & BIL & 91.63 & $+$0.04\% & $+$0.3\% & 0.3\% & $-$0.0\% & 3.19 & sideways \\
\end{tabular}\\[2pt]
{\scriptsize \textit{Each asset also carries: daily return series, cumulative return curve, sector/sub-class tag.}}
\end{tcolorbox}

\vspace{2pt}
\begin{tcolorbox}[colback=nearwhite, colframe=coolblue, boxrule=0.3pt, left=2pt, right=2pt, top=1pt, bottom=1pt, title={\scriptsize \textbf{Daily Price Series} (excerpt from the 60-day lookback, normalized to 100 at excerpt start)}]
\centering
\includegraphics[width=0.95\textwidth]{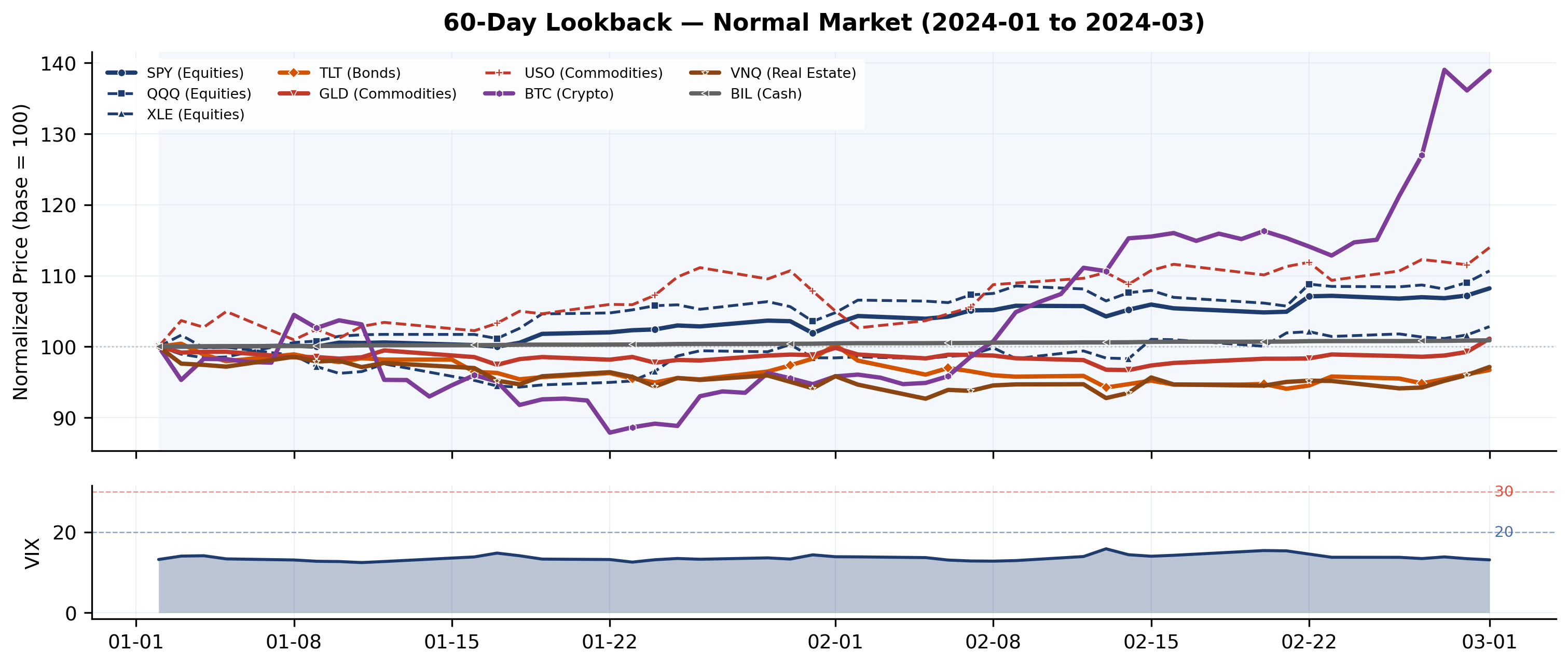}\\[2pt]
{\scriptsize \textcolor{gray}{Representative assets from each class over the displayed interval (2024-01-02 to 2024-03-01). BTC surges $+$46\% while TLT declines $-$3\%, illustrating the cross-class divergence that drives correlation-aware allocation. Daily closes and returns are drawn from the lookback window for the assets available in each snapshot.}}
\end{tcolorbox}

\vspace{2pt}
\begin{tcolorbox}[colback=nearwhite, colframe=softperi, boxrule=0.3pt, left=2pt, right=2pt, top=1pt, bottom=1pt, title={\scriptsize \textbf{Macroeconomic Indicators}}]
\scriptsize
\begin{tabular}{llll}
Fed Funds Rate: 5.33\% & VIX: 13.11 & Unemployment: 3.90\% & 10Y--2Y Spread: $-$0.35 \\
30Y Mortgage: 6.94\% & HY OAS: 3.32 & 10Y Breakeven: 2.32\% & TED Spread: 0.09 \\
\end{tabular}
\end{tcolorbox}

\vspace{2pt}
\begin{tcolorbox}[colback=nearwhite, colframe=warmcream!80!black, boxrule=0.3pt, left=2pt, right=2pt, top=1pt, bottom=1pt, title={\scriptsize \textbf{Two-Layer Correlation Interface} \hfill trailing 60-day window}]
\scriptsize
\begin{tabular}{lrr}
\textbf{Class} & \textbf{Intra-$\bar{\rho}$} & \textbf{versus Equities $\bar{\rho}$} \\
\midrule
Equities & $+$0.48 & --- \\
Bonds & $+$0.77 & $-$0.18 \\
Commodities & $+$0.23 & $+$0.12 \\
Crypto & $+$0.60 & $+$0.20 \\
Real Estate & $+$0.77 & $+$0.53 \\
Cash & $+$0.21 & $-$0.07 \\
\end{tabular}
\end{tcolorbox}

\vspace{2pt}
\begin{tcolorbox}[colback=nearwhite, colframe=lavblue, boxrule=0.3pt, left=2pt, right=2pt, top=1pt, bottom=1pt, title={\scriptsize \textbf{Portfolio State}}]
\scriptsize NAV = \$100{,}000 \quad$\mid$\quad Weights: equal-weight ($\frac{1}{n_t}$ per eligible security)
\end{tcolorbox}

\end{tcolorbox}
\caption{Illustration of \texttt{MarketSnapshot} fields using the 2024-03-01 market data and an illustrative equal-weight portfolio over eligible securities. The four input layers contain per-asset price data, macroeconomic indicators, a two-layer correlation interface, and portfolio state.}
\label{fig:showcase_snapshot}
\end{figure*}

\begin{figure*}[!ht]
\centering
\footnotesize
\begin{tcolorbox}[colback=white, colframe=deepslate, title={\footnotesize\textbf{MarketSnapshot} {\scriptsize COVID Crash} \hfill 2020-03-13 $\mid$ Conservative}, fonttitle=\footnotesize\bfseries, boxrule=0.4pt, left=2pt, right=2pt, top=1pt, bottom=1pt]

\begin{tcolorbox}[colback=nearwhite, colframe=coolblue, boxrule=0.3pt, left=2pt, right=2pt, top=0pt, bottom=0pt, title={\scriptsize \textbf{Per-Asset Summary} (representative rows)}]
\scriptsize
\begin{tabular}{llrrrrrl}
\textbf{Class} & \textbf{Ticker} & \textbf{Close} & \textbf{20d} & \textbf{60d} & \textbf{Vol} & \textbf{MaxDD} & \textbf{Regime} \\
\midrule
Equities & SPY & 270.06 & $-$9.8\% & $-$13.4\% & 44.2\% & $-$19.8\% & crisis \\
         & XLE & 230.83 & $-$22.1\% & $-$31.3\% & 56.8\% & $-$38.4\% & crisis \\
Bonds    & TLT & 151.57 & $+$8.3\% & $+$17.1\% & 18.3\% & $-$5.2\% & bull \\
Commod.  & USO & 68.26 & $-$28.7\% & $-$36.8\% & 62.3\% & $-$42.1\% & crisis \\
Crypto   & BTC & 5,418 & $-$4.2\% & $-$7.8\% & 45.2\% & $-$38.9\% & bear \\
RE / Cash & VNQ / SHV & 69.24 / 110.13 & $-$12.4\% / $+$0.3\% & $-$18.9\% / $+$0.7\% & 51.7\% / 0.4\% & $-$26.3\% / 0\% & crisis / side. \\
\end{tabular}
\end{tcolorbox}

\vspace{1pt}
\begin{tcolorbox}[colback=nearwhite, colframe=coolblue, boxrule=0.3pt, left=2pt, right=2pt, top=0pt, bottom=0pt, title={\scriptsize \textbf{Daily Price Series} (60-day lookback)}]
\centering
\includegraphics[width=0.82\textwidth]{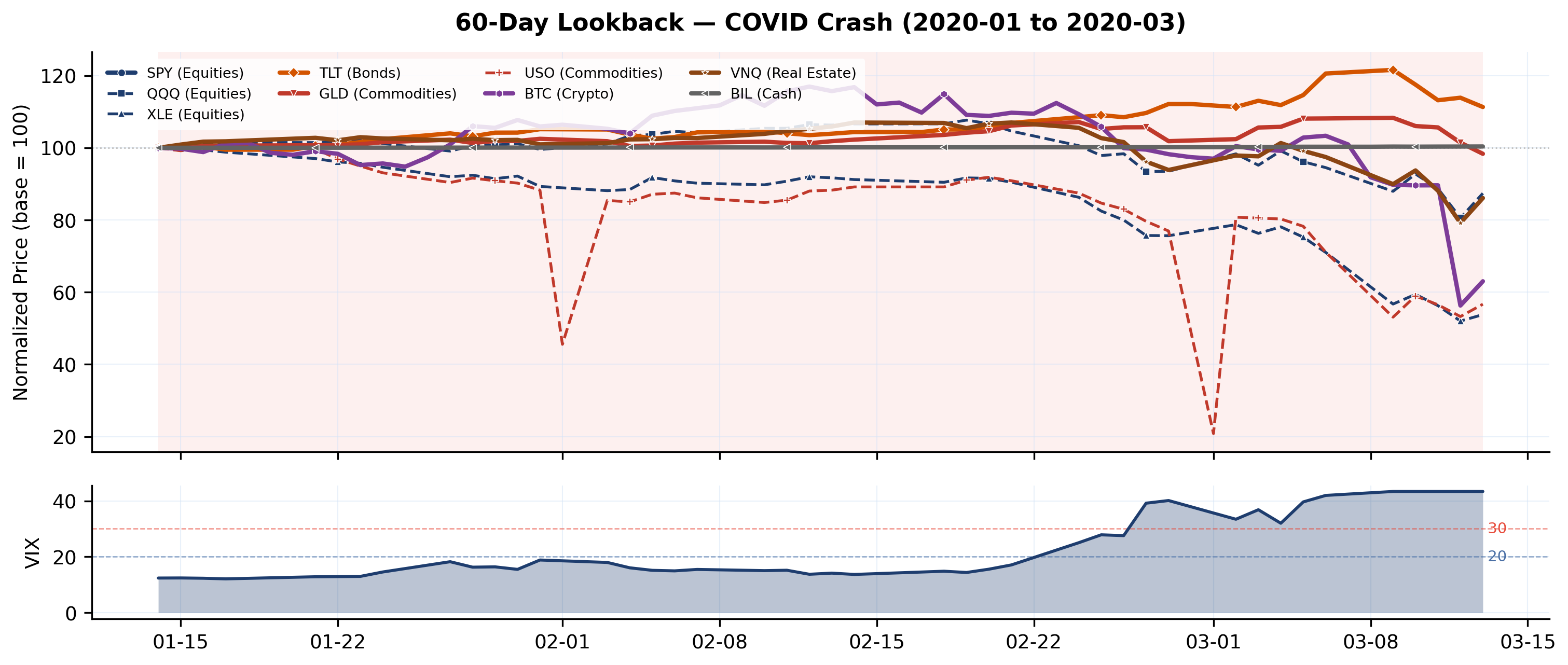}
\end{tcolorbox}

\vspace{1pt}
\begin{tcolorbox}[colback=nearwhite, colframe=softperi, boxrule=0.3pt, left=2pt, right=2pt, top=0pt, bottom=0pt, title={\scriptsize \textbf{Macros} $\mid$ VIX 43.35, HY OAS 7.31, TED 0.57}]
\scriptsize
Fed Funds 1.10\% \quad Unemployment 4.40\% \quad 10Y--2Y 0.45 \quad 10Y Breakeven 1.07\% \quad 30Y Mortgage 3.36\%
\end{tcolorbox}

\vspace{1pt}
\begin{tcolorbox}[colback=nearwhite, colframe=warmcream!80!black, boxrule=0.3pt, left=2pt, right=2pt, top=0pt, bottom=0pt, title={\scriptsize \textbf{Two-Layer Correlations} (stress; RE--Eq $\bar{\rho}=0.61$)}]
\scriptsize
\centering
\includegraphics[width=0.72\textwidth]{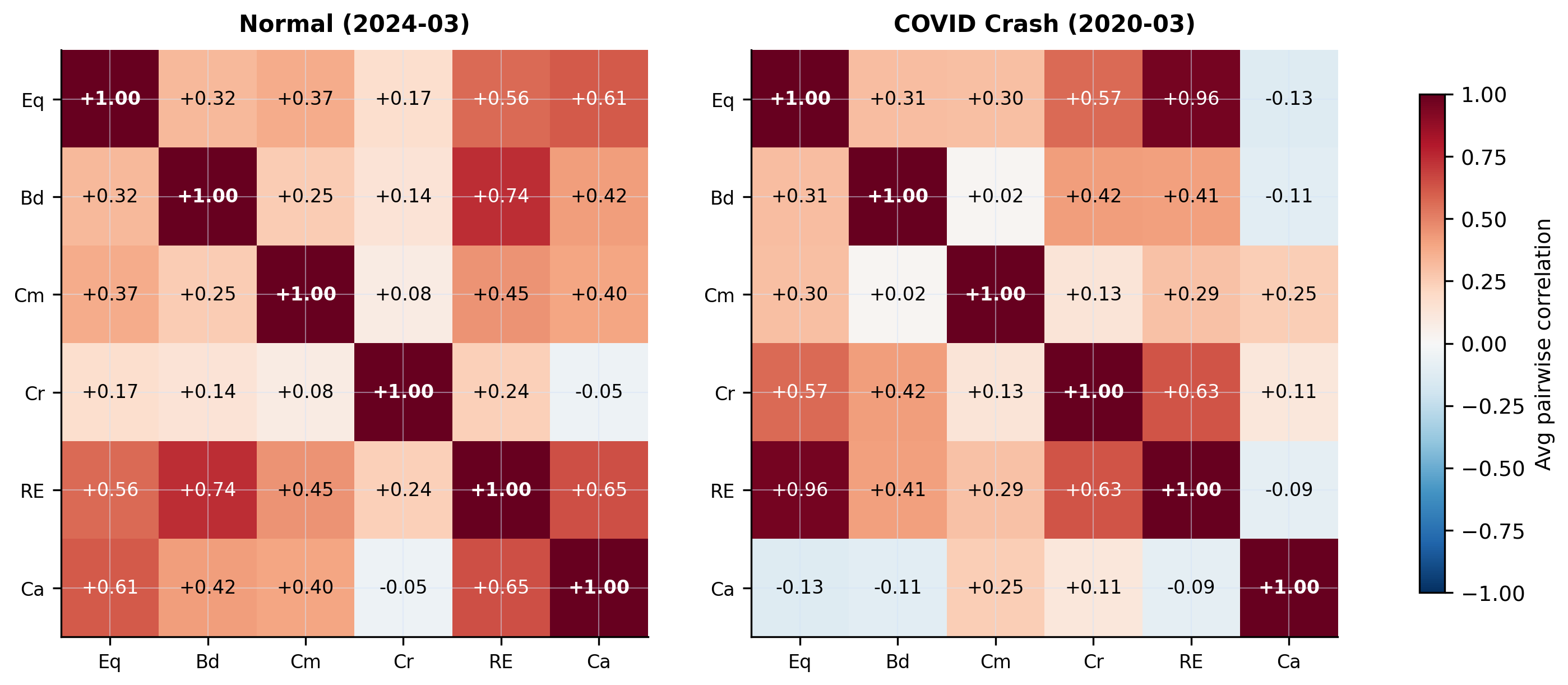}
\end{tcolorbox}

\vspace{1pt}
\begin{tcolorbox}[colback=nearwhite, colframe=lavblue, boxrule=0.3pt, left=2pt, right=2pt, top=0pt, bottom=0pt, title={\scriptsize \textbf{Portfolio State}}]
\scriptsize NAV = \$94{,}900 \quad$\mid$\quad equal-weight ($\frac{1}{n_t}$ over eligible securities) \quad$\mid$\quad Drawdown from peak: $-$5.1\%
\end{tcolorbox}

\end{tcolorbox}
\caption{Illustration of \texttt{MarketSnapshot} fields during the 2020 COVID Crash with an illustrative equal-weight portfolio over eligible securities. Compared with Figure~\ref{fig:showcase_snapshot}, VIX is higher (43 versus 13), equities/commodities sell off, bonds rally, and credit spreads widen.}
\label{fig:showcase_snapshot_stress}
\end{figure*}

A \texttt{MarketSnapshot} is the structured input constructed from the trailing lookback window and provided to the model at each rebalance date for the five-stage decision process. At each rebalance date, the model receives the snapshot through three isolated prompts and produces S1--S3; the sandbox then executes the S3 allocation and computes the S4/S5 execution and risk scores used in the ranking. The snapshot contains four structured layers: (1)~per-asset price summaries with trailing returns and volatilities, (2)~macroeconomic indicators, (3)~a pairwise return correlation matrix with intra-class and inter-class aggregation, and (4)~the current portfolio state. The two-layer correlation interface exposes within-class co-movement and cross-class hedging for joint allocation. Figures~\ref{fig:showcase_snapshot} and~\ref{fig:showcase_snapshot_stress} illustrate these fields using representative assets from the six-class market dataset under contrasting conditions: a calm bull market (2024-03) and the COVID crash (2020-03). The field illustrations draw representative series from the full six-class dataset of 183 market series. Historical comparisons use the date-specific tradable subset in Appendix~\ref{app:evaluation_details}; portfolio states are illustrative.

\subsection{QA Dataset Samples}
\label{app:showcase_qa}

Figure~\ref{fig:showcase_qa} presents one representative sample from each of the seven QA templates (T1--T7), spanning complexity levels L1--L4. Each shows the context, ground-truth answer, and key reasoning step.

\begin{figure*}[!ht]
\centering
\small

\begin{tcolorbox}[colback=white, colframe=blue!40!black, title={\small\textbf{T1: Return Prediction} \hfill Complexity 1 $\mid$ Single Asset $\mid$ Train}, fonttitle=\small, boxrule=0.4pt, left=3pt, right=3pt, top=2pt, bottom=2pt]
\textbf{Context:} ETH-USD over past 60 days: start=\$320.88, end=\$441.72, cumulative return $+$37.7\%, ann.\ vol 79.6\%. Macro: fed\_funds=1.16\%, VIX=9.73. Regime: sideways.\\[2pt]
\textbf{Question:} Predict whether the return of ETH-USD over the next 21 trading days will be: \textit{positive} ($>+1\%$), \textit{negative} ($<-1\%$), or \textit{flat} ($\pm1\%$).\\[2pt]
\textbf{Answer:} \texttt{positive} \quad {\scriptsize (actual forward return: $+$49.99\%)}
\end{tcolorbox}

\vspace{3pt}

\begin{tcolorbox}[colback=white, colframe=blue!40!black, title={\small\textbf{T2: Risk Assessment (VaR)} \hfill Complexity 1 $\mid$ Single Asset $\mid$ Train}, fonttitle=\small, boxrule=0.4pt, left=3pt, right=3pt, top=2pt, bottom=2pt]
\textbf{Context:} XLV daily returns (60 days): mean=0.0000, std=0.0053, min=$-$0.0123, max=$+$0.0133. Regime: sideways.\\[2pt]
\textbf{Question:} Using the historical simulation method, compute the 1-day VaR at 95\% confidence level for XLV.\\[2pt]
\textbf{Answer:} $-$0.0080 \quad {\scriptsize (sort 60 returns, take 5th percentile)}
\end{tcolorbox}

\vspace{3pt}

\begin{tcolorbox}[colback=white, colframe=blue!40!black, title={\small\textbf{T3: Position Sizing} \hfill Complexity 1 $\mid$ Single Asset $\mid$ Train}, fonttitle=\small, boxrule=0.4pt, left=3pt, right=3pt, top=2pt, bottom=2pt]
\textbf{Context:} VCIT daily returns (60 days): std=0.0015, worst\_day=$-$0.0043. Max drawdown threshold: 10\%.\\[2pt]
\textbf{Question:} Determine the maximum fraction of total portfolio to allocate to VCIT using the fixed-fractional formula $f^* = \min(1,\, \delta_\text{max} / |\text{VaR}_{99\%}|)$.\\[2pt]
\textbf{Answer:} 1.0000 \quad {\scriptsize ($f^* = 0.10 / 0.0034 = 29.4 \to$ capped at 1.0)}
\end{tcolorbox}

\vspace{3pt}

\begin{tcolorbox}[colback=white, colframe=teal!50!black, title={\small\textbf{T4: Pairwise Min-Variance Allocation} \hfill Complexity 2 $\mid$ 2 Assets $\mid$ Val}, fonttitle=\small, boxrule=0.4pt, left=3pt, right=3pt, top=2pt, bottom=2pt]
\textbf{Context:} XLY: daily std$=$0.0123, annualized mean$=$0.2016. STIP: daily std$=$0.0021, annualized mean$=$0.0504. Minimum annualized portfolio return: 0.1224.\\[2pt]
\textbf{Question:} Compute minimum-variance weights satisfying the return floor, $\sum w = 1$, and $w \geq 0$.\\[2pt]
\textbf{Answer:} $w_\text{XLY} = 0.4762$, $w_\text{STIP} = 0.5238$ \quad {\scriptsize (binding return floor: $w_\text{XLY}=(0.1224-0.0504)/(0.2016-0.0504)$)}
\end{tcolorbox}

\vspace{3pt}

\begin{tcolorbox}[colback=white, colframe=orange!50!black, title={\small\textbf{T5: Max-Sharpe Optimization} \hfill Complexity 3 $\mid$ 4 Assets $\mid$ Val}, fonttitle=\small, boxrule=0.4pt, left=3pt, right=3pt, top=2pt, bottom=2pt]
\textbf{Context:} Assets: VLUE, XRP-USD, DBC, BIL. Risk-free rate: 4.00\%. Covariance matrix provided.\\[2pt]
\textbf{Question:} Compute portfolio weights that maximize the Sharpe ratio. Constraints: $\sum w_i = 1$, $w_i \geq 0$.\\[2pt]
\textbf{Answer:} $w_\text{VLUE}{=}0.000$, $w_\text{XRP}{=}0.001$, $w_\text{DBC}{=}0.000$, $w_\text{BIL}{=}0.999$ \quad {\scriptsize (Sharpe = 1.544)}
\end{tcolorbox}

\vspace{3pt}

\begin{tcolorbox}[colback=white, colframe=orange!50!black, title={\small\textbf{T6: Rebalancing Decision} \hfill Complexity 3 $\mid$ 4 Assets $\mid$ Val}, fonttitle=\small, boxrule=0.4pt, left=3pt, right=3pt, top=2pt, bottom=2pt]
\textbf{Context:} Portfolio: IVV, BNB-USD, SGOV, PDBC. Current versus target weights given. Rebalance threshold $\delta_\text{reb}{=}5\%$; largest $|w^\text{current}-w^\text{target}|{=}6.5\%$.\\[2pt]
\textbf{Question:} (A) Should the portfolio be rebalanced? (B) If yes, identify the largest trade needed.\\[2pt]
\textbf{Answer:} Yes; buy 0.065 of IVV \quad {\scriptsize ($6.5\% > \delta_\text{reb}{=}5\%$ $\Rightarrow$ rebalance triggered)}
\end{tcolorbox}

\vspace{3pt}

\begin{tcolorbox}[colback=white, colframe=red!40!black, title={\small\textbf{T7: Regime Detection and Allocation} \hfill Complexity 4 $\mid$ 6 Classes $\mid$ Test}, fonttitle=\small, boxrule=0.4pt, left=3pt, right=3pt, top=2pt, bottom=2pt]
\textbf{Context:} SPY (60d): \$664.67$\to$\$685.14, return $+$3.1\%, vol 12.9\%. Macro: fed\_funds=3.64\%, VIX=13.47.\\[2pt]
\textbf{Part A:} Identify market regime from \{bull, bear, sideways, crisis\}.\\
\textbf{Part B:} For each of six asset classes, recommend: increase, decrease, or hold.\\[2pt]
\textbf{Answer:} Regime: \texttt{sideways}. Equities: \textit{hold}, Bonds: \textit{hold}, Commodities: \textit{hold}, RE: \textit{hold}, Crypto: \textit{decrease}, Cash: \textit{increase}.
\end{tcolorbox}

\caption{Representative QA samples from all seven templates (T1--T7). Color indicates complexity level: blue = L1 single-asset prediction/sizing (T1--T3), teal = L2 pairwise allocation (T4), orange = L3 multi-asset optimization/rebalancing (T5--T6), red = L4 multi-signal judgment (T7).}
\label{fig:showcase_qa}
\end{figure*}

\subsection{Pipeline Evaluation Traces}
\label{app:showcase_pipeline}

Figures~\ref{fig:showcase_pipeline_qwen} to~\ref{fig:showcase_pipeline_qwen37} show complete five-stage traces for three models across normal and stress windows. Each trace reports a prompt summary, selected model outputs, reference targets, and stage scores. Scoring and execution use the eligible securities in each snapshot. Model outputs are shown as recorded. Their weights are normalized for execution, with excluded allocations held as uninvested cash; VIX appears as market context and any proposed VIX weight is held as cash. S1/S2 use the trailing-return targets, and S3 uses the profile-constrained lookback reference in Eq.~\ref{eq:app_wstar}. S4 applies the common next-close execution and cost policy, while S5 evaluates the resulting lookback risk state. The examples cover broad equity exposure, a large defensive reallocation, and a cash-heavy allocation.

\begin{figure*}[!ht]
\centering
\small
\begin{tcolorbox}[colback=white, colframe=deepslate, title={\small\textbf{Qwen3.6-Plus} \hfill 2024-03-01 $\mid$ Balanced $\mid$ Normal Period}, fonttitle=\small\bfseries, boxrule=0.4pt, left=3pt, right=3pt, top=2pt, bottom=2pt]

\begin{tcolorbox}[colback=nearwhite, colframe=coolblue, boxrule=0.3pt, left=2pt, right=2pt, top=1pt, bottom=1pt, title={\scriptsize \textbf{S1: Market Interpretation} \hfill Score = 0.771}]
{\scriptsize \textcolor{gray}{\textbf{Prompt summary:} Assess the market on 2024-03-01 from the supplied asset and macro features. Return JSON with a view for each supplied asset, a regime, confidence, and a macro summary.}}\\[3pt]
\begin{tabular}{p{0.46\textwidth}p{0.46\textwidth}}
\textbf{Model Output} & \textbf{Reference Target} \\
\midrule
\scriptsize SPY: $+$0.60, QQQ: $+$0.80, TLT: $+$0.10, GLD: $+$0.60, BIL: $+$0.60 & \scriptsize SPY: $+$1.00, QQQ: $+$1.00, TLT: $-$0.25, GLD: $+$0.27, BIL: $+$0.01 \\[2pt]
\multicolumn{2}{l}{\scriptsize \textit{Scoring:} $\sigma_1 = 1 - \frac{1}{2n}\sum_i |\hat{v}_i-v_i^*|$ over eligible securities.}
\end{tabular}
\end{tcolorbox}

\vspace{2pt}
\begin{tcolorbox}[colback=nearwhite, colframe=softperi, boxrule=0.3pt, left=2pt, right=2pt, top=1pt, bottom=1pt, title={\scriptsize \textbf{S2: Signal Generation} \hfill Score = 0.477}]
{\scriptsize \textcolor{gray}{\textbf{Prompt summary:} Use the S1 views to return JSON with a buy, hold, or sell signal and signal strength for each supplied asset.}}\\[3pt]
\begin{tabular}{p{0.46\textwidth}p{0.46\textwidth}}
\textbf{Model Output} & \textbf{Reference Target} \\
\midrule
\scriptsize SPY: buy, QQQ: buy, TLT: hold, GLD: buy, BIL: buy & \scriptsize SPY: buy, QQQ: buy, TLT: sell, GLD: buy, BIL: hold \\[2pt]
\multicolumn{2}{l}{\scriptsize \textit{Scoring:} Fraction of eligible-security signals matching the reference target.}
\end{tabular}
\end{tcolorbox}

\vspace{2pt}
\begin{tcolorbox}[colback=nearwhite, colframe=warmcream!80!black, boxrule=0.3pt, left=2pt, right=2pt, top=1pt, bottom=1pt, title={\scriptsize \textbf{S3: Weight Optimization} \hfill Score = 0.313}]
{\scriptsize \textcolor{gray}{\textbf{Prompt summary:} Use the S2 signals and supplied correlation information to return JSON with portfolio weights, expected return, volatility, and Sharpe.}}\\[3pt]
\begin{tabular}{p{0.46\textwidth}p{0.46\textwidth}}
\textbf{Model Output} & \textbf{S3 Reference} \\
\midrule
\scriptsize $w_\text{NVDA}{=}0.034$, $w_\text{QQQ}{=}0.028$, $w_\text{AMAT}{=}0.028$, $w_\text{AVGO}{=}0.027$ & \scriptsize $w_\text{BIL}{=}0.567$, $w_\text{DBA}{=}0.092$, $w_\text{STIP}{=}0.090$, $w_\text{XLV}{=}0.086$ \\[2pt]
\multicolumn{2}{l}{\scriptsize \textit{Scoring:} $\sigma_3=0.5\,s_\text{acc}+0.5\,s_\text{corr}$ with $s_\text{acc}=1-\operatorname{TV}$ (Eq.~\ref{eq:app_sacc}). Here $s_\text{acc}=0.129$ and $s_\text{corr}=0.497$.}
\end{tabular}
\end{tcolorbox}

\vspace{2pt}
\begin{tcolorbox}[colback=nearwhite, colframe=lavblue, boxrule=0.3pt, left=2pt, right=2pt, top=1pt, bottom=1pt, title={\scriptsize \textbf{S4: Execution Simulation (deterministic)} \hfill Score = 0.030}]
{\scriptsize \textcolor{gray}{The deterministic S4 calculation defined above produces the following model and reference turnover.}}\\[3pt]
\begin{tabular}{p{0.46\textwidth}p{0.46\textwidth}}
\textbf{Model Turnover (from S3)} & \textbf{Reference Turnover} \\
\midrule
\scriptsize 5.1\% (change to the model target) & \scriptsize 174.1\% (change to the reference target) \\[2pt]
\multicolumn{2}{l}{\scriptsize \textit{Scoring:} $\sigma_4=1-|\tau-\tau^*|/\max(\tau,\tau^*,10^{-4})\approx 0.030$. Both changes start from the same pre-trade state.}
\end{tabular}
\end{tcolorbox}

\vspace{2pt}
\begin{tcolorbox}[colback=nearwhite, colframe=coralred, boxrule=0.3pt, left=2pt, right=2pt, top=1pt, bottom=1pt, title={\scriptsize \textbf{S5: Risk Monitoring (deterministic)} \hfill Score = 0.000}]
{\scriptsize \textcolor{gray}{The deterministic S5 calculation defined above produces the following model and reference risk states.}}\\[3pt]
\begin{tabular}{p{0.46\textwidth}p{0.46\textwidth}}
\textbf{Computed from Model Weights} & \textbf{Computed from Reference Weights} \\
\midrule
\scriptsize VaR: $-$1.10\%, MaxDD: $-$2.45\%, Drift: 2.2\%, Rebalance: No & \scriptsize VaR: $-$0.18\%, MaxDD: $-$0.22\%, Drift: 55.6\%, Rebalance: Yes \\[2pt]
\multicolumn{2}{l}{\scriptsize \textit{Scoring:} $\sigma_5=0.5\,d_\text{reb}+0.5\,\operatorname{clip}(1-(e_\text{VaR}+e_\text{DD})/2,0,1)$. Rebalance flags and lookback risk magnitudes differ.}
\end{tabular}
\end{tcolorbox}

\vspace{3pt}
{\scriptsize \textbf{Summary:} The model spreads weight broadly, while the reference places its largest weight in BIL. Turnover is 5.1\% versus 174.1\%, giving S4 of 0.030. Adjacent score drops total 0.771, giving CEPS of 0.241.}

\end{tcolorbox}
\caption{Pipeline trace for Qwen3.6-Plus under normal conditions. A small model reallocation (5.1\% turnover) contrasts with 174.1\% for the reference.}
\label{fig:showcase_pipeline_qwen}
\end{figure*}

\begin{figure*}[!ht]
\centering
\small
\begin{tcolorbox}[colback=white, colframe=deepslate, title={\small\textbf{GLM-5.1} \hfill 2020-02-28 $\mid$ Balanced $\mid$ COVID Flash Crash}, fonttitle=\small\bfseries, boxrule=0.4pt, left=3pt, right=3pt, top=2pt, bottom=2pt]

\begin{tcolorbox}[colback=nearwhite, colframe=coolblue, boxrule=0.3pt, left=2pt, right=2pt, top=1pt, bottom=1pt, title={\scriptsize \textbf{S1: Market Interpretation} \hfill Score = 0.802}]
{\scriptsize \textcolor{gray}{\textbf{Prompt summary:} Assess the market on 2020-02-28 from the supplied asset and macro features. Return JSON with a view for each supplied asset, a regime, confidence, and a macro summary.}}\\[3pt]
\begin{tabular}{p{0.46\textwidth}p{0.46\textwidth}}
\textbf{Model Output} & \textbf{Reference Target} \\
\midrule
\scriptsize IWM: $-$0.70, XLE: $-$0.70, TLT: $+$0.70, GLD: $+$0.40, VIX: $+$0.80 & \scriptsize IWM: $-$0.86, XLE: $-$1.00, TLT: $+$1.00, GLD: $+$0.91, VIX: context \\[2pt]
\multicolumn{2}{l}{\scriptsize \textit{Scoring:} $\sigma_1 = 1 - \frac{1}{2n}\sum_i |\hat{v}_i-v_i^*|$ over eligible securities.}
\end{tabular}
\end{tcolorbox}

\vspace{2pt}
\begin{tcolorbox}[colback=nearwhite, colframe=softperi, boxrule=0.3pt, left=2pt, right=2pt, top=1pt, bottom=1pt, title={\scriptsize \textbf{S2: Signal Generation} \hfill Score = 0.488}]
{\scriptsize \textcolor{gray}{\textbf{Prompt summary:} Use the S1 views to return JSON with a buy, hold, or sell signal and signal strength for each supplied asset.}}\\[3pt]
\begin{tabular}{p{0.46\textwidth}p{0.46\textwidth}}
\textbf{Model Output} & \textbf{Reference Target} \\
\midrule
\scriptsize XLE: sell, USO: sell, TLT: buy, GLD: buy, VIX: buy, QQQ: sell & \scriptsize XLE: sell, USO: sell, TLT: buy, GLD: buy, VIX: context, QQQ: buy \\[2pt]
\multicolumn{2}{l}{\scriptsize \textit{Scoring:} Fraction of eligible-security signals matching the reference target.}
\end{tabular}
\end{tcolorbox}

\vspace{2pt}
\begin{tcolorbox}[colback=nearwhite, colframe=warmcream!80!black, boxrule=0.3pt, left=2pt, right=2pt, top=1pt, bottom=1pt, title={\scriptsize \textbf{S3: Weight Optimization} \hfill Score = 0.339}]
{\scriptsize \textcolor{gray}{\textbf{Prompt summary:} Use the S2 signals and supplied correlation information to return JSON with portfolio weights, expected return, volatility, and Sharpe.}}\\[3pt]
\begin{tabular}{p{0.46\textwidth}p{0.46\textwidth}}
\textbf{Model Output} & \textbf{S3 Reference} \\
\midrule
\scriptsize $w_\text{TLT}{=}0.079$, $w_\text{GLD}{=}0.071$, $w_\text{IAU}{=}0.071$, $w_\text{VIX}{=}0.063$ & \scriptsize $w_\text{ICSH}{=}0.775$, $w_\text{TLH}{=}0.100$, $w_\text{SHY}{=}0.042$, $w_\text{IEF}{=}0.041$ \\[2pt]
\multicolumn{2}{l}{\scriptsize \textit{Scoring:} $\sigma_3=0.5\,s_\text{acc}+0.5\,s_\text{corr}$ with $s_\text{acc}=1-\operatorname{TV}$ (Eq.~\ref{eq:app_sacc}). Here $s_\text{acc}=0.166$ and $s_\text{corr}=0.511$.}
\end{tabular}
\end{tcolorbox}

\vspace{2pt}
\begin{tcolorbox}[colback=nearwhite, colframe=lavblue, boxrule=0.3pt, left=2pt, right=2pt, top=1pt, bottom=1pt, title={\scriptsize \textbf{S4: Execution Simulation (deterministic)} \hfill Score = 0.492}]
{\scriptsize \textcolor{gray}{The deterministic S4 calculation defined above produces the following model and reference turnover.}}\\[3pt]
\begin{tabular}{p{0.46\textwidth}p{0.46\textwidth}}
\textbf{Model Turnover (from S3)} & \textbf{Reference Turnover} \\
\midrule
\scriptsize 87.4\% (change to the model target) & \scriptsize 177.4\% (change to the reference target) \\[2pt]
\multicolumn{2}{l}{\scriptsize \textit{Scoring:} $\sigma_4=1-|\tau-\tau^*|/\max(\tau,\tau^*,10^{-4})\approx 0.492$. Both changes start from the same pre-trade state.}
\end{tabular}
\end{tcolorbox}

\vspace{2pt}
\begin{tcolorbox}[colback=nearwhite, colframe=coralred, boxrule=0.3pt, left=2pt, right=2pt, top=1pt, bottom=1pt, title={\scriptsize \textbf{S5: Risk Monitoring (deterministic)} \hfill Score = 0.500}]
{\scriptsize \textcolor{gray}{The deterministic S5 calculation defined above produces the following model and reference risk states.}}\\[3pt]
\begin{tabular}{p{0.46\textwidth}p{0.46\textwidth}}
\textbf{Computed from Model Weights} & \textbf{Computed from Reference Weights} \\
\midrule
\scriptsize VaR: $-$0.71\%, MaxDD: $-$2.78\%, Drift: 6.7\%, Rebalance: Yes & \scriptsize VaR: $-$0.03\%, MaxDD: $-$0.12\%, Drift: 76.4\%, Rebalance: Yes \\[2pt]
\multicolumn{2}{l}{\scriptsize \textit{Scoring:} $\sigma_5=0.5\,d_\text{reb}+0.5\,\operatorname{clip}(1-(e_\text{VaR}+e_\text{DD})/2,0,1)$. Rebalance flags agree; lookback risk magnitudes differ.}
\end{tabular}
\end{tcolorbox}

\vspace{3pt}
{\scriptsize \textbf{Summary:} The model makes a large defensive reallocation. Its turnover of 87.4\% versus 177.4\% gives S4 of 0.492, the highest in this example set. The proposed VIX weight of .063 is held as uninvested cash. Adjacent score drops total 0.463, giving CEPS of 0.478.}

\end{tcolorbox}
\caption{Pipeline trace for GLM-5.1 during the COVID flash crash (balanced profile). The large defensive reallocation yields the highest S4 (0.492) among the three examples; the VIX output is held as uninvested cash.}
\label{fig:showcase_pipeline_glm}
\end{figure*}

\begin{figure*}[!ht]
\centering
\small
\begin{tcolorbox}[colback=white, colframe=deepslate, title={\small\textbf{Qwen3.7-Max} \hfill 2022-10-07 $\mid$ Aggressive $\mid$ 2022 Market Decline}, fonttitle=\small\bfseries, boxrule=0.4pt, left=3pt, right=3pt, top=2pt, bottom=2pt]

\begin{tcolorbox}[colback=nearwhite, colframe=coolblue, boxrule=0.3pt, left=2pt, right=2pt, top=1pt, bottom=1pt, title={\scriptsize \textbf{S1: Market Interpretation} \hfill Score = 0.768}]
{\scriptsize \textcolor{gray}{\textbf{Prompt summary:} Assess the market on 2022-10-07 from the supplied asset and macro features. Return JSON with a view for each supplied asset, a regime, confidence, and a macro summary.}}\\[3pt]
\begin{tabular}{p{0.46\textwidth}p{0.46\textwidth}}
\textbf{Model Output} & \textbf{Reference Target} \\
\midrule
\scriptsize QQQ: $-$0.40, XLE: $+$0.20, TLT: $-$0.20, BIL: $+$0.60, VIX: $+$0.60 & \scriptsize QQQ: $-$0.58, XLE: $+$1.00, TLT: $-$1.00, BIL: $+$0.00, VIX: context \\[2pt]
\multicolumn{2}{l}{\scriptsize \textit{Scoring:} $\sigma_1 = 1 - \frac{1}{2n}\sum_i |\hat{v}_i-v_i^*|$ over eligible securities.}
\end{tabular}
\end{tcolorbox}

\vspace{2pt}
\begin{tcolorbox}[colback=nearwhite, colframe=softperi, boxrule=0.3pt, left=2pt, right=2pt, top=1pt, bottom=1pt, title={\scriptsize \textbf{S2: Signal Generation} \hfill Score = 0.545}]
{\scriptsize \textcolor{gray}{\textbf{Prompt summary:} Use the S1 views to return JSON with a buy, hold, or sell signal and signal strength for each supplied asset.}}\\[3pt]
\begin{tabular}{p{0.46\textwidth}p{0.46\textwidth}}
\textbf{Model Output} & \textbf{Reference Target} \\
\midrule
\scriptsize QQQ: sell, XLE: buy, TLT: sell, BIL: buy, VIX: buy & \scriptsize QQQ: sell, XLE: buy, TLT: sell, BIL: hold, VIX: context \\[2pt]
\multicolumn{2}{l}{\scriptsize \textit{Scoring:} Fraction of eligible-security signals matching the reference target.}
\end{tabular}
\end{tcolorbox}

\vspace{2pt}
\begin{tcolorbox}[colback=nearwhite, colframe=warmcream!80!black, boxrule=0.3pt, left=2pt, right=2pt, top=1pt, bottom=1pt, title={\scriptsize \textbf{S3: Weight Optimization} \hfill Score = 0.398}]
{\scriptsize \textcolor{gray}{\textbf{Prompt summary:} Use the S2 signals and supplied correlation information to return JSON with portfolio weights, expected return, volatility, and Sharpe.}}\\[3pt]
\begin{tabular}{p{0.46\textwidth}p{0.46\textwidth}}
\textbf{Model Output} & \textbf{S3 Reference} \\
\midrule
\scriptsize $w_\text{SGOV}{=}0.232$, $w_\text{BIL}{=}0.231$, $w_\text{SHV}{=}0.165$, $w_\text{CSHI}{=}0.120$ & \scriptsize $w_\text{SGOV}{=}0.679$, $w_\text{PPLT}{=}0.264$, $w_\text{WEAT}{=}0.031$, $w_\text{CSHI}{=}0.026$ \\[2pt]
\multicolumn{2}{p{0.92\textwidth}}{\scriptsize \textit{Scoring:} $\sigma_3=0.5\,s_\text{acc}+0.5\,s_\text{corr}$ with $s_\text{acc}=1-\operatorname{TV}$ (Eq.~\ref{eq:app_sacc}). Here $s_\text{acc}=0.258$ and $s_\text{corr}=0.538$.}
\end{tabular}
\end{tcolorbox}

\vspace{2pt}
\begin{tcolorbox}[colback=nearwhite, colframe=lavblue, boxrule=0.3pt, left=2pt, right=2pt, top=1pt, bottom=1pt, title={\scriptsize \textbf{S4: Execution Simulation (deterministic)} \hfill Score = 0.025}]
{\scriptsize \textcolor{gray}{The deterministic S4 calculation defined above produces the following model and reference turnover.}}\\[3pt]
\begin{tabular}{p{0.46\textwidth}p{0.46\textwidth}}
\textbf{Model Turnover (from S3)} & \textbf{Reference Turnover} \\
\midrule
\scriptsize 3.4\% (change to the model target) & \scriptsize 140.0\% (change to the reference target) \\[2pt]
\multicolumn{2}{l}{\scriptsize \textit{Scoring:} $\sigma_4=1-|\tau-\tau^*|/\max(\tau,\tau^*,10^{-4})\approx 0.025$. Both changes start from the same pre-trade state.}
\end{tabular}
\end{tcolorbox}

\vspace{2pt}
\begin{tcolorbox}[colback=nearwhite, colframe=coralred, boxrule=0.3pt, left=2pt, right=2pt, top=1pt, bottom=1pt, title={\scriptsize \textbf{S5: Risk Monitoring (deterministic)} \hfill Score = 0.500}]
{\scriptsize \textcolor{gray}{The deterministic S5 calculation defined above produces the following model and reference risk states.}}\\[3pt]
\begin{tabular}{p{0.46\textwidth}p{0.46\textwidth}}
\textbf{Computed from Model Weights} & \textbf{Computed from Reference Weights} \\
\midrule
\scriptsize VaR: $-$2.18\%, MaxDD: $-$2.40\%, Drift: 22.1\%, Rebalance: Yes & \scriptsize VaR: $-$0.74\%, MaxDD: $-$2.05\%, Drift: 66.7\%, Rebalance: Yes \\[2pt]
\multicolumn{2}{l}{\scriptsize \textit{Scoring:} $\sigma_5=0.5\,d_\text{reb}+0.5\,\operatorname{clip}(1-(e_\text{VaR}+e_\text{DD})/2,0,1)$. Rebalance flags agree; lookback risk magnitudes differ.}
\end{tabular}
\end{tcolorbox}

\vspace{3pt}
{\scriptsize \textbf{Summary:} Both portfolios hold cash equivalents, with different weights and turnover. Model turnover is 3.4\% versus 140.0\%, giving S4 of 0.025. The proposed VIX weight of .072 is held as uninvested cash. Adjacent score drops total 0.743, giving CEPS of 0.373.}

\end{tcolorbox}
\caption{Pipeline trace for Qwen3.7-Max during the 2022 Market Decline (aggressive profile). The cash-heavy model allocation changes little relative to the reference, giving S4 of 0.025.}
\label{fig:showcase_pipeline_qwen37}
\end{figure*}

\end{document}